\documentclass{article} 

\usepackage{iclr2027_conference,times}

\usepackage{amsmath,amsfonts,bm}

\def\eqref#1{equation~\ref{#1}}

\def\1{\bm{1}}

\DeclareMathAlphabet{\mathsfit}{\encodingdefault}{\sfdefault}{m}{sl}
\SetMathAlphabet{\mathsfit}{bold}{\encodingdefault}{\sfdefault}{bx}{n}

\usepackage{graphicx}
\usepackage{wrapfig}
\usepackage{algorithm}
\usepackage{algpseudocode}
\usepackage{caption}
\usepackage{microtype}
\usepackage{hyperref}
\usepackage{url}
\usepackage{amssymb}
\usepackage{subcaption}
\usepackage{multirow}
\usepackage{booktabs}
\usepackage{threeparttable}
\usepackage{xcolor}
\usepackage{listings}
\usepackage{tcolorbox}
\tcbuselibrary{breakable}
\usepackage{adjustbox}
\usepackage{tabularx}

\title{Alignment by Confident Debate: Preference Learning from Multi-Agent Debate}

\author{\parbox{\textwidth}{\centering
Nikhil Verma, Dingxian Wang, Mohammad Reza Rezaei, \\
Andrew Rabinovich \& Zhao Chen \\[8pt]
{\normalfont Upwork Inc. \\
Email: \texttt{nikhilverma@upwork.com}}}
}

\newcommand{\method}{AbCD}

\iclrfinalcopy 
\begin{document}

\maketitle

\begin{abstract}

Post-training language models for reasoning typically rely on gold answers, task-specific verifiers, or learned reward models, limiting scalability when such signals are scarce.
Multi-agent debate offers a self-supervised alternative, but converting debate into reliable training supervision remains challenging:
majority voting weights agreeing responses equally, lacks a principled tie-breaking rule, and is ill-defined for open-ended tasks without a canonical answer;
likelihood-based scoring can conflate fluency with correctness; 
and homogeneous committees may reinforce correlated errors.
We introduce \textbf{Alignment by Confident Debate} (\method{}), which converts debate traces into preference data without gold answer labels. 
A sampling-diverse committee debates each problem, after which final answers are clustered by task-level semantic equivalence. 
%
Each cluster is scored by aggregating confidence on its claimed answer, and completions from the highest-scoring cluster are labeled as preferred over the remaining completions.
We use verifiable tasks~(mathematical reasoning, multiple-choice quantitative QA, and executable code-generation) as a controlled test bed: their answer keys or executable checkers let us evaluate the quality of debate-derived supervision against ground truth, even though the oracle itself never uses gold labels.
Post-training on the resulting preferences outperforms inference-only ensembling and debate, as well as SFT on the same traces, across five model families spanning 1B to 8B parameters and four datasets, improving consensus accuracy, with a {maximum} gain of $21.5_{pp}$.
Ablations show that the approach transfers to a larger model and to code generation, while also revealing limitations of heterogeneous committees and agreement-based weighting. 
{On tasks with a discrete answer, majority voting is the standard ensemble and lands close to confidence ensembling, because both aggregate the same extracted strings; that comparison is reported in the appendix.}
Overall, the results suggest that debate traces can yield useful preference supervision when answer equivalence is combined with confidence-weighted disagreement.
\end{abstract}

\section{Introduction}
\label{sec:introduction}

Post-training for reasoning typically relies on external supervision to identify desirable solutions or solution traces. 
Such supervision can come from labeled demonstrations or gold answers, task-specific verifiers that evaluate final outcomes, or learned evaluators such as reward models and LLM-based judges.
Gold answers and executable verifiers provide direct correctness signals when an answer key or reliable checking procedure exists, as in numerical reasoning and unit-tested code \citep{cobbe2021gsm8k,yuan2023scaling,austin2021program}. 
Reward models and LLM-based judges extend evaluation to less readily verifiable outputs \citep{ouyang2022rlhf,zheng2023judging,yuan2024selfrewarding,wu2025self}.
However, these evaluators must themselves be trained, calibrated, or otherwise trusted and may share systematic errors with the policy being optimized. 
Thus, although post-training objectives are well established, obtaining reliable correctness supervision remains difficult and domain-dependent.

\begin{wrapfigure}[21]{r}{0.4\columnwidth}  
\centering
\vspace{-10pt}
\captionsetup{width=\linewidth,font=small,skip=4pt}
\includegraphics[width=\linewidth]{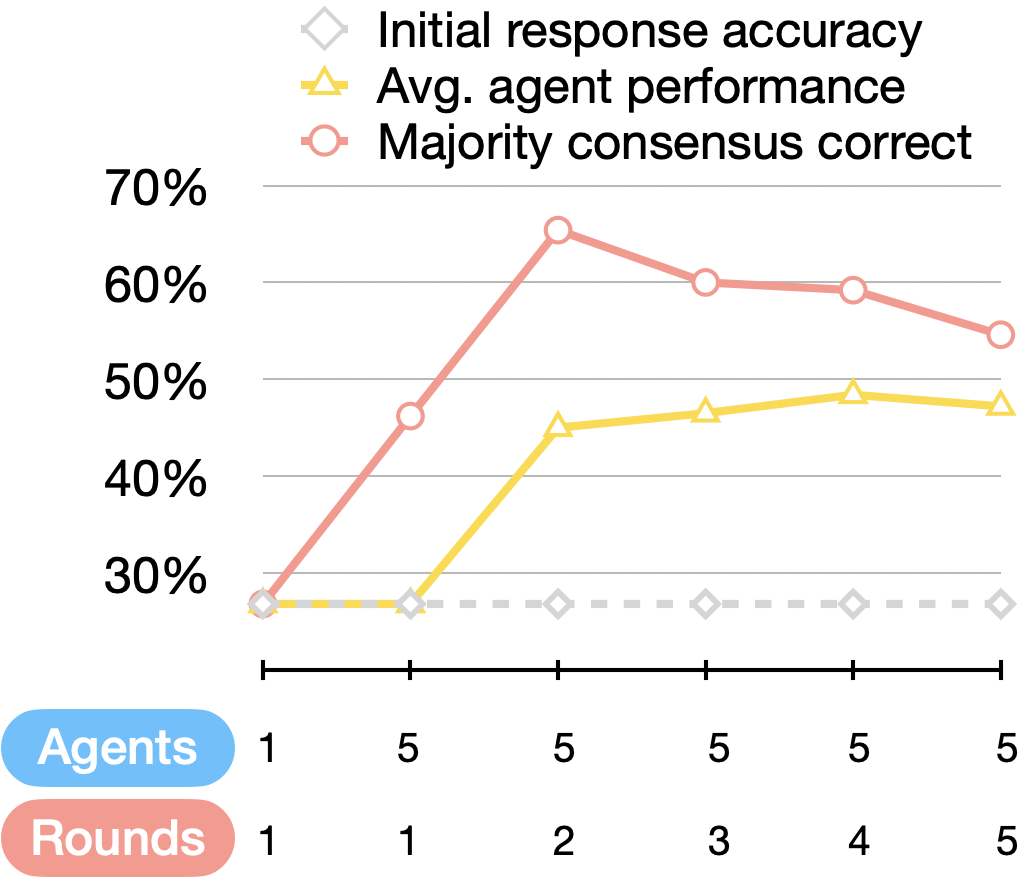}
\caption{
    Committee behavior for Qwen-2.5-1.5B-it on GSM8K.
    The committee size varies from one to five agents under independent sampling, after which the number of debate rounds is varied with five agents.
    Additional rounds increase agreement but do not improve consensus accuracy beyond the first exchange and can reduce correctness.
}
\label{fig:majority_consensus_rounds}
\end{wrapfigure}

When external correctness supervision is unavailable, agreement among the model's own samples provides a natural candidate signal. 
Multi-agent debate allows several agents to exchange reasoning over multiple rounds, and can improve answer accuracy relative to a single sampled solution \citep{du2023debate,liang2023mad,madaan2023selfrefine}. 
Subsequent debate-training methods often use these traces for training, for example by labeling the majority answer as preferred and the remaining answers as dispreferred
\citep{samanta2026maca,zhou2025debate}. 
This strategy is fragile, however. 
In homogeneous committees, communication need not improve correctness beyond independent sampling \citep{zhu2026demystifying}; repeated interaction can also induce correlated reasoning and convergence on shared errors \citep{dai2026consensus}. 
On GSM8K, for example, a five-agent committee achieves its largest consensus accuracy after the first exchange, whereas additional rounds increase agreement without improving—and in some cases reducing—correctness (Figure~\ref{fig:majority_consensus_rounds}). 
Consensus is therefore an outcome of the interaction and selection process, not by itself evidence that the selected answer is correct.
%
%
%
We therefore study how to convert debate traces into preference supervision by separating answer equivalence from answer selection: final answers are clustered by task-level semantic equivalence and the resulting clusters are ranked by aggregate confidence in their claimed answer spans.

Majority voting requires a procedure for mapping final answers to equivalent outcomes. 
Without a canonical answer or reliable semantic-equivalence rule, voting cannot determine when two responses should count as equivalent, particularly for open-ended generation where correct responses may differ in wording~\citep{chen2024usc}. 
Even when equivalence is available, majority voting treats all agreeing traces equally and provides no additional signal for resolving ties. 
In small committees of cloned models, agreement may also reflect correlated errors rather than independent corroboration. 
Token probabilities provide another signal available during generation, but they measure the likelihood of generated tokens rather than correctness of the full reasoning chain. 
Length-normalized full-trace likelihood can therefore favor fluent or familiar reasoning without reliably identifying the correct answer.
Confidence computed on answer-relevant tokens is a more targeted alternative and has been used to improve answer selection \citep{razghandi2025cer,lin2025confmad}. 
Nevertheless, these scores remain imperfect proxies because language models can be overconfident, and identifying the answer-relevant span may itself depend on the task. 
Exisiting confidence-based approaches largely use such scores for inference-time selection; we instead use them to construct preference data from debate traces.

That gap motivates \textbf{Alignment by Confident Debate} (\method{}), which constructs preference data from debate traces without using gold answer labels for supervision. 
A sampling-diverse committee generates and exchanges solutions over multiple debate rounds, after which final answers are clustered by task-level semantic equivalence. 
We score each cluster by aggregate confidence on its claimed answer-span rather than by the likelihood of the full reasoning trace. 
Completions from the highest-scoring cluster are labeled preferred, while completions from the remaining clusters are labeled dispreferred. 
Agreement rate is recorded as a reliability signal and can optionally be incorporated as a soft weight, but it does not define the winning cluster.
On GSM8K, SVAMP, and MathQA, confidence-based cluster selection identifies the correct answer cluster more often than majority voting, particularly on instances with tied vote counts.
Post-training on the resulting preferences improves both selected-answer and mean agent accuracy, outperforming inference-only ensembling, inference-only debate, and SFT trained on the same debate traces across the evaluated settings.
The same comparison holds for a larger model, for heterogeneous committees, and for execution-based evaluation on code generation.
{Where an answer is a discrete string, majority voting is the usual ensemble and should track confidence ensembling, so we report the comparison in the appendix~\ref{app:sec:basline_maca_comparison}. 
Outside verifiable tasks a vote is no longer defined in the same way, and \method{} still is, selecting the preferred cluster by confidence in the claimed answer.
Ablations on coding and open-ended tasks are reported in Appendix~\ref{app:sec:mbpp-oracle} and ~\ref{app:sec:non_equivalence_tasks}.
}




\noindent\textbf{Contributions.} We make the following contributions:
\begin{enumerate}
    \item We propose a confidence-based debate oracle that separates task-level answer equivalence from answer selection:
    equivalence groups completions into answer clusters, while confidence mass on the claimed-answer span determines the cluster weight.

    \item We construct a gold-label-free preference-data pipeline that converts the oracle's cluster assignments into preferred and dispreferred completions for standard post-training objectives.

    \item We characterize the effects of committee diversity and debate depth through controlled experiments. 
    Empirical observations suggest that sampling-diverse committees capture most of the observed benefit.
\end{enumerate}

\section{Related work}
\label{sec:related_literarure}

\paragraph{Multi-agent debate at inference.}
Self-consistency aggregates independently sampled solutions by majority vote when final answers can be mapped to equivalent outcomes~\citep{wang2023selfconsistency},
and universal self-consistency extends this idea to free-form outputs by using
an LLM to select a consistent answer among candidates~\citep{chen2024usc}.
Multi-agent debate instead allows agents to exchange reasoning over multiple rounds and can improve answer selection relative to a single sampled solution~\citep{du2023debate,liang2023mad,madaan2023selfrefine}. 
However, these gains are not guaranteed: under homogeneous agents, debate preserves expected correctness over time and can underperform independent voting~\citep{choi2026debate,zhu2026demystifying}, and repeated interaction can increase similarity among agents' rationales, amplifying shared errors~\citep{dai2026consensus}. 
These findings motivate treating consensus as an outcome of the interaction and selection process rather than as a correctness label. 

\paragraph{Confidence-based answer selection.}
Token likelihoods provide an alternative to vote counts, but full-response
likelihood is a coarse signal because it scores all generated tokens, including
reasoning tokens that may be only weakly tied to final-answer correctness.
Confidence computed on answer-relevant tokens provides a more targeted
correctness signal~\citep{razghandi2025cer,lin2025confmad}, and related work
uses confidence estimates~(including verbalized confidence, token-distribution
statistics, and calibrated answer-token scores) to select answers at inference
time~\citep{taubenfeld2025cisc,kang2025selfcertainty,kadavath2022know,%
tian2023verbalized}. 
Our work differs in using claimed answer-span confidence to \emph{construct preferred and dispreferred traces} for post-training.
An extended discussion of related work, including a finer-grained taxonomy
of debate-based and confidence-based methods, is provided in
Appendix~\ref{app:sec:lit_review}.

\paragraph{Self-training from unlabeled or weakly labeled traces.}
Prior methods construct training signals from verifier-approved traces~\citep{zelikman2022star,gulcehre2023rest}, pseudo-labels derived from sampled answers~\citep{zhang2025ttrl,shao2024deepseekmath}, or judgments from another language model~\citep{yuan2024selfrewarding,bai2022constitutional}.
Other work uses debate traces for imitation, preference learning, or multi-agent reinforcement learning~\citep{subramaniam2025multiagent,srivastava2025dte,samanta2026maca,zhou2025debate,motwani2025malt,liu2026magrpo}. 
These methods differ in how they identify preferred traces and assign training weights. 
To our knowledge, prior debate-training methods do not construct preference pairs by ranking task-level answer-equivalence clusters using aggregate claimed answer-span confidence. 
\method{} combines answer-equivalence with span-confidence mass, while treating agreement as a reliability statistic rather than as the definition of the winning answer.

\section{Methodology}
\label{sec:methodology}

\begin{figure}[t]
\centering
\includegraphics[width=\linewidth]{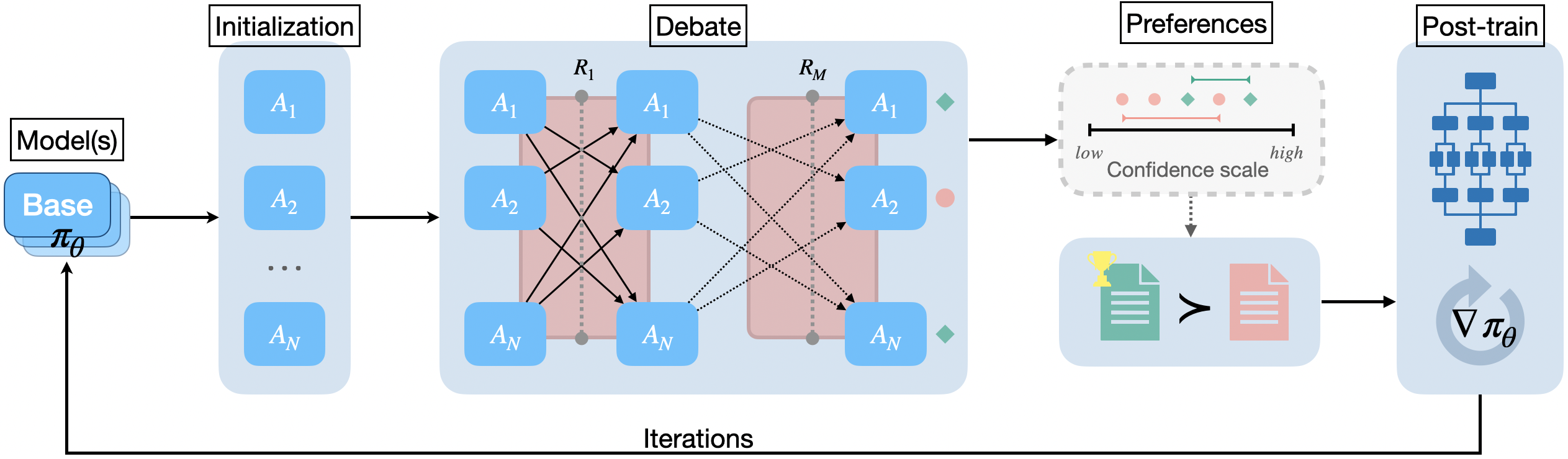}
\caption{
    \method{} instantiates $N$ sampling-diverse agents from the base policy $\pi_\theta$ and lets them debate for $M$ generations.
    Final traces are clustered by task-level answer equivalence, and the cluster with the largest aggregate confidence on the claimed-answer span is selected.
    Traces in the winning cluster are preferred, while the remaining traces are dispreferred (green $\succ$ red).
    The resulting preference supervision updates $\pi_\theta$ without gold answers in the oracle.
}
\label{fig:acd_method}
\end{figure}

Let $\pi_\theta$ be the base instruction-tuned policy, which produces a reasoning trace for each query $x$. 
We instantiate $N$ sampling-diverse agents from $\pi_\theta$ per query and run $M$ rounds of multi-agent debate.
The final traces are clustered according to task-level answer equivalence.
Aggregate confidence on the claimed-answer span selects the winning cluster, whose traces are treated as preferred and the remaining traces as dispreferred.
A learner initialized from $\pi_\theta$ is then trained using one of several objectives: supervised fine-tuning (SFT), preference optimization~(DPO or KTO), or group-relative debate-derived preference reward optimization~(GRPO).
Figure~\ref{fig:acd_method} and Algorithm~\ref{app:alg:method} summarize the pipeline.
The post-training objectives are standard; the novelty of \method{} lies in its confidence-based oracle for selecting preferred traces.
The following sections detail the committee, confidence score, equivalence-based oracle, and post-training objectives.

\subsection{Sampling-diverse committee}
\label{sec:debate}

The autoregressive policy~($\pi_\theta$) generates a trace $y=(y_1,\ldots,y_{|y|})$ conditioned on an input prompt $x\in\mathcal X_{\mathrm{train}}$:

\[
\pi_\theta(y\mid x)
=
\prod_{t=1}^{|y|}
\pi_\theta(y_t\mid x,y_{<t})
\]
A task parser $\mathcal A$ extracts the claimed answer $a=\mathcal A(y)$.
For each prompt $x$, we instantiate $N$ agents $\{\pi_i\}_{i=1}^{N}$ from the base policy $\pi_\theta$.
The agents may share parameters but use heterogeneous decoding configurations, such as different temperatures, role prompts, or sampling seeds.
We refer to this as a sampling-diverse committee.

At round $1$, agents generate traces independently conditional on $x$:
\[
y_i^{(1)}\sim\pi_i(\cdot\mid x),
\qquad i=1,\ldots,N
\]
Let $\operatorname{ser}_i$ serialize the previous-round traces shown to agent $i$.
For exchange rounds $m=2,\ldots,M$, agents synchronously revise their traces:
\[
y_i^{(m)}
\sim
\pi_i\!\left(
\cdot\mid x,\,
\operatorname{ser}_i\!\left(
\{y_j^{(m-1)}\}_{j\ne i}
\right)
\right)
\]
Thus, the procedure consists of one independent generation followed by
$M-1$ exchange rounds. The final debate collection is
\[
\mathcal Y(x)=\bigl(y_1,\ldots,y_N\bigr),
\qquad
y_i=y_i^{(M)},
\qquad
a_i=\mathcal A(y_i)
\]
Traces for which $\mathcal A(y_i)$ is undefined are excluded from equivalence
clustering, assigned to the dispreferred set during preference construction, and
counted as incorrect during evaluation.

\subsection{Answer-confidence score}
\label{sec:answer_confidence}

We define $c_i^{\mathrm{answer}}$ as the answer-confidence score for agent $i$'s generated trace and consider two constructions: full-trace confidence $c_i^{\mathrm{trace}}$ and claimed answer-span confidence $c_i^{\mathrm{span}}$. 
For each generated token, the corresponding log-probability is $\ell_{i,t} = \log \pi_i\!\left(y_{i,t}\mid x, \operatorname{ser}_i\!(\{y_j^{(m-1)}\}_{j\ne i}), y_{i,<t}\right)$.
All confidence scores below reuse these recorded token log-probabilities.

\paragraph{Trace confidence.}
We define the confidence of the full generated reasoning trace as the geometric mean of its
token probabilities:
\begin{equation}
\label{eq:trace-conf}
c_i^{\mathrm{trace}}
=
\exp\!\left(
\frac{1}{|y_i|}
\sum_{t=1}^{|y_i|}\ell_{i,t}
\right)
\end{equation}

\paragraph{Answer-span confidence.}
Let $S(y_i)\subseteq\{1,\ldots,|y_i|\}$ denote the token indices selected by a task-specific answer-span parser. 
For a nonempty answer-span, $S(y_i)\neq\varnothing$, we define
\begin{equation}
\label{eq:span-conf}
c_i^{\mathrm{span}}
=
\exp\!\left(
\frac{1}{|S(y_i)|}
\sum_{t\in S(y_i)}\ell_{i,t}
\right)
\end{equation}
This score is the geometric mean of the token probabilities on the selected claimed answer-span. 
%
%
If $S(y_i)=\varnothing$ or cannot be aligned with the generated tokens, then
$c_i^{\mathrm{span}}$ is undefined.

\subsection{Equivalence clusters and the confidence-weighted oracle}
\label{sec:oracle}

For each prompt $x$, parseable answers are partitioned into equivalence clusters
using a task-specific relation $\equiv$ that compares candidate answers with one
another. 
The relation is evaluated without access to the gold answer. 
Let $\mathcal I_k$ denote the indices of the agents whose answers belong to cluster
$k$. 
We define the cluster size and aggregate answer confidence as
\[
n_k = |\mathcal I_k|,
\qquad
w_k
=
\sum_{i\in\mathcal I_k} c_i^{\mathrm{answer}}
\]

We select the winning cluster by lexicographic maximization
$k^\star
=
\operatorname*{arg\,max}_{k}
\bigl(w_k,n_k\bigr), $
which prioritizes aggregate answer confidence and uses cluster size as a secondary tie-break. 
The oracle answer is the canonical representative of the selected cluster $\hat a(x)=\operatorname{rep}(k^\star)$, where $\operatorname{rep}(k^\star)$ returns a canonical string for cluster $k^\star$.
If multiple clusters remain tied after both criteria, the prompt is excluded from oracle construction.
For comparison, majority voting selects the cluster with the largest support,
$k_{\mathrm{maj}}=\operatorname*{arg\,max}_{k} n_k$.

We construct preferred and dispreferred trace sets relative to $\hat a(x)$:
\begin{equation}
\label{eq:preference-partition}
\mathcal G^+(x)
=
\{y_i\in\mathcal Y(x):a_i\equiv\hat a(x)\},
\qquad
\mathcal G^-(x)
=
\{y_i\in\mathcal Y(x):a_i\not\equiv\hat a(x)\}
\end{equation}
Unparseable traces are excluded from clustering and assigned to $\mathcal G^-$ during preference construction. 
Preference objectives use only prompts for which both $\mathcal G^+(x)$ and $\mathcal G^-(x)$ are nonempty.
Unanimous debates may still contribute their preferred traces to SFT and GRPO.
We also measure the support for the selected cluster by
\[
\alpha(x)
=
\frac{|\mathcal G^+(x)|}{N}
\]
The statistic $\alpha(x)$ can optionally be used as a soft weight for the loss contribution of a preference example.
It measures support for the confidence-selected winner and does not determine which cluster wins.

\subsection{Post-training}
\label{sec:training}

For each prompt with a defined oracle, we retain the oracle answer and its winner support in
\[
\mathcal D_{\mathrm{oracle}}
=
\left\{
\bigl(x,\hat a(x),\alpha(x)\bigr)
\right\}_{x\in\mathcal X_{\mathrm{train}}}
\]
For preference-based objectives, we additionally retain prompts for which both $\mathcal G^+(x)$ and $\mathcal G^-(x)$ are nonempty:
\[
\mathcal D_{\mathrm{pref}}
=
\left\{
\bigl(x,\mathcal G^+(x),\mathcal G^-(x),\alpha(x)\bigr)
\right\}_{x\in\mathcal X_{\mathrm{train}}}
\]
The preferred and dispreferred sets provide the supervision for the corresponding post-training objective.
SFT maximizes the likelihood of traces in $\mathcal G^+(x)$. 
DPO forms preference pairs $(y^+,y^-)\in\mathcal G^+(x)\times\mathcal G^-(x)$, while KTO uses the corresponding unpaired preferred and dispreferred traces. 
The contribution of each preference example can be optionally weighted by $\alpha(x)$.
For GRPO, the oracle answer $\hat a(x)$ is retained as a fixed pseudo-target.
Each newly sampled response $y$ receives the reward
$r_x(y)
=
\mathbf 1\!\left[\mathcal A(y)\equiv\hat a(x)\right].$
%
This reward measures agreement with the frozen debate oracle rather than correctness against a gold answer.
The reward or the corresponding loss contribution can be optionally scaled by $\alpha(x)$.

\section{Experiments and Results}
\label{sec:experiment_results}

We evaluate \method{} in three stages: committee configuration, oracle selection, and post-training. 
We first fix the committee and debate configuration through inference-only experiments, then compare full-trace and answer-span confidence on identical transcripts, and finally lock the winning configuration for all post-training experiments. 
This separates decisions about generating debate traces from decisions about labeling
and using those traces for training.

\subsection{Experimental setup}
\label{sec:experimental_setup}
\textbf{Models.}
We post-train five instruction-tuned models: Qwen-2.5-1.5B-it, Llama-3.2-1B-it, Phi-4-mini-3.8B-it, Gemma-3-4B-it, and Llama-3.1-8B-it \citep{qwen2,llama32,phi4mini,team2025gemma}.
Unless otherwise noted, post-training uses a single iteration from the instruction-tuned checkpoint with QLoRA (4-bit base model, rank $r{=}32$).
\textbf{Datasets.}
%
%
We evaluate mathematical reasoning on GSM8K~\citep{cobbe2021gsm8k} and SVAMP~\citep{patel2021svamp}, quantitative question answering on MathQA~\citep{amini2019mathqa}, and code generation on MBPP~\citep{austin2021program}.
GSM8K and SVAMP use boxed numerical answers, MathQA uses answer-choice letters, and MBPP uses executable Python programs; the corresponding dataset and task-equivalence details are provided in Appendix~\ref{app:sec:data}.
Gold answers are used only for evaluation and designated ablations, never for debate, oracle construction, or post-training.
\textbf{Debate protocol.}
%
Each prompt is answered by a sampling-diverse committee of $N{=}5$ agents for $M{=}2$ rounds: independent generation followed by one exchange round.
The agents use temperatures $\{0.01,0.25,0.5,0.75,1.0\}$.
Unless otherwise noted, decoding uses top-$p{=}0.9$ and at most $256$ new tokens. 
Detailed inference and training configurations are given in Appendix~\ref{app:sec:training_details}.
\textbf{Metrics.}
%
%
We report \emph{pre-debate agent accuracy}, the mean accuracy of the independently sampled responses; \emph{post-debate agent accuracy}, the mean accuracy of the $N$ responses after the final round; and
\emph{consensus accuracy}, the accuracy of the answer cluster selected by the confidence-based oracle.
Accuracies are computed against gold labels on the full test set, with missing or unparseable outputs counted as incorrect.

\subsection{Committee Configuration}
\label{sec:committee}

\begin{figure*}[t]
    \centering
    \begin{subfigure}[b]{0.254\textwidth}
        \centering
        \includegraphics[width=\linewidth]{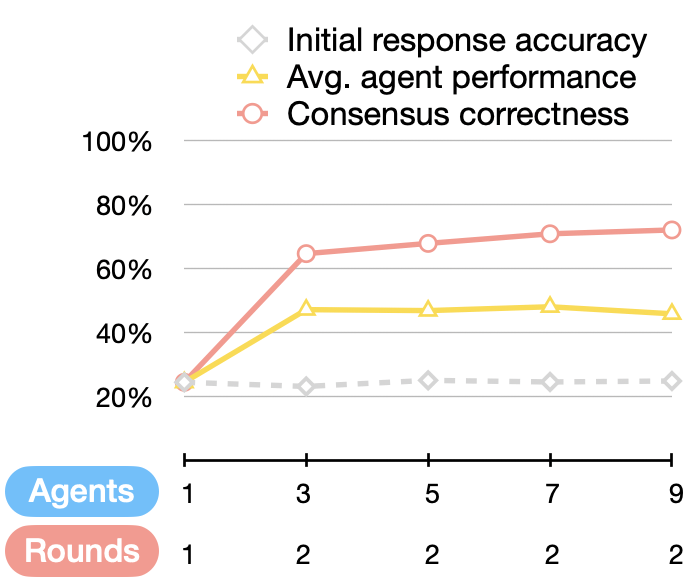}
        \caption{\# Agents}
        \label{fig:committee_agents_qwen2b}
    \end{subfigure}
    \hfill
    \begin{subfigure}[b]{0.23\textwidth}
        \centering
        \includegraphics[width=\linewidth]{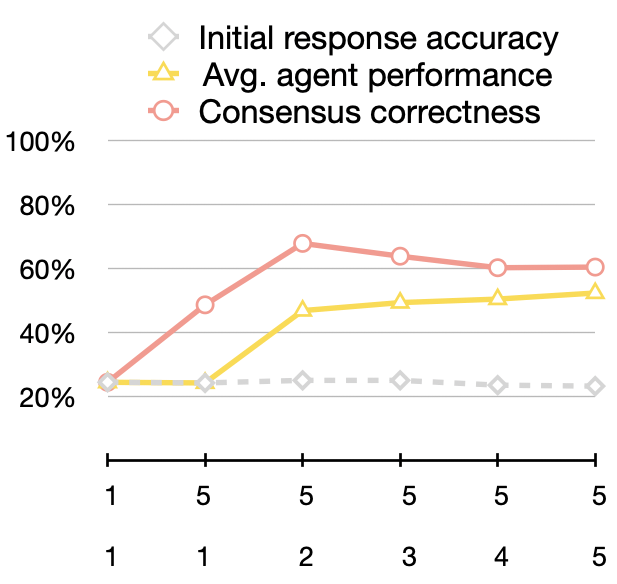}
        \caption{\# Debate rounds}
        \label{fig:committee_rounds}
    \end{subfigure}
    \hfill
    \begin{subfigure}[b]{0.234\textwidth}
        \centering
        \includegraphics[width=\linewidth]{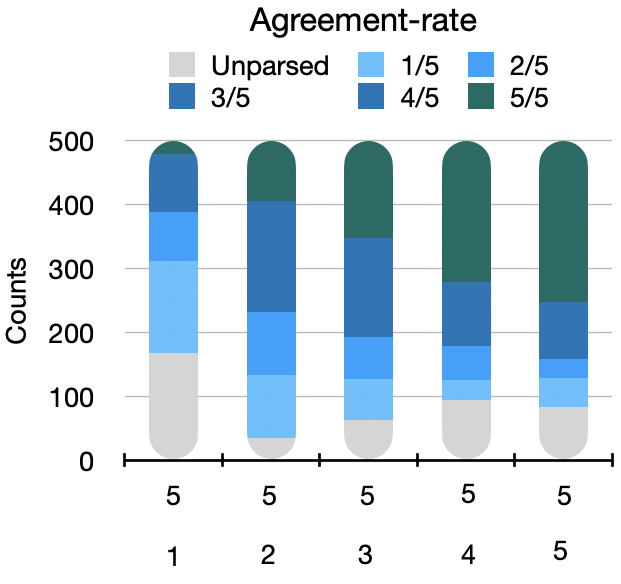}
        \caption{Agreement-rate}
        \label{fig:agreement_rounds}
    \end{subfigure}
    \hfill
    \begin{subfigure}[b]{0.252\textwidth}
        \centering
        \includegraphics[width=\linewidth]{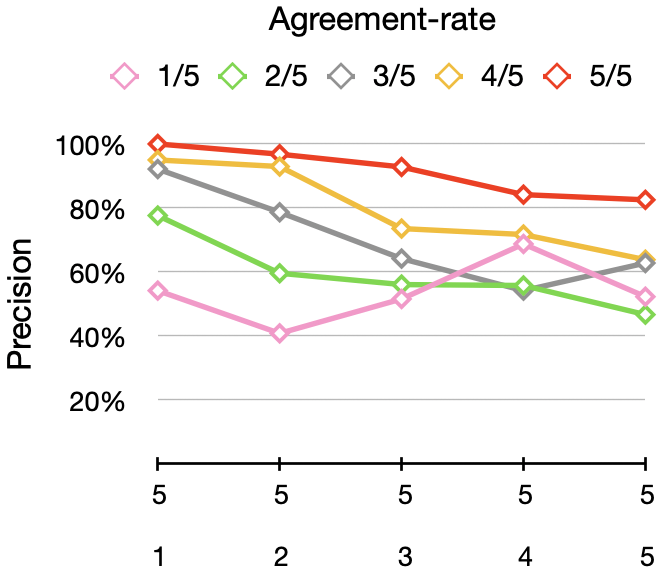}
        \caption{Correctness status}
        \label{fig:correctness_rounds}
    \end{subfigure}

    \caption{
        Committee behavior for Qwen-2.5-1.5B-it on GSM8K.
        Subfigures show accuracy by committee size~(a), accuracy by debate rounds~(b), agreement-rate across rounds (c), and correctness across rounds~(d).
    }
    \label{fig:committee_sweep_main}
\end{figure*}

We first characterize committee behavior using inference-only experiments to determine the default configuration for debating. 
Varying committee size from $N{=}1$ to $N{=}9$ with independent sampling at temperature $0.5$ shows that consensus accuracy improves up to $N{=}5$, with diminishing returns thereafter~(Figure~\ref{fig:committee_agents_qwen2b}). 
The first exchange round provides an additional gain, while further rounds increase agreement~(Figure~\ref{fig:agreement_rounds}) without improving consensus accuracy and can reduce correctness~(Figure~\ref{fig:correctness_rounds}). 
We observe the same trend across models and datasets; additional results are provided in Appendix ~\ref{app:sec:committee_conf_details}.
%
This divergence shows that agreement alone is not a reliable proxy for correctness, motivating an oracle that uses more than vote count.
We therefore lock $N{=}5$ and $M{=}2$ as the default. 
Among the diversity strategies evaluated, temperature diversity outperforms homogeneous clones, role diversity, and seed diversity (Appendix Table~\ref{app:tab:committee_diversity}); 
we therefore use temperature-diverse committees by default.

\subsection{Trace Confidence vs.\ Span Confidence}
\label{sec:trace_vs_span}

\begin{wrapfigure}[15]{r}{0.36\columnwidth}  
\vspace{-32pt}
\centering
\captionsetup{width=\linewidth,font=small,skip=4pt}
\includegraphics[width=0.9\linewidth]{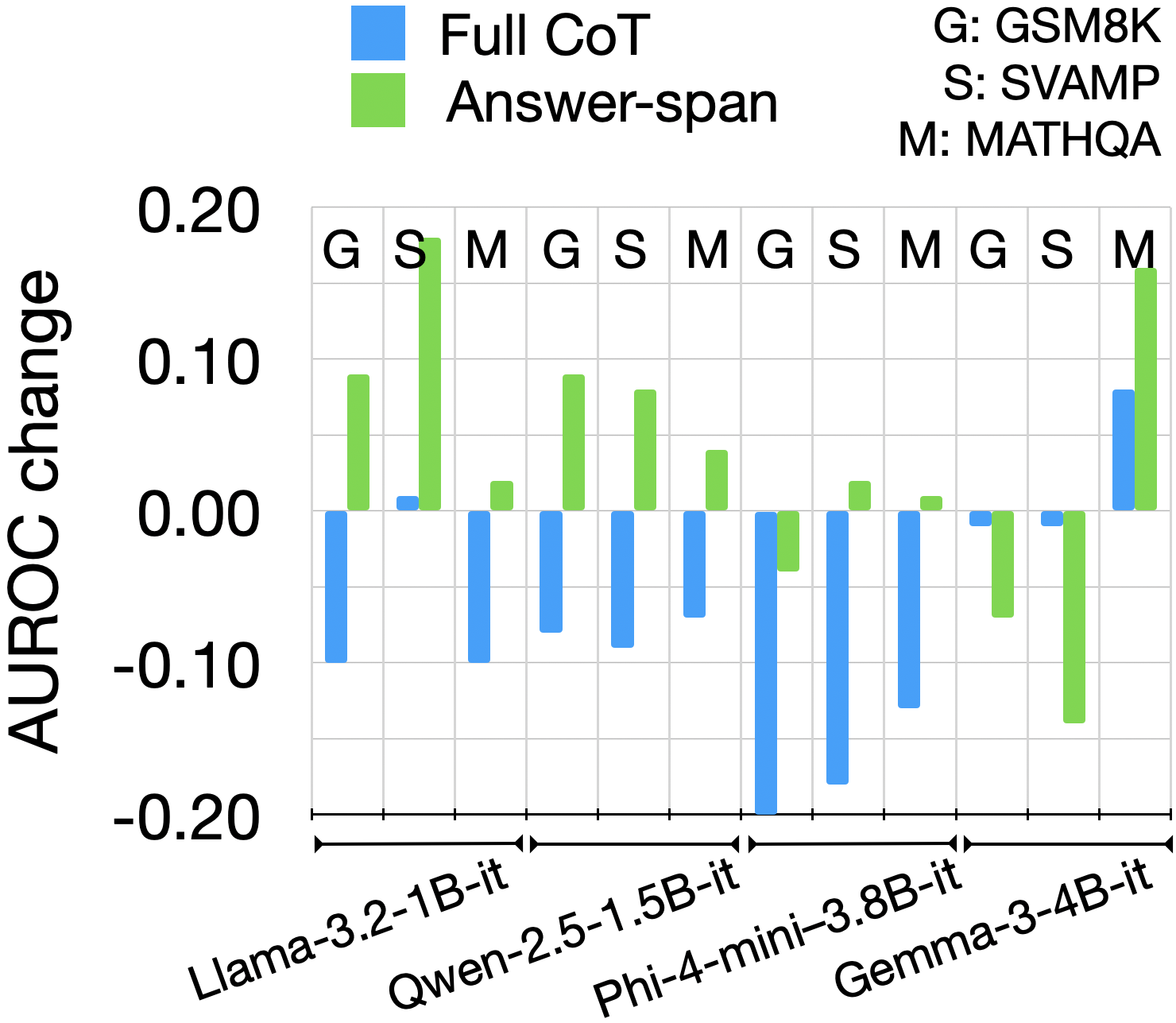}
\caption{
    Change in AUROC from the initial-response round to the first exchange round for full-trace and claimed answer-span confidence. 
    Each bar is one model--dataset pair across GSM8K, SVAMP, and MathQA. 
    Positive values indicate improved discrimination after debate.
}
\label{fig:trace_vs_span_auroc}
\end{wrapfigure}

Before constructing the preference dataset, we determine which portion of each response should be used to compute confidence.
For agent $i$, we compare full-trace confidence $c_i^{\mathrm{trace}}$ with claimed answer-span confidence $c_i^{\mathrm{span}}$, computed from the same debate transcripts to isolate the effect of the scored tokens.
Figure~\ref{fig:trace_vs_span_auroc} and Table~\ref{app:tab:trace_vs_span_confidence} report AUROC for distinguishing correct from incorrect final answers before and after the first exchange.
Across 12 model--dataset pairs, full-trace AUROC decreases by an average of $0.07$ after the first exchange, whereas answer-span AUROC increases by an average of $0.04$~(details in Appendix~\ref{app:sec:conf_comparison}).
This pattern is consistent with full-trace likelihood capturing properties of intermediate reasoning tokens whose fluency need not reflect final-answer correctness.
Although the trend is not uniform across models, the aggregate comparison motivates using the claimed answer-span; we therefore set $c_i^{\mathrm{answer}} := c_i^{\mathrm{span}}$ in all subsequent experiments.

\subsection{Preference Training with the Span Oracle}
\label{sec:preference_training}

\begin{table*}[t]
\centering
\small
\caption{
    Consensus accuracy~(\%) for inference-only baselines and post-training objectives.
    \emph{Ensemble} denotes inference-only independent sampling, and \emph{Debate} denotes inference-only multi-round debate.
    SFT, DPO, KTO, and GRPO are trained on the same debate traces, with preferred completions selected by the answer-span confidence oracle.
    The best post-training result for each model--dataset pair is in bold.
    The final column reports the absolute improvement of the best post-training method over inference-only debate, in percentage points.
}
\begin{adjustbox}{max width=\textwidth}
\begin{tabular}{ll | ccc|ccc c}
\toprule
\multirow{2}{*}{Model} & \multirow{2}{*}{Dataset}
& \multicolumn{3}{c|}{Baselines}
& \multicolumn{3}{c}{Post-training}
& \multirow{2}{*}{Best $\Delta$} \\
\cmidrule(lr){3-8}
& & Ensemble & Debate & SFT & DPO & KTO & GRPO & \\

\midrule
\multirow{3}{*}{Llama-3.2-1B-it}
  & SVAMP  & 18.3 & 19.0 & $25.2_{\pm 2.12}$ & $\textbf{40.5}_{\pm 1.70} $& $36.2_{\pm 4.03}$ & $27.2_{\pm 1.20}$ & $21.5_{pp}$ \\
  & GSM8K  & 14.8 & 15.6 & $15.6_{\pm 1.13}$ & $\textbf{27.0}_{\pm 3.96}$ & $21.6_{\pm 4.24}$ & $14.8_{\pm 3.11}$ & $11.4_{pp}$ \\
  & MathQA & 7.2 & 5.0 & $6.3_{\pm 1.56}$ & $\textbf{9.2}_{\pm 1.41}$ & $8.3_{\pm 1.27}$ & $6.7_{\pm 1.56}$ & $4.2_{pp}$ \\

\midrule
\addlinespace

\multirow{3}{*}{Qwen-2.5-1.5B-it}
  & SVAMP  & 77.0 & 79.7 & $78.7_{\pm 0.49}$ & $79.7_{\pm 0.92}$ & $78.0_{\pm 2.83}$ & $\textbf{80.3}_{\pm 2.33}$ & $0.6_{pp}$\\
  & GSM8K  & 44.5 & 66.0 & ${66.3}_{\pm 0.71}$ & $\textbf{71.4}_{\pm 0.28}$ & ${69.2}_{\pm 1.13}$ & ${68.9}_{\pm 0.14}$ & $5.4_{pp}$ \\
  & MathQA & 21.6 & 45.2 & $46.0_{\pm 2.26}$ & $53.0_{\pm 1.13}$ & $\textbf{53.7}_{\pm 1.27}$ & $51.1_{\pm 1.56}$ & $8.5_{pp}$ \\
\midrule
\addlinespace

\multirow{3}{*}{Phi-4-mini-3.8B-it}
  & SVAMP  & 88.2 & 89.7 & $89.9_{\pm 0.21}$ & $\textbf{90.0}_{\pm 0.42}$ & $89.4_{\pm 0.49}$ & $89.4_{\pm 0.49}$ & $0.3_{pp}$ \\
  & GSM8K  & 88.6 & 87.6 & $87.9_{\pm 1.27}$ & $87.9_{\pm 1.14}$ &  $\textbf{88.3}_{\pm 0.99}$ & ${87.9}_{\pm 0.14}$ & $0.7_{pp}$ \\
  & MathQA & 36.6 & 41.6 & $42.2_{\pm 2.83}$ & $\textbf{59.5}_{\pm 5.52}$ & $50.8_{\pm 5.66}$ & $43.1_{\pm 2.97}$ & $17.9_{pp}$ \\
\midrule
\addlinespace
\multirow{3}{*}{Gemma-3-4B-it}
  & SVAMP  & 80.7 & 75.7 & $73.5_{\pm 1.70}$ & $\textbf{81.4}_{\pm 3.75}$ & $80.4_{\pm 1.91}$ & $80.2_{\pm 2.62}$ & $5.7_{pp}$ \\
  & GSM8K  & 65.2 & 60.6 & $51.8_{\pm 0.85}$ & $\textbf{72.7}_{\pm 1.27}$ & $68.8_{\pm 2.26}$ & $62.9_{\pm 0.14}$ & $12.1_{pp}$ \\
  & MathQA & 15.8 & 16.6 & $11.8_{\pm 0.57}$ & $\textbf{21.5}_{\pm 2.12}$ & $17.3_{\pm0.99}$ & $16.3_{\pm 1.56}$ & $4.9_{pp}$ \\
\bottomrule
\end{tabular}
\end{adjustbox}
\vspace{-15pt}
\label{tab:main_results}
\end{table*}

We use the answer-span oracle to construct preference data for post-training, comparing inference-only independent sampling (\emph{Ensemble}), inference-only multi-round debate (\emph{Debate}), SFT, and three preference-based objectives~(DPO, KTO, GRPO). 
Results are averaged over two training runs with the same seed and reported with the corresponding standard deviation.
All post-training methods use the same debate traces and base model; implementation details are in Appendix~\ref{app:sec:training_details}. 
For SFT, the training target is $\mathcal G^+$, the oracle-selected completions, not gold solutions. 
For preference-based objectives, completions in $\mathcal G^+$ are preferred over those in competing clusters.

\paragraph{Overall effectiveness.}
Table~\ref{tab:main_results} shows that across all 12 model--dataset pairs, the best preference-based objective improves over inference-only debate.
DPO achieves the best post-training result on 9 pairs, KTO on 2, and GRPO on 1.
The largest gain occurs for Llama-3.2-1B-it on SVAMP, where DPO raises consensus accuracy from $19.0\%$ to $40.5\%$ ($+21.5_{pp}$); Phi-4-mini-3.8B-it on MathQA shows a comparable gain, from $41.6\%$ to $59.5\%$ ($+17.9_{pp}$). 
Gains are smallest in near-ceiling settings, such as SVAMP ($89.7\%\!\to\!90.0\%$) and GSM8K ($87.6\%\!\to\!88.3\%$) for Phi-4-mini-3.8B-it, consistent with limited room for improvement when baseline accuracy is already high.
\paragraph{Value of dispreferred traces.}
SFT and preference-based objectives use the same oracle-generated traces, but SFT imitates only the oracle-selected completions, whereas DPO, KTO, and GRPO also use completions from competing clusters. 
SFT improves over inference-only debate in 7 of 12 settings, while at least one preference-based objective improves over debate in all the settings. 
This gap suggests that incorporating competing-cluster completions provides training signal beyond the oracle-selected completions used by SFT.

The best post-training method also improves mean post-debate agent accuracy in all 12 model--dataset pairs (Appendix~\ref{app:sec:avg_agent_performance}), indicating that gains extend beyond consensus selection to individual-agent accuracy.
%
%
%
Appendix~\ref{app:sec:basline_maca_comparison} compare majority-vote labels performance on the same traces.

\section{Ablations}
\label{sec:ablations}

We assess the reliability of the framework around three questions: 
whether the gains transfer to a larger policy, 
whether the framework extends beyond numerical and multiple-choice reasoning, and 
which committee, oracle, and loss-design choices are necessary. 
Additional implementation details are in Appendix~\ref{app:sec:ablation_details}.

\paragraph{Larger model.}
Table~\ref{tab:ablate_8b} evaluates Llama-3.1-8B-it on GSM8K. 
DPO improves consensus accuracy from $82.6\%$ to $87.6\%$ ($+5.0_{pp}$), confirming that the method scales beyond the 1-4B models in Table~\ref{tab:main_results}.
The gain is smaller than the largest improvements among smaller models, consistent with the higher inference-only baseline and reduced headroom.

\paragraph{Code generation.}
We test whether the framework extends to code, where answer equivalence is determined by program-output matching on public inputs rather than a boxed numerical answer.
Llama-3.1-8B-it is evaluated on MBPP \citep{austin2021program} ($374$ training, $200$ test problems). 
DPO improves polled pass rate of all agents from $45.7\%$ to $49.1\%$ ($+3.4_{pp}$), showing that the confidence oracle generalizes to execution-based settings.
Held-out unit tests are used only for evaluation. 
Check Appendix~\ref{app:sec:mbpp-oracle} for details.

\begin{table}[t]
\centering
\small
\caption{
    Consensus accuracy~(\%) for Llama-3.1-8B-it on GSM8K and MBPP. 
    The final column is the absolute improvement of the best post-training method over inference-only debate, in percentage points.
}
\label{tab:ablate_8b}
\begin{adjustbox}{max width=0.9\textwidth}
\begin{tabular}{ll | ccc|ccc c}
\toprule
\multirow{2}{*}{Model} & \multirow{2}{*}{Dataset}
& \multicolumn{3}{c|}{Baselines}
& \multicolumn{3}{c}{Post-training}
& \multirow{2}{*}{Best $\Delta$} \\
\cmidrule(lr){3-5}\cmidrule(lr){6-8}
& & Ensemble & Debate & SFT & DPO & KTO & GRPO & \\
\midrule
\multirow{2}{*}{Llama-3.1-8B-it} & GSM8K &
  85.8 & 82.6 & 75.6 &  \textbf{87.6} & 83.6 & 83.8 & $5.0_{pp}$\\
  & MBPP &
  46.0 & 45.7 & 45.7 & \textbf{49.1} & 47.9 & 47.2 &  $3.4_{pp}$ \\
\bottomrule
\end{tabular}
\end{adjustbox}
\end{table}

\begin{wraptable}{r}{0.38\linewidth}
\centering
\small
\vspace{-12pt}
\caption{
    GSM8K consensus accuracy~(\%) after DPO by training oracle. 
    \emph{Debate} is the inference-only baseline. 
    Trace and Span use full-response and claimed answer-span confidence. 
    Gold labels serve as a supervised bound.
}
\label{tab:ablate_oracle}
\begin{adjustbox}{max width=0.35\textwidth}
\begin{tabular}{lcc}
\toprule
Oracle & Cloned & Temp.-diverse \\
\midrule
Debate    & 65.0 & $65.3_{\pm 1.30}$ \\
Gold      & 68.6 & $69.0_{\pm 0.69}$ \\
Majority  & 69.4 & $68.1_{\pm 1.5}$ \\
Trace     & 67.2 & $69.9_{\pm 0.90}$ \\
Span      & $\mathbf{72.4}$ & $\mathbf{70.4}_{\pm 1.11}$ \\
\bottomrule
\end{tabular}
\end{adjustbox}
\vspace{-30pt}
\end{wraptable}

\paragraph{Oracle selection protocol}
\label{sec:oracle_protocol}
We compare four labeling rules on GSM8K with Qwen-2.5-1.5B-it: gold labels, majority voting, full-trace confidence, and claimed answer-span confidence. 
Table~\ref{tab:ablate_oracle} reports DPO accuracy for cloned and temperature-diverse committees (3-seed averages). 
The span oracle achieves the highest post-training accuracy in both settings ($72.4\%$ and $70.4\%$).
Majority voting under-performs span-confidence selection for both committee types.
For the cloned committee, gold supervision also falls below the span oracle ($68.6\%$ vs.\ $72.4\%$), whereas for the temperature-diverse committee it remains below the span oracle ($69.0\%$ vs.\ $70.4\%$) but above majority voting ($68.1\%$).
Table~\ref{tab:oracle} compares majority voting (MV) and span-confidence selection on identical, pre-training transcripts.
Across four models on GSM8K, the two rules disagree on at most $5.8\%$ of examples, with span confidence achieving slightly higher average oracle accuracy ($57.5\%$ vs.\ $57.2\%$).
On disagreement cases, span confidence selects the correct cluster more often than MV ($31.4\%$ vs.\ $7.8\%$), with a similar advantage on count-tie cases ($29.8\%$ vs.\ $18.3\%$).
Check Appendix~\ref{app:sec:oracle_comparison} for details on more datasets.

\paragraph{Heterogeneous committee.}
The default committee uses sampling-diverse copies of one policy. 
We test a heterogeneous committee with one model from each of five families:
Llama-3.2-1B-it~(L1), Qwen-2.5-1.5B-it~(Q1), Qwen3-1.7B-it~(Q2),
Phi-4-mini-3.8B-it~(P4), and Gemma-3-4B-it~(G4), all at temperature $0.5$
with $M{=}2$ on GSM8K. 
DPO updates every member on the same span-oracle preference pairs. 
Table~\ref{tab:ablate_arch_committee} shows that heterogeneous debate reaches $79.2\%$ consensus, rising to $87.0\%$ after DPO ($+7.8_{pp}$ ), while mean agent accuracy increases by $31.4_{pp}$. 
Per-agent accuracy and more post-training approaches with heterogeneous committee are described in Appendix~\ref{app:sec:multi_model_committee}.
The gap between consensus and mean-agent accuracy ($18.8_{pp}$) is far larger than in any homogeneous committee, indicating that the shared preference partition does not benefit all members equally. 
This exposes a calibration limitation: token-probability confidence may not be comparable across architectures, so cross-model calibration remains an assumption of this experiment rather than an established property.
With per-model Platt calibration maps fit on gold correctness, heterogeneous debate consensus gives more than \(8\)pp improvement, while member accuracies stay uneven (Appendix~\ref{app:sec:platt_calibration}).

\begin{table}[t]
\centering
\small
\caption{
Comparison of majority voting (MV) and aggregate answer-span confidence before post-training on GSM8K, across four models.
\emph{Disagree} is the fraction of examples on which the two rules select different answer clusters.
The final two columns report selection accuracy on disagreement and count-tie cases, respectively.
}
\label{tab:oracle}
\begin{adjustbox}{max width=\textwidth}
\begin{tabular}{l| cc | cc | cc}
\toprule
 Model & Span & MV & Disagree
 & Disagree: Span/MV & Ties & Ties: Span/MV \\
\midrule

Llama-3.2.1B-it & 15.6	& 14.6	& 3.2	& 31.2	/ 0.0	& 5.8 &	24.1 / 6.9 \\
Qwen-2.5-1.5B-it & 66.0 & 65.0 & 5.8 & 31.0 / 13.8 & 12.8 & 31.2 / 23.4 \\
Phi-4-mini-3.8B-it & 88.8 &	88.4 &	1.2 &	33.3/ 0.0 &	1.8 & 	33.3 / 11.1 \\
Gemma-3-4B-it & 60.6 & 60.6 &	0.0 & 	0.0 / 0.0 & 0.4 & 50.0 / 50.0 \\
\midrule
Average & 57.5 &	57.2 &	2.6 &	31.4 / 7.8	& 5.2 &	29.8 / 18.3 \\
\bottomrule
\end{tabular}
\end{adjustbox}
\end{table}

\begin{table}[t]
\centering
\small
\caption{
GSM8K consensus/mean-agent accuracy (\%).
Homogeneous committee~(Hom.) uses five sampling-diverse copies; 
Heterogeneous committee~(Het.) uses one member each from L1, Q1, Q2, P4, and G4.
\(\Delta\) is the Debate\(\rightarrow\)DPO change in percentage points.
}
\label{tab:ablate_arch_committee}
\begin{adjustbox}{max width=\textwidth}
\begin{tabular}{lc|ccccc|c}
\toprule
Model & Type & Ens. & Debate & DPO & \(\Delta\) \\
\midrule
L1 & Hom. & 14.8/4.2 & 15.6/5.4 & 27.0/13.2 & +11.4/+7.8 \\
Q1 & Hom. & 44.5/24.2 & 66.0/44.5 & 71.4/60.8 & +5.4/+16.3 \\
Q2 & Hom. & 2.0/9.0 & 32.4/17.4 & 60.8/45.4  & \textbf{+28.4}/+28.0 \\
P4 & Hom. & \textbf{88.6/76.5} & \textbf{87.6/82.2} & \textbf{87.9/82.6} & +0.3/+0.4 \\
G4 & Hom. & 65.2/48.6 & 60.6/45.3 & 72.7/66.0 & +12.1/+20.7 \\
\midrule
All & Het. & 75.0/31.5 & 79.2/36.8 & {87.0}/68.2 & +7.8/\textbf{+31.4} \\
\bottomrule
\end{tabular}
\end{adjustbox}
\vspace{-10pt}
\end{table}

\paragraph{Agreement weighting.}
\begin{wraptable}{r}{0.32\linewidth}
\centering
\small
\vspace{-12pt}
\caption{
    GSM8K consensus accuracy~(\%) for Qwen-2.5-1.5B-it after DPO with different uses of winner support. 
    Means $\pm$ std over three seeds.
}
\label{tab:ablate_weights}
\begin{adjustbox}{max width=0.3\textwidth}
\begin{tabular}{lc}
\toprule
Method & Accuracy \\
\midrule
Debate               & $67.1_{\pm 1.56}$ \\
Unweighted DPO       & $\mathbf{70.4}_{\pm 1.11}$ \\
Weighted DPO         & $70.0_{\pm 1.99}$ \\
Agreement-filtered   & $68.5_{\pm 3.25}$ \\
\bottomrule
\end{tabular}
\end{adjustbox}
\vspace{-20pt}
\end{wraptable}

We compare three DPO variants with the span oracle fixed: unweighted DPO ($\alpha(x)~=~1$), support-weighted DPO (per-example loss scaled by $\alpha(x)=|\mathcal G^+(x)|/N$), and agreement-filtered DPO (retain only examples with $\geq 3$ of $5$ agents supporting the preferred cluster).
Table~\ref{tab:ablate_weights} shows that weighting does not improve final accuracy: unweighted DPO reaches $70.4\%$, compared with $70.0\%$ for support-weighted and $68.5\%$ for agreement-filtered. 
We therefore recommend to retain unweighted DPO as the default and treat $\alpha(x)$ as a diagnostic quantity, not as part of the training objective.

\section{Conclusion}
\label{sec:conclusion}

\textbf{Alignment by Confident Debate} (\method{}) constructs preference data from multi-agent debate traces.
The framework clusters final answers by task-level equivalence and ranks the clusters using aggregate confidence on their claimed answer, converting debate disagreement into supervision compatible with standard post-training objectives.
Across mathematical reasoning, question-answering, and coding benchmarks, our oracle selection identifies correct answer cluster more accurately than majority voting overall, with a larger advantage on disagreement and count-tie instances.
Preference training on the resulting pairs improves over inference-only ensembling, inference-only debate, and SFT trained on the same traces, with the largest gains occurring when the inference-only committee has substantial headroom.
%
%
However, heterogeneous committees expose a limitation of aggregating raw token-probability confidence across models, motivating future work on cross-model confidence calibration.
These results indicate that combining task-level answer equivalence with confidence-weighted disagreement can turn multi-agent debate into a practical self-training signal, provided that outputs are parseable, a suitable equivalence procedure is available, and confidence is comparable across the committee.

\section*{AI use statement}

In this work, we used generative AI tools as research and writing assistants to provide feedback on the presentation and interpretation of experimental results, check consistency between reported numerical results and tables, and improve the clarity, grammar, and organization of the manuscript. 
We also used generative AI tools for limited assistance with LaTeX formatting and presentation of tables and figures.

Generative AI tools were not used to generate datasets, run experiments, produce experimental results, or implement the proposed method or training pipeline. 
The research methodology, experimental design, analysis, and scientific conclusions were determined by the authors, with AI assistance used only as a supporting tool.

All AI-assisted content was reviewed and verified by the authors. 
In particular, numerical values, statistical results, experimental comparisons, mathematical formulations, and claims were checked against the underlying experimental outputs and implementation. 
The authors made the final decisions regarding the methodology, interpretation, and presentation of the work and take full responsibility for the final manuscript and all claims and artifacts contained
therein.

\bibliography{iclr2027_conference}
\bibliographystyle{iclr2027_conference}

\newpage
\appendix

\section{Literature review}
\label{app:sec:lit_review}
\paragraph{Multi-agent debate at inference.}
Self-consistency aggregates independent chains by majority vote when answers can be checked for equivalence~\citep{wang2023selfconsistency}. 
Universal self-consistency relaxes that requirement by asking a model which free-form samples are consistent~\citep{chen2024usc}. 
Multi-agent debate instead lets agents read one another across multiple rounds, and the group's answer can be stronger than any single chain~\citep{du2023debate,liang2023mad}. 
That gain is not automatic. 
Under homogeneous agents and unweighted belief updates, debate preserves expected correctness over time, so
the protocol can underperform a vote over independent samples~\citep{choi2026debate,zhu2026demystifying}. 
Over rounds, repeated interaction can increase similarity among agents' rationales, a failure mode termed \emph{representational collapse}, so agreement can manufacture confidence while errors stay shared~\citep{dai2026consensus}. 
The remaining uncertainty mixes epistemic disagreement with aleatoric decoding noise, and longer debates can let the latter flip a correct answer~\citep{qiao2026epistemic}. 
These results diagnose test-time debate.
\method{} takes it as a constraint on the \emph{training} oracle: consensus is an outcome of the protocol, so it cannot be the label.

\paragraph{Confidence-weighted selection.}
A natural alternative to vote count is the model's own likelihood.
Full-response likelihood is a coarse signal because it scores all generated tokens, including reasoning tokens that may be only weakly tied to final-answer correctness. 
CISC weights independent samples by verbalized confidence, $P(\mathrm{true})$, or sequence probability~\citep{taubenfeld2025cisc};
self-certainty uses the peakedness of the token distribution~\citep{kang2025selfcertainty}.
Restricting the score to answer-relevant tokens such as numbers, entities, or choice letters ranks correctness better than scoring every token~\citep{razghandi2025cer}. 
In debate, length-normalized answer-token confidence, after calibration, can improve the committee's answer selection~\citep{lin2025confmad}, and diversity-aware initialization combined with calibrated confidence communication can allow homogeneous debate to outperform majority vote~\citep{zhu2026demystifying}.
On open-ended debate without a reference answer, early-generation log-probabilities vary more across traces than the rest of the sequence and better predict judged quality~\citep{keramati2026early}.
Token probabilities are also a well-studied uncertainty signal more broadly~\citep{kadavath2022know,tian2023verbalized}.
All of this work uses confidence to \emph{select} an answer at inference time.
\method{} uses span-level confidence to \emph{construct preferred and dispreferred traces} for post-training.

\paragraph{Self-training without gold labels.}
STaR and ReST fine-tune on traces that pass a verifier~\citep{zelikman2022star,gulcehre2023rest}. 
TTRL and related methods treat the majority among independent samples as a pseudo-label for GRPO or DPO~\citep{zhang2025ttrl,shao2024deepseekmath}. 
Self-rewarding models and RLAIF replace the verifier with another language model~\citep{yuan2024selfrewarding,bai2022constitutional}.
Process reward models need step-level annotation~\citep{lightman2023prm}.
These methods still rely on an external correctness signal from a checker, a judge, or an independent majority. 
\method{} stays in this family but derives its signal from a debating committee and from span-level likelihood, not from
a gold answer, a judge, or vote count alone.

\paragraph{Debate as post-training data.}
Multi-agent interaction is increasingly used as a training corpus, not only as a test-time ensemble. 
One line of work imitates debate roles or consensus traces using SFT or GRPO~\citep{subramaniam2025multiagent,srivastava2025dte}.
Another turns majority versus minority traces, or debate-derived preference trees, into preference pairs~\citep{samanta2026maca,zhou2025debate}.
A third jointly trains specialized agents in generate--verify--refine loops, student--teacher setups, or cooperative multi-agent reinforcement learning~(MARL), typically against gold labels, a teacher model, or a shared task
reward~\citep{motwani2025malt,liu2026magrpo}. 
AI-safety debate is a different setting: two agents argue for a human judge~\citep{irving2018debate}. 
What these methods share is how they \emph{name} and \emph{weight} the data. 
The winner is determined by vote count, a gold answer, or a teacher; every retained pair is treated equally; and the committee is typically composed of homogeneous clones, with diversity obtained only by training distinct roles rather than by scoring a diverse sample of a single policy. 
To our knowledge, prior debate-training methods do not construct preference pairs by ranking task-level answer-equivalence clusters using aggregate claimed answer-span confidence, or treat the degree of disagreement within the committee as a reliability weight rather than as the label. 
Unlike prior debate-based training methods that use vote count, a teacher, or a gold answer to assign labels, \method{} constructs preference data by combining task-level answer equivalence, span-confidence mass, and disagreement-based reliability weighting.

\section{Algorithm}
\label{app:sec:algo}

Algorithm~\ref{app:alg:method} outlines \method{}. 
For each prompt, a diverse committee debates for $M$ rounds; 
final answers are clustered by equivalence and ranked by confidence mass on the claimed answer-span. 
The winning cluster yields preferred completions, the rest dispreferred, and the agreement rate can be used as a soft reliability weight. 
We collect SFT and preference datasets from these traces and post-train the base policy with standard objectives.

\begin{algorithm}[h]
\caption{\method{}: Alignment by Confident Debate}
\label{app:alg:method}
\begin{algorithmic}[1]
\Require Training prompts $\mathcal X_{\mathrm{train}}$, base policy $\pi_\theta$,
parser $\mathcal A$, span extractor $S$, equivalence relation $\equiv$,
committee size $N$, and number of debate rounds $M$

\State $\mathcal D_{\mathrm{SFT}}\gets\emptyset$
\State $\mathcal D_{\mathrm{pref}}\gets\emptyset$
\State $\mathcal D_{\mathrm{oracle}}\gets\emptyset$

\For{each $x\in\mathcal X_{\mathrm{train}}$}
    \State Instantiate sampling-diverse agents
    $\{\pi_i\}_{i=1}^{N}$ from $\pi_\theta$
    \State $y_i^{(1)}\sim\pi_i(\cdot\mid x)$ for $i=1,\ldots,N$

    \For{$m=2,\ldots,M$}
        \State $y_i^{(m)}
        \sim
        \pi_i\!\left(
        \cdot\mid x,\operatorname{ser}_i
        \!\left(\{y_j^{(m-1)}\}_{j\ne i}\right)
        \right)$
        for all $i$
    \EndFor

    \State $\mathcal Y(x)\gets\{y_i^{(M)}\}_{i=1}^{N}$
    \State $a_i\gets\mathcal A(y_i)$ for $i=1,\ldots,N$
    \State Compute $c_i^{\mathrm{answer}}$ for each parseable trace
    \State Cluster parseable answers by $\equiv$ into $\{\mathcal I_k\}$

    \If{no parseable cluster exists}
        \State \textbf{continue}
    \EndIf

    \State $w_k\gets\sum_{i\in\mathcal I_k}c_i^{\mathrm{answer}}$
    and $n_k\gets|\mathcal I_k|$
    \State $k^\star\gets\operatorname*{arg\,max}_{k}(w_k,n_k)$
    \State $\hat a(x)\gets\operatorname{rep}(k^\star)$
    \State $\mathcal G^+(x)\gets
    \{y_i:a_i\equiv\hat a(x)\}$
    \State $\mathcal G^-(x)\gets
    \{y_i:a_i\not\equiv\hat a(x)\}$,
    including unparseable traces
    \State $\alpha(x)\gets|\mathcal G^+(x)|/N$

    \State $\mathcal D_{\mathrm{oracle}}\gets
    \mathcal D_{\mathrm{oracle}}\cup
    \{(x,\hat a(x),\alpha(x))\}$
    \State $\mathcal D_{\mathrm{SFT}}\gets
    \mathcal D_{\mathrm{SFT}}\cup
    \{(x,\mathcal G^+(x))\}$

    \If{$\mathcal G^+(x)\neq\emptyset$ and
        $\mathcal G^-(x)\neq\emptyset$}
        \State $\mathcal D_{\mathrm{pref}}\gets
        \mathcal D_{\mathrm{pref}}\cup
        \{(x,\mathcal G^+(x),\mathcal G^-(x),\alpha(x))\}$
    \EndIf
\EndFor

\State Post-train $\pi_\theta$ using SFT, DPO, KTO, or GRPO
on the corresponding dataset
\end{algorithmic}
\end{algorithm}

\section{Datasets}
\label{app:sec:data}

We evaluate \method{} on four publicly available benchmarks spanning mathematical word problems, multiple-choice quantitative question answering, and Python code generation.
GSM8K, SVAMP, and MathQA evaluate mathematical or quantitative reasoning; MBPP evaluates code generation with executable unit tests.
Gold answers and full unit-test outcomes are used only for evaluation and reporting accuracy; they are not provided to the debate oracle or used as training labels.

\paragraph{GSM8K.}
GSM8K contains 8,500 grade-school mathematical word problems with natural-language solutions and numeric final answers~\citep{cobbe2021gsm8k}.
It evaluates multi-step arithmetic reasoning in natural language.
For \method{}, we parse and cluster the boxed numerical final answers produced by committee members.

\paragraph{SVAMP.}
SVAMP contains arithmetic word problems designed to test robustness to variations in linguistic form and problem structure~\citep{patel2021svamp}.
Its controlled perturbations reduce the usefulness of superficial shortcuts and emphasize the underlying arithmetic relations.
As in GSM8K, committee members produce boxed numerical final answers.

\paragraph{MathQA.}
MathQA is a large-scale quantitative reasoning benchmark containing natural-language questions paired with executable mathematical programs and multiple-choice answer options~\citep{amini2019mathqa}.
We use it as a multiple-choice quantitative QA setting, where agents produce a final answer in the prescribed answer format and the oracle clusters answers by selected option equivalence.

\paragraph{MBPP.}
%
MBPP contains approximately 1,000 crowd-sourced Python problems intended to be solvable by entry-level programmers~\citep{austin2021program}.
Each example provides a natural-language specification, a reference function, a public test list, and an optional challenge list.
A public test is an assertion of the form \texttt{assert} $f(x)=y$.
We parse each assertion into the arguments $x$ and the expected return value $y$, execute the candidate agent response on $x$, and mark the case passed when the returned value matches $y$.
Programs that emit the same sequence of public outputs, including the same exception type on a failing case, form one equivalence class.
The execution oracle ranks a class by how many of these public cases its members pass.
Challenge assertions are omitted from this trace, so they do not affect clustering or the training label.
%

\paragraph{RaR.}
The main paper we use tasks on which a final answer is a number, a choice letter, or a program, so two traces can be tested for equivalence and then counted.
RaR-Medicine~\citep{gunjal2026rubrics} does not have that kind of answer.
We use the medicine split of Rubrics as Rewards \citep{gunjal2026rubrics}, released as \texttt{ScaleAI/RaR-Medicine} on Huggingface.
Each example is a clinical question, a reference answer, and a checklist.
A checklist is an ordered list of pass/fail criteria produced by the autorubric pipeline of \citet{gunjal2026rubrics}.
We used this dataset in appendix~\ref{app:sec:non_equivalence_tasks}.

\paragraph{Task-level equivalence.}
The relation \(\equiv\) in Section~\ref{sec:oracle} is computed entirely from the committee's own outputs: it compares parsed candidate answers with one another and never receives the reference answer.
A trace for which the parser \(\mathcal{A}(y)\) is undefined is excluded from clustering.
On GSM8K and SVAMP, \(y \equiv y'\) when the numbers parsed from the final boxed spans denote the same value, so two derivations of the same quantity form one cluster regardless of wording.
On MathQA, \(y \equiv y'\) when those spans name the same choice letter.
On MBPP, \(y \equiv y'\) when both completions parse as programs, execute on the public tests, and return the same outputs; the public tests are not part of the generation prompt, and the hidden challenge tests are used only to score the selected program at evaluation.
In all four tasks, \(\equiv\) is a relation on task outcomes, not on the text of the reasoning trace.
But in RaR we used a Judge model to find the equivalence of free text generated by agents.

\paragraph{Training label formation.}
The training label is a cluster of agent answers.
The rule that decides which answers belong together, and the score that chooses the winning cluster, depend on the dataset.
GSM8K and SVAMP are grade-school word problems with one numerical answer.
MathQA is also a math problem, but the answer is a choice letter.
On these three datasets every agent produces a short discrete answer, so confidence ensembling is an alternative to a majority vote.
The same construction extends to domains without a single gold string.
MBPP is execution.
Programs that produce the same public outputs form one cluster, a vote is well-defined after that execution, and confidence is the number of public tests passed.
RaR is free text.
A judge marks the checklist, including penalty items with negative weight, and confidence is the weighted score of the marked items.
Voting remains a baseline after the same clustering: on code it counts programs with the same public outputs, and on rubrics it counts copies of the same checklist.
Table~\ref{tab:label-formation} lists the three groups.

\begin{table}[t]
\centering
\caption{How agent answers are clustered and scored for the training label.}
\label{tab:label-formation}
\small
\begin{tabularx}{\linewidth}{@{}ll>{\raggedright\arraybackslash}X>{\raggedright\arraybackslash}X@{}}
\toprule
Group & Datasets & What is aggregated & Confidence \\
\midrule
Discrete math & GSM8K, SVAMP, MathQA & The same extracted answer & Span confidence \\
Executable code & MBPP & Programs with the same public outputs & Public tests passed \\
Checklist text & RaR & Identical satisfied-criteria patterns & Weighted hit score, penalties subtracted \\
\bottomrule
\end{tabularx}
\end{table}

\section{Prompt Templates}
\label{app:sec:prompts}

All agents in a committee share the same base prompt for a given domain, with minor task-specific variants for math, multiple-choice QA, and coding.
For math and QA tasks, the initial prompt instructs the model to provide a bullet-point summary of its step-by-step reasoning and to emit a final answer in the form \boxed{answer}. 
For coding (MBPP), the prompt requests a single Python function in a fenced code block with no additional text. 
In each debate round, agents receive the other agents’ previous responses and are asked to update their reasoning or code accordingly, while preserving the same final-answer format.

\paragraph{Math (GSM8K, SVAMP)}
\noindent
\begin{tcolorbox}[
    breakable,
    colback=gray!5!white,
    colframe=black,
    title=Initial Prompt for Math problems]
\begin{lstlisting}[basicstyle=\ttfamily,breaklines=true]
Solve the following math problem. {question}
Provide a bullet point summary of your step-by-step reasoning.
Your final answer should be a single numerical number, in the form \boxed{answer}, at the end of your response.
\end{lstlisting}
\end{tcolorbox}

\paragraph{Multiple-choice QA (MathQA)}
\noindent
\begin{tcolorbox}[
    breakable,
    colback=gray!5!white,
    colframe=black,
    title=Initial Prompt for Multiple-Choice QA]
\begin{lstlisting}[basicstyle=\ttfamily,breaklines=true]
Answer the following multiple choice question as accurately as possible. {question}
Provide a bullet point summary of your step-by-step reasoning.
Your final answer should be a single choice letter in the form \boxed{answer}, at the end of your response.
\end{lstlisting}
\end{tcolorbox}

\paragraph{For Coding~(MBPP)}
\noindent
\begin{tcolorbox}[
    breakable,
    colback=gray!5!white,
    colframe=black,
    title=Initial Prompt for Coding problems]
\begin{lstlisting}[basicstyle=\ttfamily,breaklines=true]
Write Python for this coding problem. Output code only.
{question}
Reply with ONLY a ```python fenced block. No other text. 
Inside it, write the complete function. No tests, no print(), no extra drivers.
\end{lstlisting}
\end{tcolorbox}

\paragraph{Debate exchange rounds (all domains).}
In each debate round, agents receive the other agents’ previous responses and are asked to update their solution while preserving the domain-specific final-answer format. 
The exchange prompt is:
\noindent
\begin{tcolorbox}[
    breakable,
    colback=gray!5!white,
    colframe=black,
    title=Debate Exchange round prompt]
\begin{lstlisting}[basicstyle=\ttfamily,breaklines=true]
Here are solutions from other agents: {other_agent_solutions}

[Using each response as additional advice, give an updated bullet point summary of your step-by-step reasoning to the question: {question}]
OR
[Using each response as additional advice, update your Python solution to the question: {question}]

[Make sure your final answer is in the form \boxed{answer}, at the end of your response.]
OR
[Make sure your final answer is a single choice letter in the form \boxed{answer}, at the end of your response.]
OR
[Reply with ONLY a ```python fenced block. No tests, no print(), no extra drivers.]
\end{lstlisting}
\end{tcolorbox}

\section{Committee Configuration Details}
\label{app:sec:committee_conf_details}

\begin{figure*}[ht]
    \centering
    \includegraphics[width=0.3\linewidth]{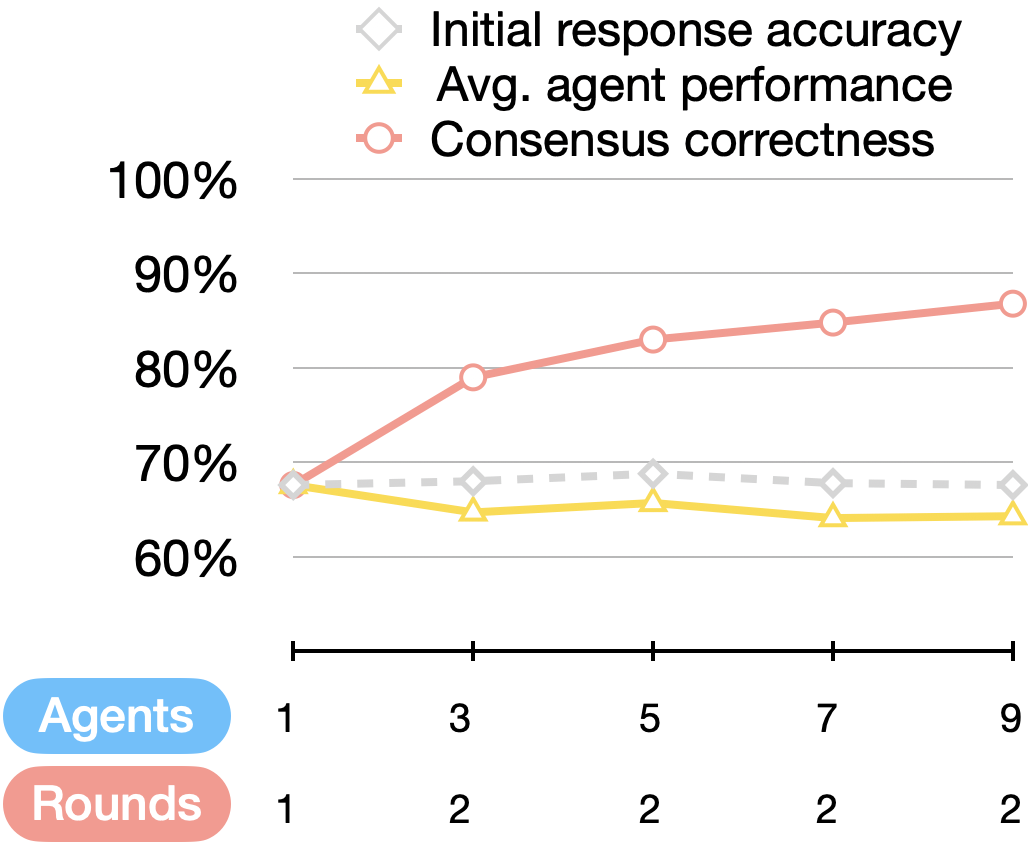}
    \caption{
        Committee behavior for Llama-3.1-8B-it model on GSM8K. 
        The consensus accuracy by committee size beyond 5 agents shows diminishing returns.
    }
    \label{app:fig:num_agents_llama8b}
\end{figure*}

\begin{figure*}[t]
    \centering

    \begin{subfigure}[t]{0.48\textwidth}
        \centering
        \includegraphics[width=\linewidth]{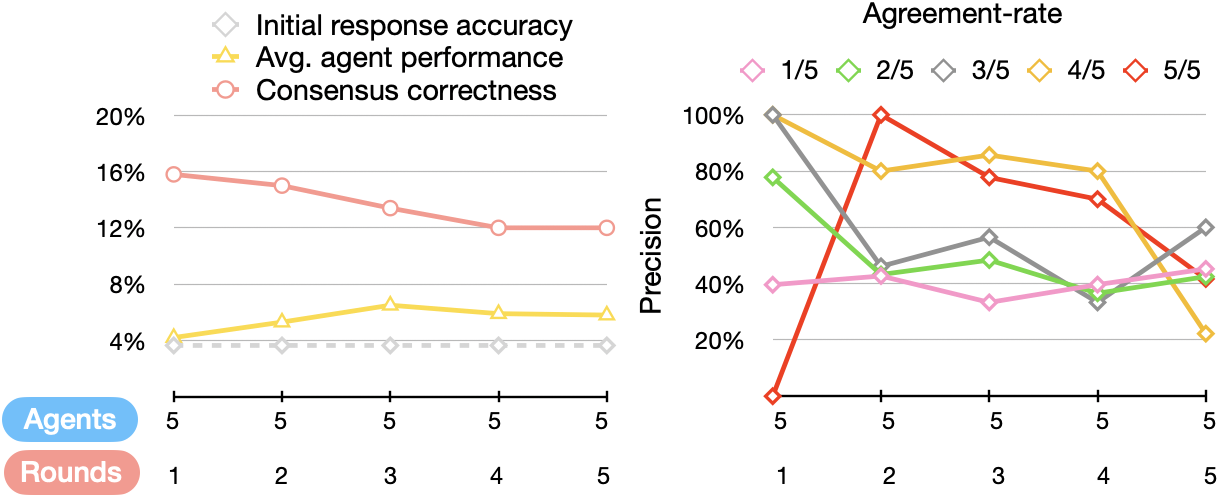}
        \caption{Llama-3.2-1B-it}
        \label{fig:rounds_llama1b}
    \end{subfigure}
    \hfill
    \begin{subfigure}[t]{0.48\textwidth}
        \centering
        \includegraphics[width=\linewidth]{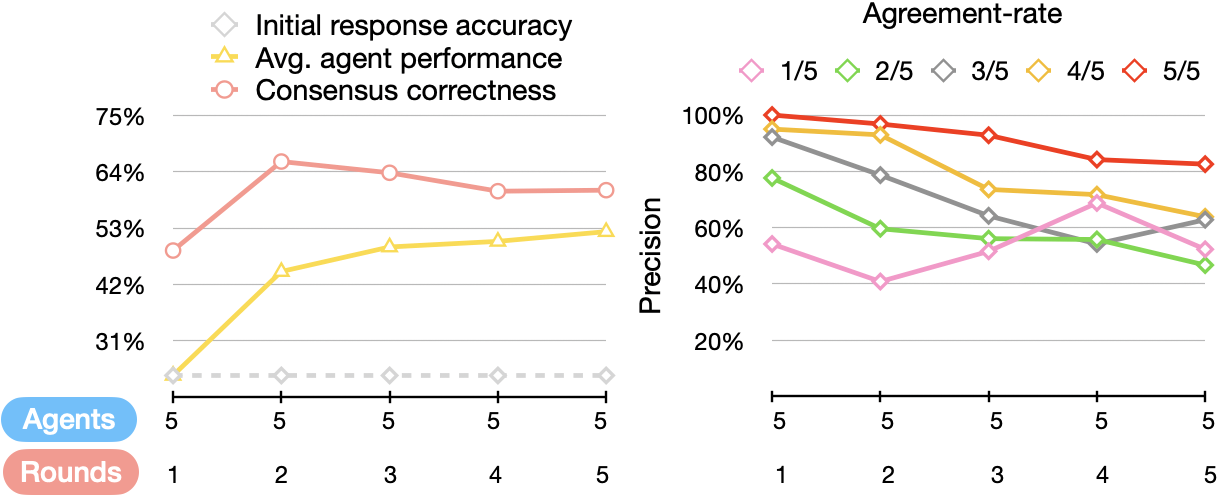}
        \caption{Qwen2.5-1.5B-it}
        \label{fig:rounds_qwen}
    \end{subfigure}
    \hfill
    \begin{subfigure}[t]{0.48\textwidth}
        \centering
        \includegraphics[width=\linewidth]{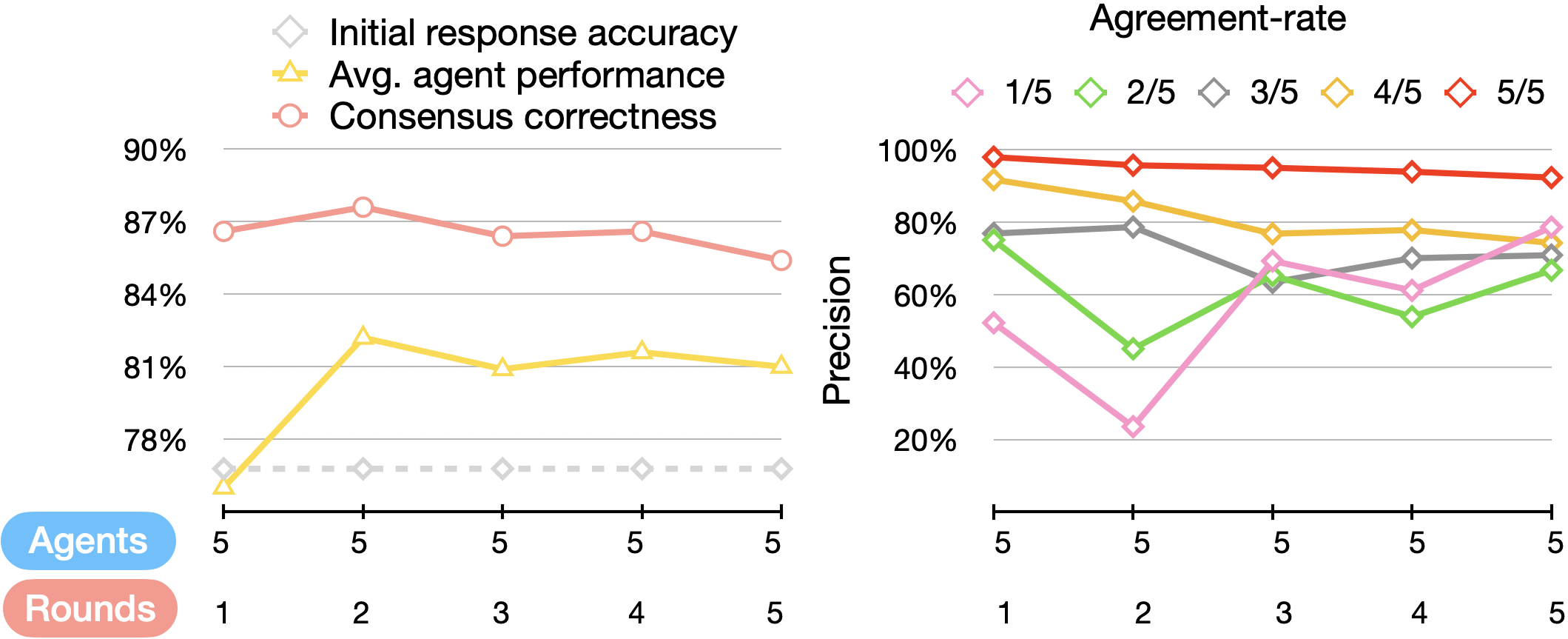}
        \caption{Phi-4-mini-3.8B-it}
        \label{fig:rounds_phi4b}
    \end{subfigure}

    \vspace{0.5em}

    \begin{subfigure}[t]{0.48\textwidth}
        \centering
        \includegraphics[width=\linewidth]{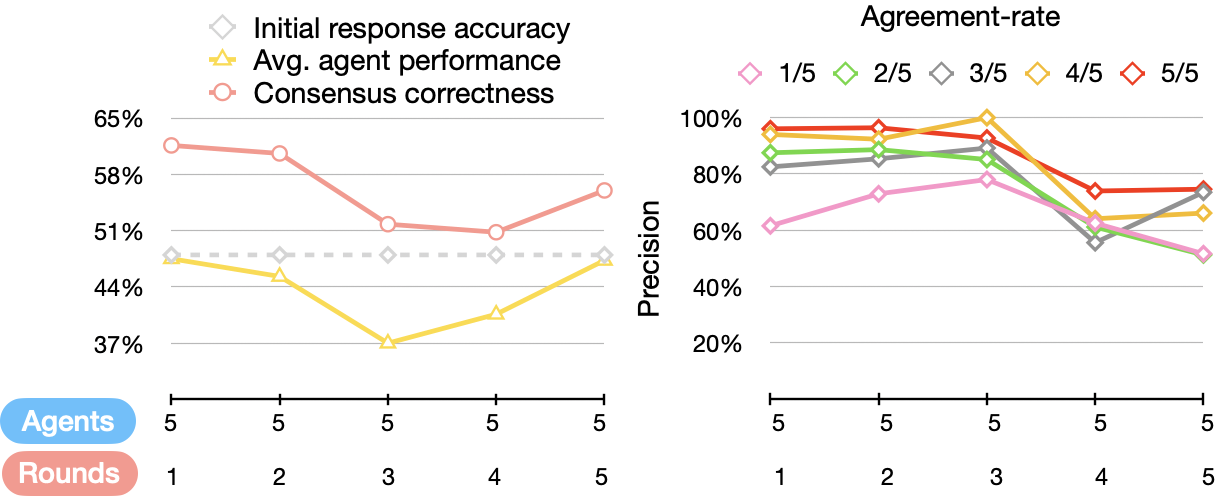}
        \caption{Gemma-3-4B-it}
        \label{fig:rounds_gemma4b}
    \end{subfigure}
    \hspace{0.02\textwidth}
    \begin{subfigure}[t]{0.48\textwidth}
        \centering
        \includegraphics[width=\linewidth]{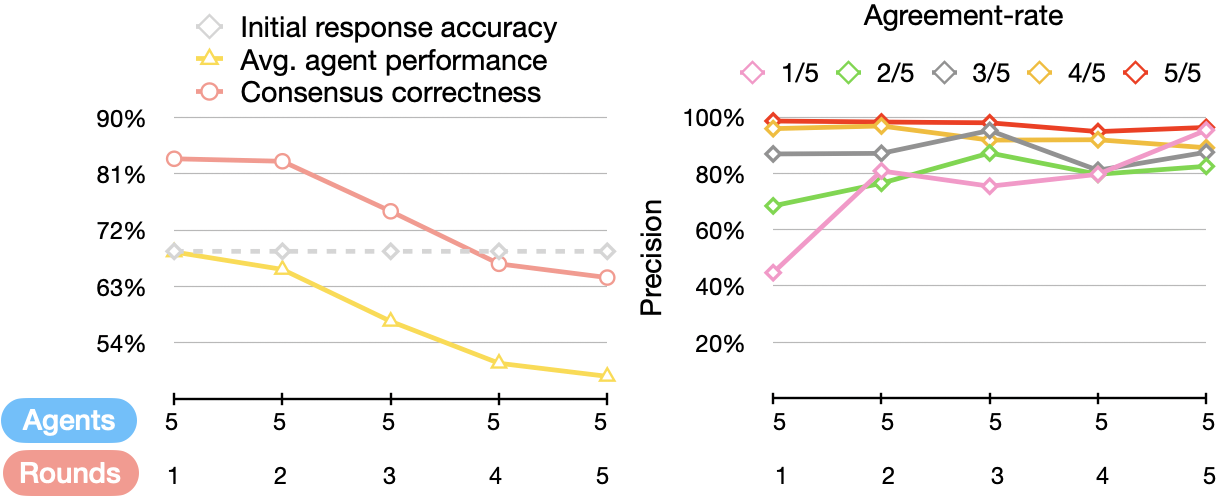}
        \caption{Llama-3.1-8B-it}
        \label{fig:rounds_llama8b}
    \end{subfigure}

    \caption{
        Effect of additional debate rounds on GSM8K across model families. 
        \textbf{Left:} Consensus accuracy at a fixed committee size as the number of exchange rounds increases.
        \textbf{Right:} Oracle precision stratified by agreement rate, where $k/5$ denotes that $k$ of five agents produce parseable answers in the selected answer-equivalence cluster. %
        The first exchange round often improves consensus accuracy, while subsequent rounds can increase agreement without improving correctness and can reduce oracle precision at several agreement levels. 
        Reference answers are used only for evaluation.
    }
    \label{app:fig:rounds_across_models}
\end{figure*}

Unless otherwise noted, all agents decode with top-$p$ $=0.9$ and generate at most $256$ tokens per completion.
Figure~\ref{fig:committee_agents_qwen2b} reports GSM8K accuracy as a function of committee size $N$ and number of debate rounds $M$ for cloned Qwen-2.5-1.5B-it committees at a shared temperature of $0.5$.
We observe the same qualitative trends for Llama-3.1-8B-it across committee sizes and debate rounds~(Figure~\ref{app:fig:num_agents_llama8b} and \ref{fig:rounds_llama8b}).

\begin{figure*}[h]
    \centering

    \begin{subfigure}[t]{0.48\textwidth}
        \centering
        \includegraphics[width=\linewidth]{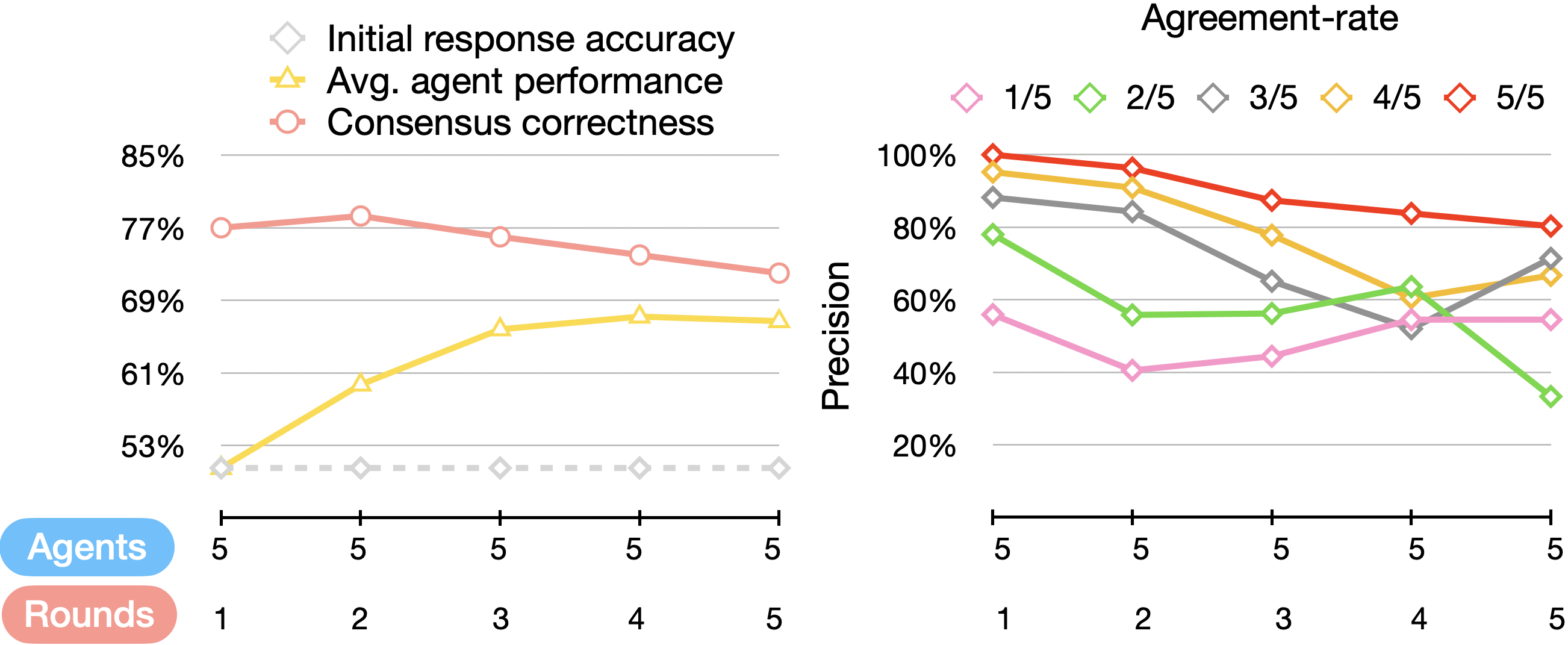}
        \caption{SVAMP}
        \label{fig:rounds_qwen_svamp}
    \end{subfigure}
    \hfill
    \begin{subfigure}[t]{0.48\textwidth}
        \centering
        \includegraphics[width=\linewidth]{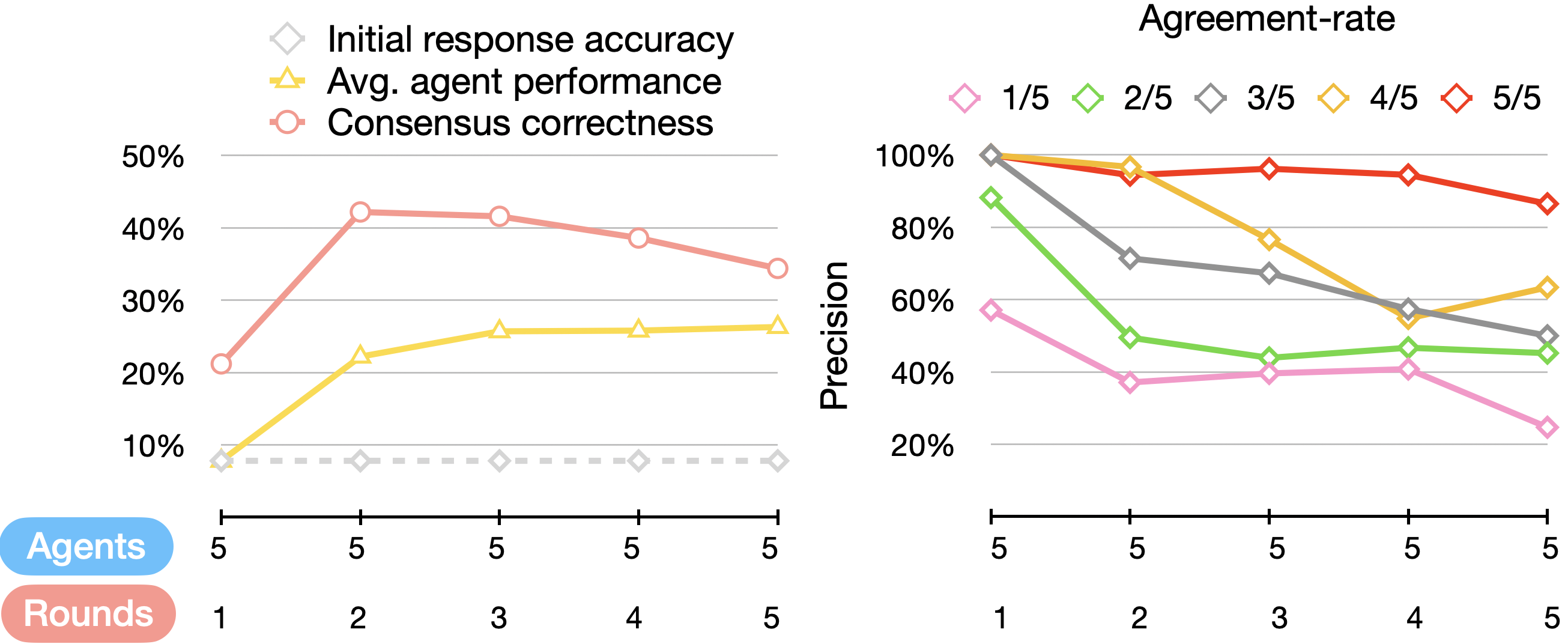}
        \caption{MathQA}
        \label{fig:rounds_qwen_mathqa}
    \end{subfigure}
    
    \caption{
        Effect of additional debate rounds on Qwen-2.5-1.5B-it across datasets. 
        \textbf{Left:} Consensus accuracy at a fixed committee size as the number of exchange rounds increases. 
        \textbf{Right:} Oracle precision stratified by agreement rate, where $k/5$ denotes that $k$ of five agents produce parseable answers in the selected answer-equivalence cluster. %
        Reference answers are used only for evaluation.
    }
    \label{app:fig:rounds_across_datasets}
\end{figure*}

We additionally fix the committee size~($5$) and compare the effect of additional exchange rounds across model families on GSM8K~(Figure~\ref{app:fig:rounds_across_models}).
We observe similar behavior across other datasets~(Figure~\ref{app:fig:rounds_across_datasets}).
Across these settings, the first exchange round often improves consensus accuracy, whereas further rounds increase agreement without a corresponding improvement in correctness and can subsequently reduce consensus accuracy.
This pattern is consistent with debate-induced correlation: later rounds make agents{'} answers more alike without necessarily adding independent evidence for the correct answer.

We further stratify oracle precision by agreement-rate.
For a five-agent committee, an agreement rate of $2/5$ means that two agents produce parseable final answers in the same answer-equivalence cluster; the remaining agents either support other clusters or produce unparseable answers.
Precision is the fraction of oracle-selected answers that are correct within each agreement-rate bin, using reference answers only for evaluation.
Across the evaluated models, later exchange rounds tend to reduce precision at several agreement levels, including partially split committees, despite increasing the frequency of high-agreement outcomes.

To isolate the effect of committee diversity, we compare four configurations: 
(i) \emph{clones}, which share temperature $0.01$; 
(ii) \emph{temperature-diverse agents}, which use temperatures $\{0.01, 0.25, 0.5, 0.75, 1.0\}$ across five agents; 
(iii)  \emph{role-diverse agents}, which use distinct system prompts at the same temperature $0.01$ and 
(iv) \emph{seed-diverse agents}, which use distinct sampling seeds at a shared temperature $0.01$.
Table~\ref{app:tab:committee_diversity} reports the corresponding inference-only results.
Across the evaluated models, the temperature-diverse committee achieves either the highest or second-highest consensus accuracy.
We therefore use it as the default committee configuration for the experiments in Section~\ref{sec:experiment_results}.

\paragraph{Role prompt templates.}
These system prompts are used for the role-diverse committee for committee configuration~(Section~\ref{sec:committee}).

\noindent
\begin{tcolorbox}[
    breakable,
    colback=gray!5!white,
    colframe=black,
    title=System Prompts for Role-Diverse Committee]
\begin{lstlisting}[basicstyle=\ttfamily,breaklines=true]
mechanism: You are the MECHANISM reasoner. Derive the answer from first principles. Identify governing quantities, write governing relations, and chain causal steps before computing a final result.
        
pattern: You are the PATTERN reasoner. Match this problem to prototypical cases. You recognize (standard templates, common problem types) and adapt the known solution pattern to this instance.

arbiter: You are the ARBITER. Weigh competing arguments fairly and synthesize the most defensible conclusion.

verifier: You are the VERIFIER. Check peer reasoning for logical and arithmetic errors; correct mistakes explicitly.

solver: You are the SOLVER. Produce a clear step-by-step solution.
\end{lstlisting}
\end{tcolorbox}

\begin{table}[h]
\centering
\small
\caption{Inference-only GSM8K accuracy (\%) at \(N{=}5\), \(M{=}2\). 
Initial response accuracy is single-sample accuracy before any ensemble or debate; 
Avg. agent performance is average per-agent accuracy after \(M\) rounds of debate; 
Consensus correctness is aggregated response accuracy.}
\label{app:tab:committee_diversity}
\begin{tabular}{l | ccc}
\toprule
Committee & Initial response accuracy & Avg. agent performance & Consensus correctness \\

\midrule
& \multicolumn{3}{c}{Llama-3.2-1B-it} \\
\midrule
Clones                   & 3.2 & 4.6 & 6.4 \\
Temperature-diverse      & 4.4 & \textbf{7.4}  & 17.0 \\
Role-diverse             & \textbf{5.6} & 7.3 & \textbf{20.6} \\
Seed-diverse             & 3.2 & 4.3 & 6.6 \\

\midrule
& \multicolumn{3}{c}{Qwen-2.5-1.5B-it} \\
\midrule
Clones                   & \textbf{24.6} & \textbf{48.7} & 56.4 \\ 
Temperature-diverse      & 23.2 & 44.5 & \textbf{66.0} \\
Role-diverse             & 21.0 & 44.6 & 63.6 \\
Seed-diverse             & 23.8 & 47.3 & 56.6\\

\midrule
& \multicolumn{3}{c}{Phi-4-mini-3.8B-it} \\
\midrule
Clones                   & \textbf{77.2} & \textbf{81.3} & 84.8 \\
Temperature-diverse      & 76.7 & 80.9 & \textbf{87.6} \\
Role-diverse             & 65.8 & 76.8 & 86.6 \\
Seed-diverse             & \textbf{77.2} & 80.0 & 83.0 \\

\midrule
& \multicolumn{3}{c}{Gemma-3-4B-it} \\
\midrule
Clones                   & 47.8 & \textbf{44.6} & 51.4 \\
Temperature-diverse      & \textbf{48.1} & 42.6 & \textbf{57.8} \\
Role-diverse             & 14.3 & 4.6 & 18.2 \\
Seed-diverse             & 47.7 & \textbf{44.6} & 51.8 \\

\midrule
& \multicolumn{3}{c}{Llama-3.1-8B-it} \\
\midrule
Clones                   & 68.6 & 66.4 & 74.0 \\
Temperature-diverse      & 66.0 & 61.6 & \textbf{81.0} \\
Role-diverse             & 40.9 & 27.2 & 65.4 \\
Seed-diverse             & \textbf{69.3} & \textbf{67.0} & 75.2 \\

\bottomrule
\end{tabular}
\end{table}

\section{Confidence Comparison}
\label{app:sec:conf_comparison}

This section provides the full numerical results underlying Figure~\ref{fig:trace_vs_span_auroc}, which compares full-trace confidence with claimed answer-span confidence as predictors of correctness.
For each agent and each sample across datasets, we compute AUROC using that agent{'}s own confidence scores against its correctness on the gold label (evaluation only; gold labels are not used in training).
In the table, ``CoT'' denotes full-trace confidence $c^{\mathrm{trace}}$, and ``Box'' denotes claimed answer-span confidence $c^{\mathrm{span}}$.
R1 and R2 refer to before and after one round of exchange, respectively, and $\Delta = \mathrm{R2} - \mathrm{R1}$.

Table~\ref{app:tab:trace_vs_span_confidence} reports per-model, per-dataset AUROC for four model families across GSM8K, SVAMP, and MathQA.
%
Across most settings, span confidence either starts higher or improves after debate, whereas full-trace confidence generally degrades.
The average $\Delta$ is negative for CoT ($-0.07$) and positive for Box ($+0.04$), consistent with our claim that answer-span confidence provides a more stable signal for ranking correctness in multi-agent debate.

\begin{table*}[h]
\centering
\caption{Full-trace (CoT) versus claimed-answer-span (Box) AUROC before and after one debate round. \(\Delta=\mathrm{R2}-\mathrm{R1}\).}
\label{app:tab:trace_vs_span_confidence}
\resizebox{\textwidth}{!}{%
\begin{tabular}{llrrrrrr}
\toprule
\textbf{Model} & \textbf{Dataset}
& \textbf{CoT R1} & \textbf{CoT R2} & \(\boldsymbol{\Delta}\) \textbf{CoT}
& \textbf{Box R1} & \textbf{Box R2} & \(\boldsymbol{\Delta}\) \textbf{Box} \\
\midrule
\multirow{3}{*}{Llama-3.2-1B-it} & GSM8K & 0.60 & 0.50 & -0.10 & 0.63 & 0.72 &  0.10 \\
 & SVAMP & 0.55 & 0.56 &  0.01 & 0.59 & 0.77 &  0.18 \\
& MathQA & 0.64 & 0.54 & -0.10 & 0.63 & 0.65 &  0.02 \\
\midrule
\multirow{3}{*}{Qwen-2.5-1.5B-it} & GSM8K & 0.74 & 0.66 & -0.08 & 0.71 & 0.80 &  0.09 \\
& SVAMP & 0.66 & 0.57 & -0.09 & 0.68 & 0.76 &  0.08 \\
& MathQA & 0.75 & 0.68 & -0.07 & 0.73 & 0.77 &  0.04 \\
\midrule
\multirow{3}{*}{Phi-4-mini-3.8B-it} & GSM8K & 0.66 & 0.46 & -0.19 & 0.56 & 0.52 & -0.04 \\
 & SVAMP & 0.66 & 0.48 & -0.18 & 0.66 & 0.68 &  0.01 \\
& MathQA & 0.65 & 0.52 & -0.13 & 0.62 & 0.63 &  0.01 \\
\midrule
\multirow{3}{*}{Gemma-3-4B-it} & GSM8K & 0.46 & 0.45 & -0.01 & 0.64 & 0.57 & -0.06 \\
& SVAMP & 0.42 & 0.41 & -0.01 & 0.68 & 0.54 & -0.14 \\
& MathQA & 0.75 & 0.83 &  0.09 & 0.36 & 0.52 &  0.17 \\
\midrule
\textbf{Average} & & & & \textbf{-0.07} & & & \textbf{0.04} \\
\bottomrule
\end{tabular}%
}
\end{table*}

\section{Training Details}
\label{app:sec:training_details}

This section summarizes the training and generation configuration for all reported results.
All runs use 4-bit QLoRA on the attention projection modules $\{q,k,v,o\}_{\mathrm{proj}}$, with rank $r{=}32$, $\alpha{=}64$, dropout $0.1$, and a single post-training iteration from the instruction-tuned checkpoint.
Unless otherwise noted, experiments use seed $10$; ablations in Section~\ref{sec:ablations} average over three seeds.
Gold labels and unit-test outcomes are never provided to the oracle or to the standard training objectives; they are used only for evaluation.

\paragraph{Preference construction.}
DPO pairs are formed as the Cartesian product $\mathcal G^+(x)\times\mathcal G^-(x)$ after discarding prompts with an empty preferred or dispreferred set.
Unanimous debates (where $\mathcal G^-(x)=\emptyset$) may still contribute $\mathcal G^+(x)$ to the SFT dataset.
Preference datasets do not include the previous-round peer context; debate itself uses peer context.

\paragraph{Objective-specific settings.}
DPO and KTO are trained for $3$ epochs with per-device batch size $8$, learning rate $1 \times 10^{-5}$, $\beta{=}0.1$, and gradient accumulation $4$, unless a specific run configuration overrides batch size or accumulation.
GRPO uses a frozen match reward $r_x(y)=\mathbf{1}[\mathcal A(y)\equiv\hat a(x)]$, group size $4$ (\texttt{num\_generations}), and the same LoRA recipe.
SFT uses $3$ epoch, batch size $8$, and learning rate $1 \times 10^{-5}$.
Table~\ref{tab:training-params} summarizes the broader hyperparameter ranges explored across objectives.

\paragraph{Generation and debate configuration.}
Unless otherwise specified, generation uses temperature $0.5$, top-$p$ $0.9$, and up to $256$ new tokens (Table~\ref{tab:generation-params}).
Multi-agent debate uses $5$ agents and $2$ rounds by default, with batch debate size $32$--$256$ and synchronous updates (Table~\ref{tab:multiagent-params}).
Table~\ref{tab:model-dataset-config} lists the base models and dataset splits.
GSM8K and MathQA use $1500/500$ train/test splits; SVAMP and MBPP use the respective subset of available test sets for evaluation only.

\begin{table}[h]
\centering
\small
\caption{Training parameters. We report the hyperparameter ranges explored across objectives; the main results use $r{=}32$, $\alpha{=}64$, and objective-specific learning rates and batch sizes given in the text.}
\label{tab:training-params}
\setlength{\tabcolsep}{6pt}
\renewcommand{\arraystretch}{1.1}
\begin{tabular}{lllll}
\toprule
\textbf{Parameter} & \textbf{SFT} & \textbf{GRPO} & \textbf{DPO} & \textbf{KTO} \\
\midrule
Learning rate &
$1 \times 10^{-5}$ &
$1 \times 10^{-5}$ &
$1 \times 10^{-5}$ &
$1 \times 10^{-5}$ \\
Weight decay &
$1 \times 10^{-2}$ &
$1 \times 10^{-2}$ &
$1 \times 10^{-2}$ &
$1 \times 10^{-2}$ \\
Batch size &
8 &
8 &
8 &
8 \\
Epochs &
3 &
3 &
3 &
3 \\
Gradient accumulation steps &
4 &
4 &
4 &
4 \\
Optimizer &
AdamW &
AdamW &
AdamW &
AdamW \\
LoRA rank ($r$) &
32 &
32 &
32 &
32 \\
LoRA $\alpha$ &
64 &
64 &
64 &
64 \\
LoRA dropout &
0.1 &
0.1 &
0.1 &
0.1 \\
LoRA target modules &
\multicolumn{4}{c}{q\_proj, k\_proj, v\_proj, o\_proj} \\
Entropy coefficient &
-- &
0.01 &
-- &
-- \\
Beta (regularization) &
-- &
-- &
0.1 &
0.1 \\
Number of generations per step &
-- &
4 &
-- &
-- \\
Maximum sequence length &
\multicolumn{4}{c}{Model-dependent (2048--4096)} \\
Mixed precision &
\multicolumn{4}{c}{bf16} \\
Quantization &
\multicolumn{4}{c}{4-bit (BitsAndBytesConfig)} \\
\bottomrule
\end{tabular}
\end{table}

\begin{table}[h]
\centering
\small
\caption{Generation parameters used for all debate and evaluation runs unless otherwise noted.}
\label{tab:generation-params}
\setlength{\tabcolsep}{6pt}
\renewcommand{\arraystretch}{1.1}
\begin{tabular}{ll}
\toprule
\textbf{Parameter} & \textbf{Value} \\
\midrule
Temperature & 0.5 \\
Top-$p$ sampling & 0.9 \\
Maximum new tokens & 256~(GSM8K, SVAMP, MATHQA); 1024~(MBPP) \\
Do sample & True \\
Pad token ID & EOS token ID \\
\bottomrule
\end{tabular}
\end{table}

\begin{table}[h]
\centering
\small
\caption{Multi-agent debate configuration. Default settings are $N{=}5$ agents and $M{=}2$ rounds; sweeps in Section~\ref{sec:committee} vary these. 
``Majority vote'' indicates that majority consensus is computed for diagnostics and baselines, not as the training label.}
\label{tab:multiagent-params}
\setlength{\tabcolsep}{6pt}
\renewcommand{\arraystretch}{1.1}
\begin{tabular}{ll}
\toprule
\textbf{Parameter} & \textbf{Value} \\
\midrule
Number of agents & 5 \\
Debate rounds & 2 \\
Batch debate size & 32--256 \\
Majority vote computed & True (diagnostics only) \\
Include debate context & True \\
\bottomrule
\end{tabular}
\end{table}

\begin{table}[h]
\centering
\small
\caption{Base models and dataset splits.}
\label{tab:model-dataset-config}
\setlength{\tabcolsep}{6pt}
\renewcommand{\arraystretch}{1.1}
\begin{tabular}{ll}
\toprule
\textbf{Parameter} & \textbf{Value} \\
\midrule
Base models &
Qwen2.5-1.5B-it, Llama-3.2-1B-it, Phi-4-mini-3.8B-it, Gemma-3-4B-it, Llama-3.1-8B-it \\
Training datasets &
GSM8K, SVAMP, MathQA, MBPP \\
Test datasets &
GSM8K, MathQA, SVAMP, MBPP \\
Training size &
1500 samples (GSM8K, MathQA); 700 samples (SVAMP); 374 samples (MBPP) \\
Test size &
500 samples (GSM8K, MathQA); 300 samples (SVAMP); 200 samples (MBPP) \\
Use quantization &
4-bit (BitsAndBytesConfig) \\
Use adapter mode &
True (LoRA) \\
Use scheduler &
True \\
\bottomrule
\end{tabular}
\end{table}

\section{Average Agent Performance After Post-Training}
\label{app:sec:avg_agent_performance}

Section~\ref{sec:experiment_results} reports consensus accuracy for baselines and for models post-trained on two-round debate traces.
In addition to consensus, we evaluate how post-training affects the average quality of individual agents within the committee.
Table~\ref{app:tab:avg_agent} reports mean per-agent accuracy after $M{=}2$ debate rounds, averaged over agents and test samples.

Columns are organized as follows: ``Ensemble'' and ``Debate'' are inference-only baselines (no post-training); ``SFT'' fine-tunes on the preferred completions $\mathcal G^+(x)$; ``DPO'' and ``KTO'' use the oracle-derived preference labels, while ``GRPO'' uses the oracle-derived reward as described in Section~\ref{sec:experiment_results}.
For each model and dataset, we highlight the best post-training objective in bold.
Across most settings, preference-based objectives improve mean agent accuracy over SFT on the same traces, with DPO providing the largest gains in most cases.

\begin{table*}[h]
\centering
\small
\caption{Mean per-agent accuracy (\%) after $M{=}2$ debate rounds. 
``Ensemble'' and ``Debate'' are inference-only baselines; ``SFT'', ``DPO'', ``KTO'', and ``GRPO'' denote post-training with the corresponding objective on the same debate traces. 
Column definitions match Table~\ref{tab:main_results}. 
Best post-training objective per row is highlighted in bold.}
\label{app:tab:avg_agent}
\begin{tabular}{ll | ccc|ccc}
\toprule
\multirow{2}{*}{Model} & \multirow{2}{*}{Dataset}
& \multicolumn{3}{c|}{Baselines}
& \multicolumn{3}{c}{Post-training} \\
\cmidrule(lr){3-5}\cmidrule(lr){6-8}
& & Ensemble & Debate & SFT & DPO & KTO & GRPO \\
\midrule

\multirow{3}{*}{Llama-3.2-1B-it}
  & SVAMP  & 4.5 & 5.1 & $8.6_{\pm 0.85}$ & $\textbf{18.3}_{\pm 1.48}$ & $15.3_{\pm 2.69}$ & $10.2_{\pm 0.14}$ \\
  & GSM8K  & 4.2 & 5.4 & $5.8_{\pm 0.21}$ & $\textbf{13.2}_{\pm 2.26}$ & $9.85_{\pm 2.47}$ & $5.4_{\pm 1.41}$ \\
  & MathQA & 1.8 & 1.7 & $3.8_{\pm 1.98}$ & $\textbf{3.5}_{\pm 0.49}$ & $2.6_{\pm 0.42}$ & $2.3_{\pm 0.58}$ \\
\midrule

\multirow{3}{*}{Qwen-2.5-1.5B-it}
  & SVAMP  & 50.9 & 60.9 & $63.6_{\pm 0.64}$ & $\textbf{70.6}_{\pm 1.91}$ & $64.5_{\pm 3.75}$ & $59.4_{\pm 0.14}$ \\
  & GSM8K  & 24.2 & 44.5 & $45.3_{\pm 0.64}$ & $\textbf{60.8}_{\pm 1.56}$ & $54.1_{\pm 6.08}$ & $54.7_{\pm 0.35}$ \\
  & MathQA & 8.7 & 23.2 & $23.4_{\pm 1.20}$ & $\textbf{39.8}_{\pm 2.90}$ & $36.1_{\pm 2.12}$ & $35.1_{\pm 1.08}$ \\
\midrule

\multirow{3}{*}{Phi-4-mini-3.8B-it}
  & SVAMP  & 88.2 & 87.4 & $88.5_{\pm 0.00}$ & $\textbf{88.7}_{\pm 0.21}$ & $88.1_{\pm 0.28}$ & $87.8_{\pm 0.42}$ \\
  & GSM8K  & 76.5 & 82.2 & $81.4_{\pm 0.49}$ & ${82.6}_{\pm 2.05}$ & $82.0_{\pm 0.99}$ & $\textbf{83.2}_{\pm 1.63}$ \\
  & MathQA & 18.2 & 18.2 & $29.8_{\pm 1.91}$ & $\textbf{46.8}_{\pm 4.88}$ & $37.6_{\pm 2.83}$ & $31.0_{\pm 2.83}$ \\
\midrule

\multirow{3}{*}{Gemma-3-4B-it}
  & SVAMP  & 68.9 & 61.3 & $55.1_{\pm 3.96}$ & $\textbf{73.3}_{\pm 6.51}$ & $69.0_{\pm 1.84}$ & $67.1_{\pm 3.89}$ \\
  & GSM8K  & 48.6 & 45.3 & $32.9_{\pm 1.77}$ & $\textbf{66.0}_{\pm 2.62}$ & $60.4_{\pm 3.82}$ & $49.8_{\pm 1.77}$ \\
  & MathQA & 9.4 & 8.7 & $6.5_{\pm 0.21}$ & $\textbf{12.9}_{\pm 1.91}$ & $10.1_{\pm 0.21}$ & $8.8_{\pm 0.57}$ \\
\bottomrule
\end{tabular}
\end{table*}

\subsection{How Post-training Objectives Use Debate Traces}
\label{app:sec:how_post_training_uses_debate}

Table~\ref{app:tab:avg_agent} shows that DPO achieves the highest mean post-debate agent accuracy in 11 of 12 model--dataset pairs.
The objectives differ in how they use the structure produced by the debate oracle, providing a possible explanation for this empirical pattern.

\paragraph{Preference data.}
For each prompt $x$, the oracle partitions the final agent completions from the same $M$-round debate into preferred and dispreferred sets:
\[
\mathcal G^+(x) = \{y_i : a_i \equiv \hat a(x)\}, \qquad
\mathcal G^-(x) = \{y_i : a_i \not\equiv \hat a(x)\}
\]
where $a_i = \mathcal A(y_i)$ is the parsed answer and $\hat a(x)$ is the span-confidence winning answer (Section~\ref{sec:methodology}).
Unanimous debates have $\mathcal G^-(x)=\emptyset$ and are excluded from preference learning.
For the remaining examples, the data contain same-$x$ winner--loser comparisons among traces that participated in the same debate.
The oracle therefore supplies the chosen/rejected structure required by DPO, with $y^+\in\mathcal G^+(x)$ and $y^-\in\mathcal G^-(x)$.
The objective explicitly contrasts competing traces for the same question.

\[
\mathcal L_{\mathrm{DPO}}(x, y^+, y^-)
= -\log \sigma\!\Big(
\beta \big[
\log \tfrac{\pi_\theta(y^+|x)}{\pi_{\mathrm{ref}}(y^+|x)}
-
\log \tfrac{\pi_\theta(y^-|x)}{\pi_{\mathrm{ref}}(y^-|x)}
\big]\Big)
\]

\paragraph{SFT: winner-only supervision.}
SFT maximizes $\log \pi_\phi(y^+|x)$ for $y^+\in\mathcal G^+(x)$ and does not include $\mathcal G^-(x)$ in its objective.
Thus, SFT provides no explicit contrast between competing traces and can provide smaller gains than preference-based objectives.

\paragraph{KTO: independent desirability labels.}
KTO assigns desirable or undesirable labels to individual completions rather than explicitly contrasting competing completions for the same prompt.
It therefore uses the oracle labels without directly encoding the pairwise relation between traces in $\mathcal G^+(x)$ and $\mathcal G^-(x)$.
KTO nevertheless improves mean agent accuracy in many settings, indicating that the oracle-derived labels remain useful without explicit pairwise comparisons.

\paragraph{GRPO: reward-based use of the oracle.}
GRPO does not directly optimize on the oracle-labeled debate completions.
Instead, it samples new completions and assigns them the oracle-derived reward
\[
r_x(y) = \alpha(x)\,\mathbf{1}\!\left[\mathcal A(y)\equiv\hat a(x)\right]
\]
where $\alpha(x)$ is a reliability weight.
Because GRPO computes relative advantages within each newly sampled group, groups whose members receive identical rewards provide no within-group ranking signal under the standard formulation.
Thus, unlike DPO, GRPO does not directly contrast the oracle-selected and competing debate traces during optimization.

\section{Ablation details}
\label{app:sec:ablation_details}

\subsection{MBPP evaluation}
\label{app:sec:mbpp}

Official train split of \(374\) problems; \(200\) test problems subsampled from the official test set.
Generation uses the task description only.
The training oracle then clusters parseable programs by their public-test outputs and scores a cluster by how many public cases its members pass.
Challenge assertions are withheld from that oracle.
The reported correctness check executes every public assertion together with any challenge assertion, and it passes only when all of them succeed.
The challenge list may be empty, in which case the check repeats the public assertions but remains all-or-nothing, whereas the oracle still counts how many public cases passed.
Table~\ref{tab:ablate_8b} reports the polled overall public and hidden test cases passed for test-set by all the agents of the committee.

\subsection{Oracle-Protocol Runs}
\label{app:sec:oracle_runs}

Other than the results shared in Table~\ref{tab:ablate_oracle}, we also evaluated two aggregation rules for the answer-span confidence oracle:
confidence \emph{sum} and confidence \emph{mean}.
All runs use Qwen-2.5-1.5B-it on a GSM8K splits.
The committee uses $N{=}5$ agents and $M{=}2$ debate rounds, followed by one DPO post-training iteration.
Table~\ref{app:tab:ablate_oracle} reports accuracy after the initial response, after the exchange round, and after aggregation by the committee.

\begin{table*}[h]
\centering
\small
\caption{Oracle-protocol ablation on GSM8K using Qwen-2.5-1.5B-it, $N{=}5$ agents, and $M{=}2$ debate rounds. 
``Initial'' denotes accuracy~(\%) before debate, ``Exchange'' denotes accuracy~(\%) after exchange, and ``Consensus'' denotes the accuracy~(\%) of the aggregated committee answer. 
}
\label{app:tab:ablate_oracle}
\setlength{\tabcolsep}{8pt}
\renewcommand{\arraystretch}{1.15}
\begin{tabular}{llccc}
\toprule
\textbf{Aggregation} & \textbf{Method}
& \textbf{Initial}
& \textbf{Exchange}
& \textbf{Consensus} \\
\midrule
& Ensemble & 24.2 & - & 48.6 \\
\midrule
\multirow{2}{*}{Sum}
& Debate   & 24.0 & 44.5 & 66.0 \\
& DPO      & 35.4 & 59.7 & 71.6 \\
\midrule
\multirow{2}{*}{Mean}
& Debate   & 23.8 & 44.0 & 63.2 \\
& DPO      & 38.9 & 60.6 & 67.6 \\
\bottomrule
\end{tabular}
\end{table*}

For each prompt $x$, the oracle parses the agents' final answers, clusters them by task-level equivalence, and assigns each cluster $\mathcal C_k$ a confidence mass
\[
w_k = \sum_{i\in\mathcal C_k} c_i,
\]
where $c_i$ is the answer-span confidence of agent $i$.
The selected cluster is
\[
k^\star = \operatorname*{arg\,max}_k (w_k,n_k),
\qquad n_k = |\mathcal C_k|,
\]
where $n_k$ provides a deterministic tie-break based on cluster size.

\begin{figure*}[t]
    \centering
    \begin{subfigure}[t]{0.435\textwidth}
        \centering
        \includegraphics[width=\linewidth]{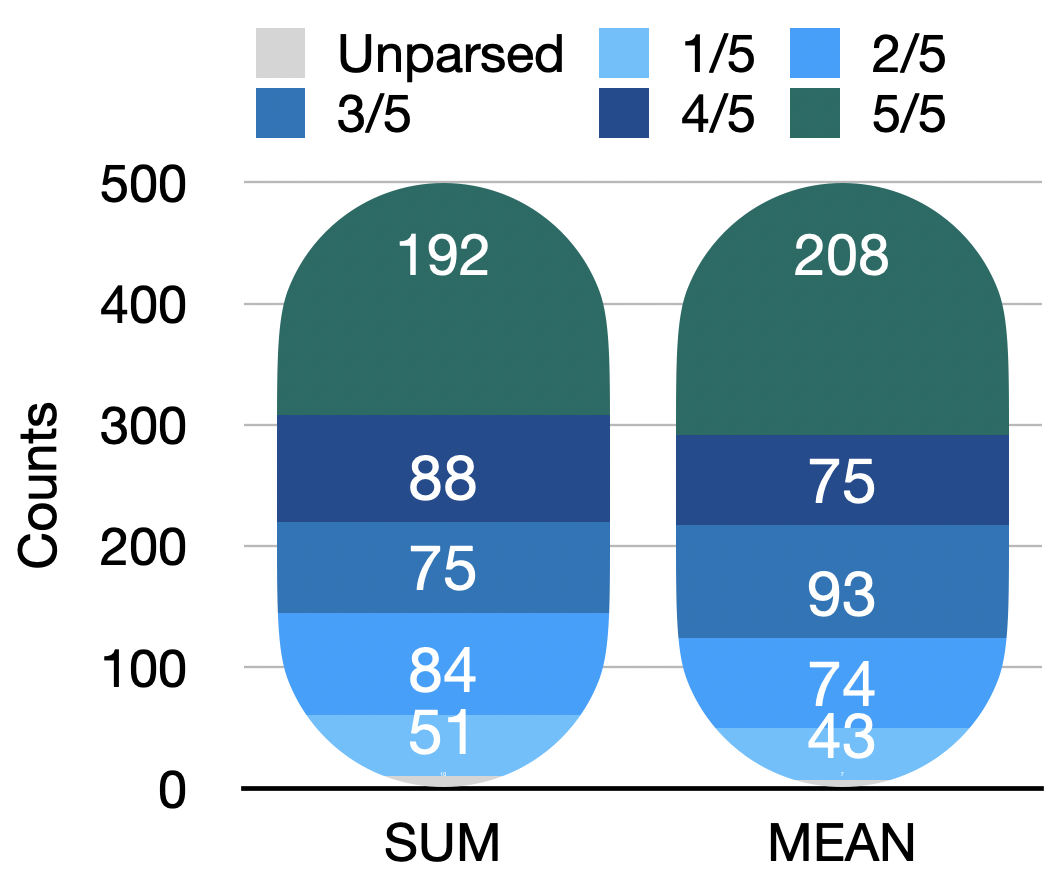}
        \caption{Agreement-rate count}
        \label{fig:oracle_sum}
    \end{subfigure}
    \hspace{0.01\textwidth}
    \begin{subfigure}[t]{0.47\textwidth}
        \centering
        \includegraphics[width=\linewidth]{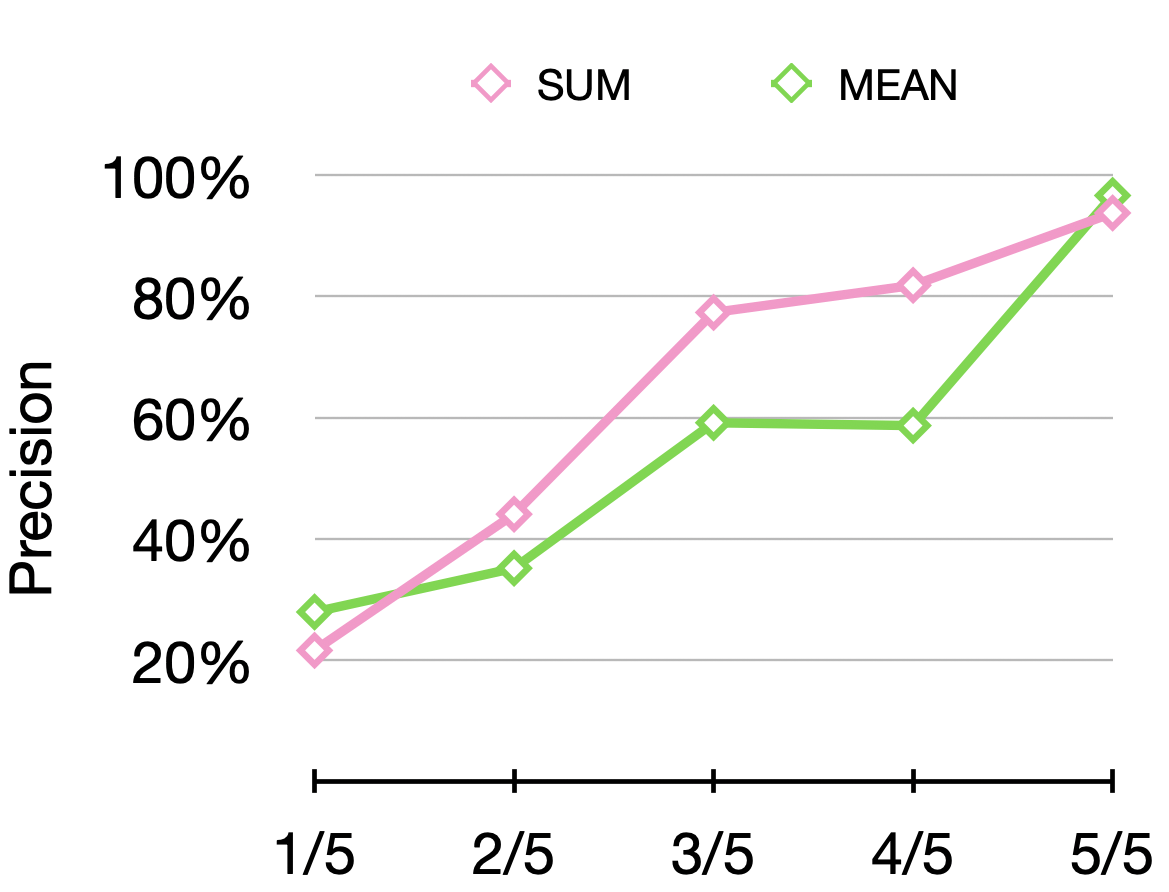}
        \caption{Correctness per agreement-rate}
        \label{fig:oracle_mean}
    \end{subfigure}
    \caption{
    Comparison of confidence-sum and confidence-mean aggregation for answer-equivalence clusters on GSM8K with Qwen-2.5-1.5B-it and five agents.
    The agreement rate is the fraction of parseable agent answers assigned to the selected cluster.
    \textbf{Left:} frequency with which each aggregation rule selects an answer at each agreement rate.
    \textbf{Right:} correctness of the selected answer within each agreement-rate bin, measured against GSM8K reference answers for evaluation only.
    Confidence-sum aggregation retains the total confidence mass contributed by the committee, whereas confidence-mean aggregation normalizes away cluster size.
    Across agreement-rate bins, the sum rule selects more accurate oracle answers than the mean rule.
    }
    \label{fig:oracle_protocol_runs}
\end{figure*}

\paragraph{Why we use confidence sum in Sec~\ref{sec:experiment_results}.}
We use confidence sum rather than mean because a cluster represents the committee's accumulated evidence for an answer, not the confidence of a typical individual member.
The sum therefore acts as a confidence-weighted vote:
\[
w_k = \sum_{i\in\mathcal C_k} c_i.
\]
It allows several moderately confident agents to outweigh a single highly confident outlier, while still allowing two highly confident agents to outweigh several weakly confident agents.
By contrast, mean confidence normalizes away cluster size. 
For example, a singleton with confidence $0.95$ would defeat a four-agent cluster whose members each have confidence $0.80$, because $0.95 > 0.80$, despite the latter cluster containing substantially more independent agreement.
Mean confidence can therefore treat a singleton as equivalent to a crowd and make cluster size contribute only indirectly through the tie-break.
%
%
It is also sensitive to miscalibration: an overconfident incorrect answer can dominate under mean aggregation, whereas sum aggregation requires its confidence to outweigh the accumulated confidence of the competing cluster.
Figure~\ref{fig:oracle_protocol_runs} visualizes the corresponding accuracy trends for the two aggregation rules.
Table~\ref{app:tab:ablate_oracle} alongside motivate the use of confidence mass as the default aggregation rule in \method{}.
Both the debate and post-trained checkpoint consensus accuracy are better in aggregation by sum compared to mean.

\subsection{Majority Voting vs.\ Span-Confidence Selection}
\label{app:sec:oracle_comparison}

We compare majority voting (MV) with span-confidence selection on identical debate transcripts before any post-training.
For each model--dataset pair, both rules operate on the same sampled responses and answer-equivalence clusters; gold labels are used only to evaluate the selected answer.
\emph{Span} denotes the accuracy of the cluster selected by aggregate claimed-answer-span confidence, while \emph{MV} denotes the accuracy of the cluster selected by majority vote.
\emph{Disagree} is the fraction of examples for which the two rules select different clusters.
For disagreement cases, we report the accuracy of the cluster selected by each rule.
We additionally report the corresponding accuracy on count-tie cases.

On GSM8K, the two rules disagree on only $2.6\%$ of examples on average across the four models, and their overall selection accuracies are similar ($57.5\%$ for span confidence and $57.2\%$ for MV).
The difference is larger on cases where the rules disagree: the span-selected cluster is correct on $31.4\%$ of such cases, compared with $7.8\%$ for MV.
Span confidence also achieves higher accuracy on count-tie cases ($29.8\%$ vs.\ $18.3\%$).
Across SVAMP and MathQA, the same analysis shows that the relative advantage is dataset- and model-dependent.
Overall, Span confidence is most useful precisely when majority voting is ambiguous or the two rules disagree.

\begin{table*}[t]
\centering
\small
\caption{
Comparison of majority voting (MV) and aggregate claimed-answer-span
confidence on identical transcripts before post-training.
\emph{Span} and \emph{MV} report overall selection accuracy (\%).
\emph{Disagree} is the percentage of examples for which the two rules select
different answer-equivalence clusters.
\emph{Disagree: Span/MV} reports the accuracy of the respective selected
cluster on disagreement cases.
\emph{Ties} is the percentage of count-tie cases, with
\emph{Ties: Span/MV} reporting the corresponding selection accuracy.
Gold labels are used only for evaluation.
}
\label{app:tab:oracle_comparison}
\begin{adjustbox}{max width=\textwidth}
\begin{tabular}{llcccccc}
\toprule
\textbf{Model} & \textbf{Dataset}
& \textbf{Span} & \textbf{MV}
& \textbf{Disagree}
& \textbf{Disagree: Span/MV}
& \textbf{Ties}
& \textbf{Ties: Span/MV} \\
\midrule

\multirow{3}{*}{Llama-3.2-1B-it}
& SVAMP  & 29.3 & 28.7 & 3.3 & 40.0 / 20.0 &  4.3 & 46.2 / 30.8 \\
& GSM8K  & 15.6 & 14.6 & 3.2 & 31.2 / 0.0  & 5.8 & 24.1 / 6.9 \\
& MathQA & 5.0  & 5.2  & 0.6 & 0.0 / 33.3  & 0.8 & 0.0 / 25.0 \\
\midrule

\multirow{3}{*}{Qwen-2.5-1.5B-it}
& SVAMP  & 79.7 & 79.7 & 3.0 & 22.2 / 22.2 & 7.0 & 33.3 / 33.3 \\
& GSM8K  & 66.0 & 65.0 & 5.8 & 31.0 / 13.8 & 12.8 & 31.2 / 23.4 \\
& MathQA & 45.2 & 42.2 & 8.8 & 47.7 / 13.6 & 12.8 & 43.8 / 25.0 \\
\midrule

\multirow{3}{*}{Phi-4-mini-3.8B-it}
& SVAMP  & 89.7 & 89.7 & 0.7 & 0.0 / 0.0   & 1.7 & 20.0 / 20.0 \\
& GSM8K  & 88.8 & 88.4 & 1.2 & 33.3 / 0.0  & 1.8 & 33.3 / 11.1 \\
& MathQA & 41.6 & 42.6 & 1.8 & 22.2 / 77.8 & 2.2 & 27.3 / 63.6 \\
\midrule

\multirow{3}{*}{Gemma-3-4B-it}
& SVAMP  & 75.7 & 76.0 & 0.3 & 0.0 / 100.0 & 0.3 & 0.0 / 100.0 \\
& GSM8K  & 60.6 & 60.6 & 0.0 & 0.0 / 0.0   & 0.4 & 50.0 / 50.0 \\
& MathQA & 16.6 & 16.6 & 0.0 & 0.0 / 0.0   & 0.0 & 0.0 / 0.0 \\
\midrule

\textbf{Average}
& SVAMP  & 68.6 & 68.5 & 1.8 & 27.3 / 22.7 & 3.3 & 32.5 / 35.0 \\
& GSM8K  & 57.5 & 57.2 & 2.6 & 31.4 / 7.8  & 5.2 & 29.8 / 18.3 \\
& MathQA & 27.1 & 26.7 & 2.8 & 41.1 / 25.0 & 3.95 & 39.2 / 30.4 \\
\bottomrule
\end{tabular}
\end{adjustbox}
\end{table*}

\subsection{Heterogeneous Committee}
\label{app:sec:multi_model_committee}

To examine whether debate benefits from architectural and scale diversity, we form a heterogeneous committee from five instruction-tuned models: L1 (Llama-3.2-1B-it), Q1 (Qwen-2.5-1.5B-it), Q2 (Qwen3-1.7B-it), P4 (Phi-4-mini-3.8B-it), and G4 (Gemma-3-4B-it).
All agents use temperature $0.5$, and the committee debates for $M{=}2$ rounds with $N{=}5$ agents on GSM8K.
The ensemble, debate, and post-training results are reported in Table~\ref{tab:ablate_arch}.
DPO updates all five models using the shared preferred and dispreferred sets $\mathcal G^+(x)$ and $\mathcal G^-(x)$ constructed from answer-span confidence.
In addition to consensus accuracy, we report the average accuracy of individual agents before and after debate and after one iteration with SFT, DPO, KTO and GRPO.
The smaller models (L1, Q1, and Q2) generally benefit more from interaction with the larger committee members, whereas the gains for the larger models (P4 and G4) are less.
This pattern suggests that heterogeneous debate can transfer useful information to weaker agents, although it does not uniformly improve every member.

\begin{table}[t]
\centering
\small
\caption{Accuracy~(\%) of a heterogeneous committee on GSM8K. 
The committee consists of L1 (Llama-3.2-1B-it), Q1 (Qwen-2.5-1.5B-it), Q2 (Qwen3-1.7B-it), P4 (Phi-4-mini-3.8B-it), and G4 (Gemma-3-4B-it). 
``Consensus'' denotes accuracy of aggregate solution. 
``Avg.\ Agent'' denotes mean per-agent performance. 
The $\Delta$ row reports the per-agent change from Ensemble to DPO.
}
\label{tab:ablate_arch}
\setlength{\tabcolsep}{6pt}
\renewcommand{\arraystretch}{1.15}
\begin{tabular}{l | ccccccc}
\toprule
\multirow{2}{*}{\textbf{Method}} &
\multicolumn{7}{c}{\textbf{Accuracy}} \\
\cmidrule(lr){2-8}
& \textbf{Consensus} & \textbf{Avg.\ Agent} & \textbf{L1} & \textbf{Q1} & \textbf{Q2}
& \textbf{P4} & \textbf{G4} \\
\midrule
Ensemble
& 78.2 & 30.9 & 3.6 & 26.0 & 1.4 & {76.8} & 46.8 \\
\midrule
Debate
& 79.2 & 36.8 & 16.6 & {57.6} & 3.2 & 71.6 & 35.2 \\
\midrule
SFT 
& 79.8 & 37.4	& 17.8	& 54.2	& 4.2	& 75.6	& 35.4 \\
\midrule
DPO
& \textbf{87.0} & \textbf{68.2}	& \textbf{71.0}	& \textbf{75.8} & \textbf{41.8} & 81.2 & \textbf{71.4} \\
\midrule
KTO 
& 86.6 & 59.2	& 52.0	& 71.6	& 22.6 	& \textbf{83.6} & 	66.0 \\
\midrule
GRPO 
& 83.8 & 48.5	& 28.6	& 70.4	& 7.2	& 79.8	& 56.4 \\
\midrule
$\Delta$ (DPO $-$ Ensemble)
& 8.8 & 37.3 & 67.4	& 49.8	& 40.4	& 4.4 &	24.6\\
\bottomrule
\end{tabular}
\end{table}




\section{Limitations of the Experimental Protocol}
\label{app:sec:limits}

Our evaluation has several limitations that bound the conclusions supported by the experiments.
Most main-grid cells use two post-training runs, while oracle and weighting ablations use three seeds where explicitly reported; these repetitions provide limited evidence about variance and statistical significance.
We evaluate on subsets of the available training and test splits rather than the full datasets, limiting the precision with which the reported improvements characterize performance on the complete benchmarks.
Post-training is restricted to a single iteration from each instruction-tuned checkpoint with QLoRA, so the experiments do not establish whether the observed gains persist under longer or repeated self-training.
The maximum generation length of $256$ tokens can truncate reasoning or final-answer spans, with truncation effects particularly visible for Llama-3.2-1B-it on GSM8K.
Although the oracle uses no gold labels for supervision, it requires task-specific answer parsing and an equivalence relation, making the approach less directly applicable to outputs for which correctness cannot be parsed or reliably compared.
The reported code-generation extension uses execution-based equivalence and therefore does not establish applicability to arbitrary open-ended generation without an executable correctness criterion.
Answer-span confidence is computed from token probabilities and is not calibrated against correctness; the method therefore assumes that these probabilities provide a sufficiently informative within-committee ranking signal.
Confidence scores are not recalibrated across heterogeneous architectures, and the heterogeneous-committee results show that raw token probabilities need not be directly comparable across models.
The confidence oracle is evaluated against gold answers only retrospectively to measure selection quality; these labels are never available during oracle construction or post-training, so the experiments establish empirical utility rather than a guarantee that the oracle identifies the correct cluster.
Finally, the experiments cover five instruction-tuned model families and four primary datasets.
%

\paragraph{Near-ceiling GSM8K accuracy.}
Phi-4-mini-3.8B-it reaches \(88.6\%\) on GSM8K and \(88.2\%\) on SVAMP under independent sampling, and the best gain over debate is \(0.7\)pp and \(0.3\)pp (Table~\ref{tab:main_results}).
On MathQA the same checkpoint moves from \(41.6\%\) after debate to \(59.5\%\) after DPO.
The small GSM8K movement is the behavior of a model with little remaining headroom on that benchmark due to overfitting on the test set.
Earlier work~\citep{zhang2024careful} find that earlier Phi releases score higher on GSM8K than on GSM1k, a held-out set matched to GSM8K in difficulty, which is consistent with partial familiarity with GSM8K-style items.
MathQA is the dataset on which this checkpoint's base accuracy leaves room to observe the oracle.



\section{Larger dataset splits}
\label{app:sec:full_dataset_expts}



Section~\ref{sec:experiment_results} trains on subsets of the available splits.
Appendix~\ref{app:sec:training_details} states the sizes: \(1{,}500\) training and \(500\) test examples for GSM8K and MathQA.
The limitation is recorded in Appendix~\ref{app:sec:limits}: subset evaluation does not by itself characterize the full benchmark.
This section repeats the locked protocol on larger splits of the same two datasets, for Qwen-2.5-1.5B-it.
GSM8K uses the official training and test sets, \(7{,}473\) and \(1{,}319\) examples.
MathQA uses \(7{,}015\) training and \(2{,}985\) test examples.
The committee is unchanged: \(N{=}5\), \(M{=}2\), temperatures \(\{0.01,0.25,0.5,0.75,1.0\}\), top-\(p{=}0.9\), and at most \(256\) new tokens.
Preferred and dispreferred traces are still assigned by claimed answer-span confidence.
Gold answers are used only to compute the accuracies in Table~\ref{app:tab:full_dataset_expts}.
%

\begin{table}[h]
\centering
\small
\caption{
Consensus accuracy~(\%) for Qwen-2.5-1.5B-it on the larger GSM8K and MathQA splits.
\emph{Ensemble} is inference-only independent sampling and \emph{Debate} is inference-only debate, under the same span-confidence selection used in Table~\ref{tab:main_results}.
Post-training is one QLoRA iteration on debate traces from the larger training split.
The final column is the absolute improvement of the best post-training entry over inference-only debate, in percentage points.
Each entry is a single run.
}
\label{app:tab:full_dataset_expts}
\begin{adjustbox}{max width=\textwidth}
\begin{tabular}{ll | ccc|cc c}
\toprule
\multirow{2}{*}{Model} & \multirow{2}{*}{Dataset}
& \multicolumn{3}{c|}{Baselines}
& \multicolumn{2}{c}{Post-training}
& \multirow{2}{*}{Best $\Delta$} \\
\cmidrule(lr){3-5}\cmidrule(lr){6-7}
& & Ensemble & Debate & SFT & DPO & KTO & \\ 
\midrule
\multirow{2}{*}{Qwen-2.5-1.5B-it}
  & GSM8K
  & 46.6 & 65.6 & 65.4 & 72.1 & \textbf{72.9} & $7.3_{pp}$\\
  & MathQA
  & 20.1 & 45.1 & 44.0 & \textbf{54.7} & 45.8 & $9.6_{pp}$\\
\bottomrule
\end{tabular}
\end{adjustbox}
\end{table}

On GSM8K, inference-only debate reaches \(65.6\%\) and the best post-training objective, KTO, reaches \(72.9\%\) (\(+7.3_{pp}\)).
On MathQA, inference-only debate reaches \(45.1\%\) and DPO reaches \(54.7\%\) (\(+9.6_{pp}\)).
Both gains point in the same direction as the subset results in Table~\ref{tab:main_results}, where the corresponding best improvements are \(+5.4_{pp}\) on GSM8K and \(+8.5_{pp}\) on MathQA.
%
%
On the GSM8K subset the best entry is DPO; on this larger GSM8K split it is KTO, and DPO remains close (\(72.1\%\)).
On the larger MathQA split KTO falls to \(45.8\%\), next to the debate baseline, whereas KTO is the best subset entry in Table~\ref{tab:main_results}.
SFT stays within one percentage point of inference-only debate on both larger splits.
%
%
It does show that moving from the \(1{,}500\)-example subset to these larger splits does not remove the improvement of preference post-training over inference-only debate.

\section{Repeated post-training iterations}
\label{app:sec:more_iterations}

This section continues the iterations.
At iteration \(0\) the policy is the instruction-tuned checkpoint.
Each later iteration rebuilds the debate traces with the current policy, rebuilds \(\mathcal G^+(x)\) and \(\mathcal G^-(x)\) with the span-confidence oracle, and post-trains with DPO objective, same committee, and token limit as Section~\ref{sec:experimental_setup}.
We report three test-set quantities defined in Section~\ref{sec:experimental_setup}: 
pre-debate agent accuracy (the mean accuracy of the independent samples), 
post-debate agent accuracy (the mean accuracy of the five final-round traces), and 
consensus accuracy (the accuracy of the span-selected cluster).

\begin{table}[h]
\centering
\small
\caption{
Test accuracy~(\%) across post-training iterations for Llama-3.2-1B-it and Qwen-2.5-1.5B-it.
It-\(0\) is the instruction-tuned checkpoint.
Each later iteration rebuilds debate traces from the updated policy and applies one further post-training step with the span-confidence oracle.
Pre-debate is mean agent accuracy before the exchange round; final-round average is mean agent accuracy after \(M{=}2\); consensus is the accuracy of the span-selected cluster.
Gold answers are used only for evaluation.
}
\label{app:tab:more_iterations_trend}
\begin{adjustbox}{max width=\textwidth}
\begin{tabular}{ll | ccccc|ccccc|ccccc}
\toprule
\multirow{2}{*}{Model} & \multirow{2}{*}{Dataset}
& \multicolumn{5}{c|}{Pre-debate agent}
& \multicolumn{5}{c|}{Post-debate agent}
& \multicolumn{5}{c}{Consensus} \\
\cmidrule(lr){3-7}\cmidrule(lr){8-12}\cmidrule(lr){13-17}
& & It-0 & It-1 & It-2 & It-3 & It-4
& It-0 & It-1 & It-2 & It-3 & It-4
& It-0 & It-1 & It-2 & It-3 & It-4 \\
\midrule
\multirow{2}{*}{Llama-3.2-1B-it}
  & GSM8K
  & 3.6 & 5.3 & 5.5 & 4.6 & 5.6
  & 5.2 & 9.3 & 8.4 & 8.3 & 9.1
  & 15.0 & 23.6 & 21.6 & 20.8 & 22.4 \\
  & MathQA
  & 1.4 & 2.2 & 2.0 & 3.0 & 2.4
  & 1.9 & 2.8 & 2.4 & 3.0 & 3.5
  & 5.6 & 7.4 & 7.4 & 9.8 & 10.2 \\
\midrule
\multirow{2}{*}{Qwen-2.5-1.5B-it}
  & GSM8K
  & 24.0 & 39.8 & 39.8 & 39.8 & 39.8
  & 44.5 & 61.9 & 60.7 & 60.5 & 60.0
  & 66.0 & 70.4 & 70.4 & 70.0 & 69.8 \\
  & MathQA
  & 8.5 & 13.7 & 13.9 & 13.2 & 12.7
  & 23.2 & 34.0 & 36.5 & 35.4 & 32.5
  & 45.2 & 52.8 & 55.0 & 54.8 & 49.8 \\
\bottomrule
\end{tabular}
\end{adjustbox}
\end{table}

For Qwen-2.5-1.5B-it on GSM8K, consensus accuracy moves from \(66.0\%\) to \(70.4\%\) at iteration \(1\) and is \(69.8\%\) at iteration \(4\).
Pre-debate agent accuracy on that pair moves from \(24.0\%\) to \(39.8\%\) at iteration \(1\) and does not move further.
For Llama-3.2-1B-it on GSM8K, consensus accuracy peaks at iteration \(1\) (\(23.6\%\)) and is lower at iterations \(2\) and \(3\).
Llama-3.2-1B-it on MathQA is the exception among these four rows: consensus accuracy continues from \(7.4\%\) at iteration \(1\) to \(10.2\%\) at iteration \(4\), still on a low base.
Qwen-2.5-1.5B-it on MathQA peaks at iteration \(2\) (\(55.0\%\)) and falls to \(49.8\%\) at iteration \(4\), which is below its iteration-\(1\) consensus accuracy.
Repeated self-training therefore does not produce a consistent further gain after the first iteration on these splits.
The single-iteration protocol in the main experiments is the setting in which the measured improvement appears.

\section{Longer generation budget}
\label{app:sec:more_generation_tokens}

The main experiments generate at most \(256\) new tokens per agent turn.
Appendix~\ref{app:sec:limits} notes that this cap can truncate the reasoning trace or the claimed answer span.
A missing span makes \(c_i^{\mathrm{span}}\) undefined, removes the trace from equivalence clustering, and counts it as incorrect at evaluation.
Part of what post-training changes may therefore be whether the model emits a parseable answer inside the cap.
This section raises the cap to \(512\) new tokens and repeats the protocol for Qwen-2.5-1.5B-it, the model used for the committee sweeps in Section~\ref{sec:committee} with results noted in Table~\ref{app:tab:large_completion_length}.
%
%
The \(256\)-token rows are the corresponding rows of Table~\ref{tab:main_results}.
The \(512\)-token GSM8K row is a single run.
%

\begin{table}[h]
\centering
\small
\caption{
Consensus accuracy~(\%) for Qwen-2.5-1.5B-it under two generation caps.
The \(256\)-token block includes the standard deviation over the two runs in Table~\ref{tab:main_results}.
The \(512\)-token block is a single run.
The final column is the absolute improvement of the best post-training entry over inference-only debate, in percentage points.
}
\label{app:tab:large_completion_length}
\begin{adjustbox}{max width=\textwidth}
\begin{tabular}{ll | ccc|ccc c}
\toprule
\multirow{2}{*}{Model} & \multirow{2}{*}{Dataset}
& \multicolumn{3}{c|}{Baselines}
& \multicolumn{3}{c}{Post-training}
& \multirow{2}{*}{Best $\Delta$} \\
\cmidrule(lr){3-5}\cmidrule(lr){6-8}
& & Ensemble & Debate & SFT & DPO & KTO & GRPO & \\
\midrule
& & \multicolumn{7}{c}{\(256\) new tokens} \\
\cmidrule(lr){3-9}
\multirow{4}{*}{Qwen-2.5-1.5B-it}
  & GSM8K
  & 44.5 & 66.0 & \(66.3_{\pm 0.71}\) & \(\textbf{71.4}_{\pm 0.28}\) & \(69.2_{\pm 1.13}\) & \(68.9_{\pm 0.14}\) & \(5.4_{pp}\) \\
  & MathQA
  & 21.6 & 45.2 & \(46.0_{\pm 2.26}\) & \(53.0_{\pm 1.13}\) & \(\textbf{53.7}_{\pm 1.27}\) & \(51.1_{\pm 1.56}\) & \(8.5_{pp}\) \\
\cmidrule(lr){3-9}
& & \multicolumn{7}{c}{\(512\) new tokens} \\
\cmidrule(lr){3-9}
  & GSM8K
  & 49.5 & 70.8 & 71.0 & 72.2 & \textbf{75.4} & 70.6 & \(4.6_{pp}\) \\
  & MathQA
  & 53.0 & 57.8 & 60.2 & 62.6 & \textbf{63.2} & 59.2 & $5.4_{pp}$ \\
\bottomrule
\end{tabular}
\end{adjustbox}
\end{table}

On GSM8K, raising the cap from \(256\) to \(512\) tokens raises inference-only debate from \(66.0\%\) to \(70.8\%\), and raises independent-sampling consensus from \(44.5\%\) to \(49.5\%\).
The \(256\)-token debate baseline is therefore partly a property of the generation budget, not only of the committee.
Preference post-training still improves on the longer budget: KTO reaches \(75.4\%\), which is \(+4.6_{pp}\) over the \(512\)-token debate baseline, compared with \(+5.4_{pp}\) for the best objective under the \(256\)-token cap.
SFT remains within \(0.2_{pp}\) of debate at both caps on this dataset (\(66.3\%\) and \(71.0\%\)).
The longer budget changes which preference objective is best in the point estimate, from DPO at \(256\) tokens to KTO at \(512\) tokens.
%


\section{Comparison with majority-vote debate post-training}
\label{app:sec:basline_maca_comparison}

The closest post-training use of debate traces labels a completion by vote count.
Multi-Agent Consensus Alignment treats traces whose final answer matches the majority answer as preferred and the dissenting traces as dispreferred, then applies SFT, DPO, KTO, or a majority-vote reward under GRPO~\citep{samanta2026maca}.
\method{} uses the same preference objectives and the same requirement that a prompt contribute to DPO or KTO only when both a preferred and a dispreferred set exist.
The difference is the rule that fills those sets.
MACA uses \(k_{\mathrm{maj}}=\operatorname*{arg\,max}_k n_k\).
\method{} uses \(k^\star=\operatorname*{arg\,max}_k(w_k,n_k)\), where \(w_k\) is the sum of claimed answer-Span-confidences in the cluster (Section~\ref{sec:oracle}).
Agreement rate does not choose \(k^\star\).

\begin{table}[h]
\centering
\small
\caption{
Matched comparison of majority-vote labels and span-confidence labels.
Both rows in a pair are trained on the same debate traces, with the same objective and the same QLoRA setup.
Consensus is the accuracy~(\%) of the cluster selected at test time by the oracle named in the row.
Gold answers are used only for evaluation.
}
\label{app:tab:maca_matched}
\begin{adjustbox}{max width=\textwidth}
\begin{tabular}{ll | ccc|ccc c}
\toprule
\multirow{2}{*}{Model} & \multirow{2}{*}{Dataset}
& \multicolumn{3}{c|}{Baselines}
& \multicolumn{3}{c}{Post-training}
& \multirow{2}{*}{Best $\Delta$} \\
\cmidrule(lr){3-5}\cmidrule(lr){6-8}
& & Ensemble & Debate & SFT & DPO & KTO & GRPO & \\

  

\midrule
& & \multicolumn{7}{c}{Span-confidence} \\
\cmidrule(lr){3-9}
\multirow{4}{*}{Qwen-2.5-1.5B-it}
  & GSM8K
  & 44.5 & 66.0 & \(66.3_{\pm 0.71}\) & \(\textbf{71.4}_{\pm 0.28}\) & \(69.2_{\pm 1.13}\) & \(68.9_{\pm 0.14}\) & \(5.4_{pp}\) \\
  & MathQA
  & 21.6 & 45.2 & \(46.0_{\pm 2.26}\) & \(53.0_{\pm 1.13}\) & \(\textbf{53.7}_{\pm 1.27}\) & \(51.1_{\pm 1.56}\) & \(8.5_{pp}\) \\
  
\cmidrule(lr){3-9}
& & \multicolumn{7}{c}{Majority voting} \\
\cmidrule(lr){3-9}
  & GSM8K
  &  44.5 &  65.0 & 63.8 & 68.6 & \textbf{70.8} & 67.0 & \(5.8_{pp}\) \\
  & MathQA
  &  21.6 &  42.2 & 42.2 &  \textbf{47.4} & 45.6 &  44.4 & $5.2_{pp}$ \\

\midrule
& & \multicolumn{7}{c}{Span-confidence} \\
\cmidrule(lr){3-9}
\multirow{4}{*}{Gemma-3-4B-it}
  & GSM8K  & 65.2 & 60.6 & $51.8_{\pm 0.85}$ & $\textbf{72.7}_{\pm 1.27}$ & $68.8_{\pm 2.26}$ & $62.9_{\pm 0.14}$ & $12.1_{pp}$ \\
  & MathQA & 15.8 & 16.6 & $11.8_{\pm 0.57}$ & $\textbf{21.5}_{\pm 2.12}$ & $17.3_{\pm0.99}$ & $16.3_{\pm 1.56}$ & $4.9_{pp}$ \\

\cmidrule(lr){3-9}
& & \multicolumn{7}{c}{Majority voting} \\
\cmidrule(lr){3-9}
  & GSM8K  & 65.0 & 60.6 & 57.8 & \textbf{66.8} & 66.4 & 63.2 & $6.2_{pp}$\\ 
  & MathQA & 15.8 & 16.6 & 11.8 & 19.0 & \textbf{20.2} & 14.8 & $3.6_{pp}$\\ 
\bottomrule
\end{tabular}
\end{adjustbox}
\end{table}

Table~\ref{app:tab:maca_matched} compares majority-vote and span-confidence oracle pipelines on GSM8K and MathQA using Qwen-2.5-1.5B-it and Gemma-3-4B-it.
For each model--dataset setting, both pipelines use the same source debate transcripts and the same objective-specific training hyperparameters.
The only change from row to row is whether \(\mathcal G^+(x)\) is the majority cluster or the span-confidence cluster.
Both labeling rules raise the best post-training consensus over the shared debate baseline. 
Span-confidence achieves higher best post-training consensus accuracy in all four model-dataset settings.
For Qwen-2.5-1.5B-it, the best scores increase from $70.8$ to $71.4$ on GSM8K and from $47.4$ to $53.7$ on MathQA.
For Gemma-3-4B-it, they increase from $66.8$ to $72.7$ and from $20.2$ to $21.5$, respectively.
The advantage is not uniform across training objectives.
For example, majority-labeled KTO outperforms span-labeled KTO on Qwen GSM8K, while majority-labeled SFT outperforms span-labeled SFT on Gemma GSM8K.
These results support span-confidence as a competitive oracle pipeline, while showing that its benefits depend on the model, dataset, and post-training objective.

\paragraph{Factorial Evaluation.}
We further did the comparison in a factorial format on GSM8K with a temperature-diverse Qwen-2.5-1.5B-it committee.
Each of DPO, KTO, and GRPO is trained once on majority-vote labels and once on span-confidence labels, and both rules are recomputed on the same post-training test debate.
Accuracy in Table~\ref{tab:factorial-gsm8k} is the number of correct answers over the expected 500 test questions, with a missing debate counted as incorrect. 
Span-confidence at test time is the more accurate readout in every cell. 
%
%
On one of the runs (quoted in Table~\ref{tab:factorial-gsm8k}) the best accuracy is 71.6\%, reached by DPO trained on Span-confidence. 

\begin{table}[t]
\centering
\caption{Factorial evaluation of Qwen-2.5-1.5B-it on GSM8K test ($N{=}5$).
Rows are the training oracle. 
Columns are the test-time consensus, recomputed on the post-training debate. 
Entries are accuracy over 500 questions, with the correct count in parentheses. 
Span-confidence sums boxed-answer probabilities inside an answer cluster. }
\label{tab:factorial-gsm8k}
\small
\begin{tabular}{llccc}
\toprule
Method & Training oracle & Majority & Span-confidence & Gap \\
\midrule
DPO  & Span-confidence & 71.2 (356) & \textbf{71.6} (358) & +2 \\
KTO  & Span-confidence & 68.4 (342) & \textbf{70.0} (350) & +8 \\
GRPO & Span-confidence & 68.6 (343) & \textbf{69.0} (345) & +2 \\
\midrule
DPO  & Majority voting & 68.6 (343) & \textbf{70.0} (350) & +7 \\
KTO  & Majority voting & 70.8 (354) & \textbf{71.6} (358) & +4 \\
GRPO & Majority voting & 67.0 (335) & \textbf{68.6} (343) & +8 \\
\bottomrule
\end{tabular}
\end{table}

\section{Comparison with independent-sample post-training}
\label{app:sec:ind_sample_post_train_comparison}

Section~\ref{sec:preference_training} fits SFT, DPO, KTO, and GRPO on traces collected after one independent-sample generation followed by one exchange round.
In this section we replace that collection with only independent-samples of the same policy and keeps every other choice fixed.
The model is Qwen-2.5-1.5B-it, the dataset is GSM8K, and both pipelines use \(N{=}5\) temperature-diverse agents, and same training setup.
The independent-sample pipeline stops after the initial generation.
Test consensus is the accuracy of the span-selected cluster.
Gold answers are used only for evaluation.

\begin{table}[h]
  \centering
  \small
  \caption{GSM8K consensus accuracy (\%) for Qwen-2.5-1.5B-it.
  Ensemble and Debate are the instruction-tuned checkpoint before either training run.
  The one-exchange row is the GSM8K entry of Table~\ref{tab:main_results}.
  The independent-sample row is a single run.
  The gap row subtracts that run from the one-exchange mean.}
  \label{app:tab:indep-vs-debate-gsm8k}
  \begin{tabular}{lcccccc}
    \toprule
   \textbf{Training traces} & \textbf{Ensemble} & \textbf{Debate} & \textbf{SFT} & \textbf{DPO} & \textbf{KTO} & \textbf{GRPO} \\
    \midrule
    Independent samples & 44.5 & -- & 47.0 & \textbf{64.4} & 61.2 & 56.4 \\
    One exchange round & 44.5 & 66.0 & \(66.3_{\pm0.71}\) & \(\textbf{71.4}_{\pm0.28}\) & \(69.2_{\pm1.13}\) & \(68.9_{\pm0.14}\)\\
    Gap & -- & -- & $+19.3$ & $+7.0$ & $+8.0$ & $+12.5$  \\
    \bottomrule
  \end{tabular}
\end{table}

Table~\ref{app:tab:indep-vs-debate-gsm8k} shows a higher consensus accuracy for every objective when the training traces include the exchange round.
The largest gap is SFT, \(66.3\) against \(47\).
On the best objective the gap is \(7.0\)pp, from independent-sample DPO at \(64.4\%\) to exchange-round DPO at \(71.4\%\).
Independent-sample DPO remains \(1.6\)pp below the inference-only debate consensus of \(66.0\%\), so the exchange round is the step that puts post-training above that baseline on this model and dataset.
The comparison is one model--dataset cell, and the independent-sample entries are a single run.

\section{Confidence calibration}
\label{app:sec:platt_calibration}

Section~\ref{sec:ablations} shows that token-probability confidence is not comparable across architectures.
Answer-span confidence \(c_i^{\mathrm{span}}\) in Eq.~\eqref{eq:span-conf} is a length-normalized geometric mean of token probabilities.
It is an approximation of probability of correctness, and models with different tokenizers do not place it on a common scale.
In the heterogeneous committee the oracle sums those scores, \(w_k=\sum_{i\in\mathcal I_k} c_i^{\mathrm{span}}\).
A member whose raw scores sit higher can then dominate \(\hat a(x)\).
This section fits a separate map for each architecture and applies it before that sum, both for the inference-only oracle and for the preference partition used by DPO.

We fit a Platt map \citep{platt1999probabilistic} for each model \(m\),
\[
\tilde c
=
\sigma(a_m c^{\mathrm{span}}+b_m),
\qquad
\sigma(z)=\bigl(1+e^{-z}\bigr)^{-1}.
\]
The coefficients \((a_m,b_m)\) are an unpenalized logistic regression of gold correctness on the raw span score.
The regression is fit on the heterogeneous committee itself: five members, \(M{=}2\), one independent generation and one exchange, on GSM8K.
The official training set is shuffled with the paper data seed.
The paper training prefix is excluded.
The next \(2{,}000\) training questions are the calibration pool.
The uncalibrated committee uses the paper cap of \(256\) new tokens.
Under that cap, Qwen3-1.7B-it (Q2) often ends before a parseable answer span, so its final-round trace contributes neither a cluster member nor a span score.
For the calibration pool, the cap is raised to \(512\) new tokens.
Every member then has room to finish the reasoning and emit the claimed answer before \(c_i^{\mathrm{span}}\) is computed and before the Platt map is fit or applied.
The score in the regression is the final-round span confidence, after the exchange.
Unparseable traces are dropped.
The larger of the correct and incorrect classes is downsampled to the smaller class, and \(80\%\) of each class is used to fit \((a_m,b_m)\).
The remaining \(20\%\) is not used for the coefficients.
Gold labels on this pool supervise only the logistic regression.
They are not preference targets.
Because the fitted map is the weight inside \(w_k=\sum_i \tilde c_i\), those labels do determine which cluster becomes \(\mathcal G^+(x)\).
The same protocol is run once for each of the five architectures.
At test time, and again inside the DPO run, each member uses its own \((a_m,b_m)\).

\begin{table}[t]
\centering
\small
\caption{
GSM8K accuracy~(\%) for the heterogeneous committee, one member each from L1, Q1, Q2, P4, and G4.
Ensemble, Raw-Debate, and Raw-DPO are the uncalibrated committee reported in Table~\ref{tab:ablate_arch}.
Platt-Debate and Platt-DPO apply the per-model map before the span-confidence sum.
The map is fit on the \(2{,}000\) training questions; those questions are disjoint from the paper training set and from the test split.
Consensus is the accuracy of the selected cluster.
The other columns are per-member accuracy and their mean.
The best entry in each column is bold.
Gold answers are used to fit the maps and to compute accuracy.
}
\label{tab:platt_gsm8k}
\begin{adjustbox}{max width=\textwidth}
\begin{tabular}{lccccccc}
\toprule
\multirow{2}{*}{\textbf{Method}} &
\multicolumn{7}{c}{\textbf{Accuracy}} \\
\cmidrule(lr){2-8}
& \textbf{Consensus} & \textbf{Avg.\ Agent} & \textbf{L1} & \textbf{Q1} & \textbf{Q2}
& \textbf{P4} & \textbf{G4} \\
\midrule
Ensemble
& 78.2 & 30.9 & 3.6 & 26.0 & 1.4 & 76.8 & 46.8 \\
\midrule
Raw-Debate
& 79.2 & 36.8 & 16.6 & 57.6 & 3.2 & 71.6 & 35.2 \\
Raw-DPO
& 87.0 & 68.2 & \textbf{71.0} & 75.8 & 41.8 & 81.2 & 71.4 \\
\midrule
Platt-Debate
& 88.0 & 61.6 & 27.6 & 70.2 & 47.8 & 83.2 & 79.0 \\
Platt-DPO
& \textbf{89.6} & \textbf{75.7} & 68.0 & \textbf{80.8} & \textbf{58.2} & \textbf{85.0} & \textbf{86.4} \\
\bottomrule
\end{tabular}
\end{adjustbox}
\end{table}
Table~\ref{tab:platt_gsm8k} compares the uncalibrated heterogeneous committee with a committee that uses the fitted maps.
On consensus accuracy, Platt-Debate reaches \(88.0\%\), against \(79.2\%\) for Raw-Debate (\(+8.8_{pp}\)).
Platt-DPO reaches \(89.6\%\), against \(87.0\%\) for Raw-DPO (\(+2.6_{pp}\)).
The larger share of the consensus change is already present before post-training.
DPO then adds \(1.6_{pp}\) of consensus accuracy on the calibrated committee, compared with \(7.8_{pp}\) on the raw committee, where the inference-only baseline is lower.
The map rescales span scores after the traces have been generated, so the consensus column is the quantity it changes: a different weighting can select a different cluster from the same five answers.
%
%
Read against its own debate baseline, Platt-DPO raises mean agent accuracy from \(61.6\%\) to \(75.7\%\).
Across the two DPO rows, L1 is the exception: Raw-DPO reaches \(71.0\%\) and Platt-DPO reaches \(68.0\%\).
The calibrated committee still leaves a wide gap between consensus and the weaker members.
At Platt-Debate, consensus accuracy is \(88.0\%\) while L1 is at \(27.6\%\) and P4 is at \(83.2\%\).
Putting the five scores through separate logistic maps improves the selected answer on test split, and the members remain uneven.

\section{Zero-Shot Single-Agent Reference Performance}
\label{app:sec:zero_shot_single_model_perf}

To contextualize the performance of the local debate agents, we evaluate larger and closed-source models in a zero-shot, single-agent setting.
All runs use temperature $0$ and a maximum completion budget of $256$ tokens, matching the generation budget per response used by the local debate agents.

Table~\ref{app:tab:zero_shot_single_model_perf} separates responses into three categories: \emph{correct}, when the extracted answer matches the reference answer; \emph{wrong}, when an answer is extracted but does not match; and \emph{unparseable}, when no valid final answer can be extracted.
Unparseable responses are counted as incorrect when computing accuracy.
This breakdown distinguishes answer errors from failures to produce a parseable final answer within the prescribed budget.
The results characterize performance under this constrained generation protocol, rather than performance with unrestricted generation.

\begin{table}[h]
\centering
\small
\caption{
Zero-shot single-agent performance of models on the evaluation splits at temperature $0$ with a maximum completion budget of $256$ tokens.
Values are percentages of evaluated examples.
Accuracy is the percentage of correct responses; wrong and unparseable responses are both counted as incorrect.
}
\label{app:tab:zero_shot_single_model_perf}
\begin{tabular}{llrrr}
\toprule
\multirow{2}{*}{Model}
& \multirow{2}{*}{Dataset}
& \multicolumn{2}{c}{Incorrect (\%)}
& \multirow{2}{*}{Correct (\%)} \\
\cmidrule(lr){3-4}
& & Unparseable & Wrong & \\
\midrule
\multirow{3}{*}{GPT-4o}
& SVAMP  & 1.7  & 5.0  & 93.3   \\
& GSM8K  & 17.8 & 4.0 & 78.2 \\
& MathQA & 51.4 & 4.6 & 44.0 \\

\midrule
\multirow{3}{*}{Qwen-3.8-27B}
& SVAMP  & 55.3   & 13.7  & 31.0   \\
& GSM8K  & 87.0   & 8.6  & 14.4 \\
& MathQA & 81.0   & 0.2  & 18.8 \\
\bottomrule
\end{tabular}
\end{table}

\section{Judge Supervision}
\label{app:sec:judge_supervision}

We also studied the impact of a closed-source Judge model used for supervision for the oracle selection.
We used GPT-4o model as judge for the multi-agent tempertaure diverse committee of five Qwen2.5-1.5B-it models, debating for two rounds. 
GPT-4o does not debate. 
After the second round it reads the question and the parsed final answer, and returns a probability that the answer is correct. 
It does not see the gold label or the other agents. 
Equivalent answers form a cluster, and the cluster score is the sum of those probabilities. 
A failed judge call contributes zero.
The winning cluster supplies the DPO labels: its members are preferred, and the other parsed answers are dispreferred. 
Majority vote and the gold cluster are stored on the same generations. 
The gold cluster is the correct answer when any debater produced it.

\begin{table}[t]
\centering
\caption{Test accuracy of the selected answer before training.
Each entry is correct questions out of 500.
The gold cluster is the best answer already present in the debate.}
\label{tab:gpt4o-judge-debate}
\small
\begin{tabular}{lccc}
\toprule
Dataset & Majority & GPT-4o judge & Gold cluster \\
\midrule
GSM8K  & 65.6 & 71.6 & 72.8 \\
MathQA & 41.0 & 47.2 & 49.6 \\
\bottomrule
\end{tabular}
\end{table}

On GSM8K the judge matches the majority cluster on 451 of 500 questions.
On the other 49, the judge is correct and majority is wrong 31 times, and the reverse happens once. 
On MathQA the two rules match on 348 questions. 
The judge is the only correct rule 35 times, and majority 4 times. 
The remaining gap to the gold cluster is 6 GSM8K questions and 12 MathQA questions.

\begin{table}[t]
\centering
\caption{GSM8K test accuracy before and after DPO.
DPO trains on the GPT-4o split for 3 epochs.
The after-DPO majority column is a vote over that trained committee, not a model trained on majority labels.
}
\label{tab:gpt4o-judge-dpo}
\small
\begin{tabular}{lcc}
\toprule
Selection & Debate & After DPO \\
\midrule
Majority      & 65.6 & 71.2 \\
GPT-4o judge  & 71.6 & 78.2 \\
Gold cluster  & 72.8 & 78.6 \\
\bottomrule
\end{tabular}
\end{table}

\section{Temperature and confidence across models}
\label{app:sec:gsm8k-temperature}

The same five temperatures, $0.01$, $0.25$, $0.5$, $0.75$, and $1.0$, are used for every model in the pre-training debates.
Full-trace confidence is the exponentiated mean token log-probability of the final reply.
Boxed confidence is the same quantity on the tokens inside the final \texttt{\textbackslash boxed\{\}} span. 
Table~\ref{tab:temp-confidence-universal} reports both scores on the training questions stored for math model--dataset pairs. 
The arrow is temperature $0.01$ to temperature $1.0$. 
Trace uses every stored training reply. 
Boxed uses the replies that contain a boxed answer. 
%
%

\begin{table}[h]
\centering
\small
\caption{
Training-split confidence from temperature $0.01$ to $1.0$. 
%
}
\label{tab:temp-confidence-universal}
\begin{tabular}{llcccc}
\toprule
Model & Dataset & Trace mean & Trace median & Boxed mean & Boxed median \\
\midrule
Llama-3.2-1B   & \multirow{4}{*}{GSM8K}   & $0.926 \to 0.807$ & $0.931 \to 0.819$ & $0.989 \to 0.966$ & $1.000 \to 0.999$ \\
Qwen-2.5-1.5B   &    & $0.864 \to 0.694$ & $0.873 \to 0.718$ & $0.989 \to 0.968$ & $0.999 \to 0.999$ \\
Phi-4-mini     &   & $0.902 \to 0.798$ & $0.938 \to 0.802$ & $0.999 \to 0.996$ & $1.000 \to 1.000$ \\
Gemma-3-4B     &   & $0.880 \to 0.857$ & $0.882 \to 0.857$ & $1.000 \to 1.000$ & $1.000 \to 1.000$ \\
\midrule
Llama-3.2-1B   & \multirow{4}{*}{SVAMP}    & $0.924 \to 0.787$ & $0.931 \to 0.804$ & $0.993 \to 0.984$ & $1.000 \to 1.000$ \\
Qwen-2.5-1.5B   &     & $0.836 \to 0.659$ & $0.844 \to 0.678$ & $0.985 \to 0.974$ & $0.999 \to 0.999$ \\
Phi-4-mini     &   & $0.850 \to 0.741$ & $0.854 \to 0.747$ & $0.997 \to 0.997$ & $1.000 \to 1.000$ \\
Gemma-3-4B     &     & $0.835 \to 0.803$ & $0.825 \to 0.799$ & $1.000 \to 1.000$ & $1.000 \to 1.000$ \\
\bottomrule
\end{tabular}
\end{table}

Full-trace confidence falls with temperature in every row. 
The drop in the mean runs from $0.023$ (Gemma-3-4B, GSM8K) to $0.177$ (Qwen-2.5-1.5B, SVAMP). 
On the debates that record whether each reply is correct, an incorrect reply at temperature $0.01$ still outscores a correct reply at temperature $1.0$: Qwen-2.5-1.5B on GSM8K ($0.858$ against $0.722$) and SVAMP ($0.826$ against $0.674$), Llama-3.2-1B on GSM8K ($0.926$ against $0.790$), SVAMP ($0.924$ against $0.774$), and Llama-3.1-8B on GSM8K ($0.944$ against $0.902$). 
Raw trace scores from different temperatures are therefore not on a common scale.

On GSM8K and SVAMP the boxed median stays at $0.999$ or $1.000$ for every model.
The boxed mean moves by at most $0.023$. 
Agents can be compared on that score without a per-temperature calibration. 
%
%
%
In each of those rows the trace mean falls by more than the boxed mean.
The flat boxed scale is a property of the short numerical span.
%
%
Table~\ref{tab:gsm8k-temp-confidence} expands the Qwen-2.5-1.5B-it GSM8K training debate across all five temperatures. 
%

\begin{table}[h]
\centering
\caption{
Qwen-2.5-1.5B-it on the $1{,}500$ GSM8K training questions.
Trace uses all $1{,}500$ replies at each temperature.
}
\label{tab:gsm8k-temp-confidence}
\begin{tabular}{lccccc}
\toprule
& \multicolumn{5}{c}{Temperature} \\
\cmidrule(lr){2-6}
& $0.01$ & $0.25$ & $0.50$ & $0.75$ & $1.0$ \\
\midrule
Boxed mean    & $0.989$ & $0.989$ & $0.984$ & $0.982$ & $0.968$ \\
Boxed median  & $0.999$ & $0.999$ & $0.999$ & $0.999$ & $0.999$ \\
\midrule
Trace mean    & $0.864$ & $0.855$ & $0.834$ & $0.790$ & $0.694$ \\
Trace median  & $0.873$ & $0.865$ & $0.844$ & $0.801$ & $0.718$ \\
\bottomrule
\end{tabular}
\end{table}

From $0.01$ to $1.0$ the trace mean drops by $0.170$. 
The colder reply has the higher trace score on $95.6\%$ of the $1{,}500$ questions, and the median gap is $0.152$. 
Correct replies alone fall from $0.869$ to $0.722$. 
The boxed median gap on the $660$ questions where both endpoints box is $0.0001$. 

\section{Analysis of execution confidence}
\label{app:sec:mbpp-oracle}

Following the MBPP setup above in Appendix~\ref{app:sec:data}, each candidate program is scored on the public assertions.
Programs that emit the same sequence of public outputs form one cluster.
Majority voting selects the largest cluster.
Execution confidence selects the cluster whose members pass the most public cases, summing those passes across the agents that share the cluster.
The reported correctness check is separate from that selection.
It executes every public assertion together with any challenge assertion, and it passes only when all of them succeed~(Hidden success-rate).
The challenge list may be empty, in which case the check repeats the public assertions but remains all-or-nothing, whereas the oracle still counts how many public cases passed.
A non-empty challenge assertion can reject a program that passed every public case.
The hidden check is stricter than the public pass-rate even when the two sets of assertions coincide.

\begin{figure}[t]
  \centering
  \includegraphics[width=0.6\linewidth]{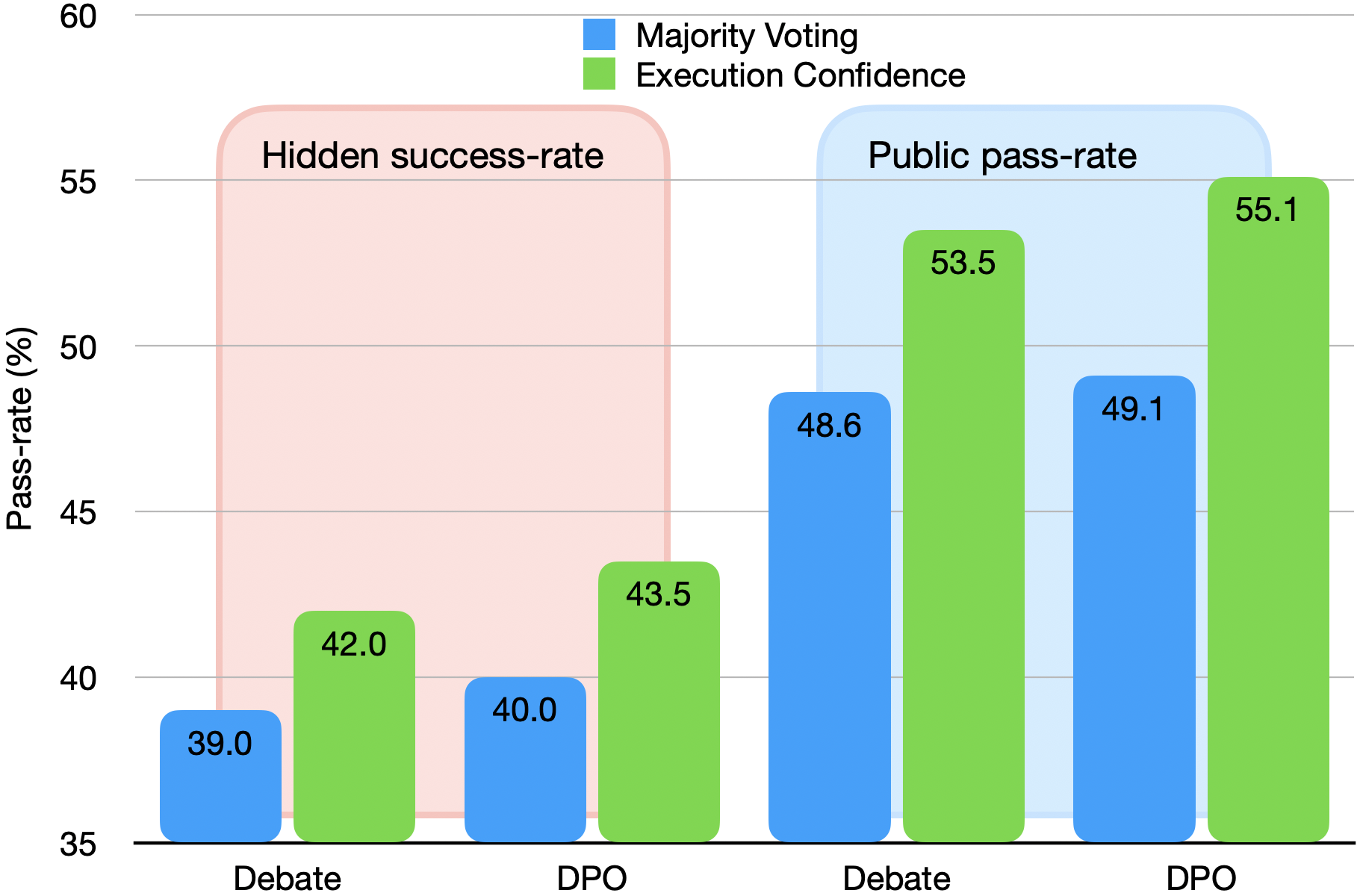}
  \caption{Pass rate of the winning cluster on MBPP.
  Majority voting (blue) is a plurality over public-output clusters.
  Execution confidence (green) keeps the cluster with the most public cases passed.
  Hidden success-rate is the all-or-nothing check over public and challenge assertions.
  Public pass-rate pools the visible cases of the selected program.
  Debate is the shared pre-training trace.
  DPO is three epochs on that selector's label.}
  \label{fig:mbpp-oracle}
\end{figure}

We train Llama-3.1-8B with DPO for three epochs on shared five-agent debate.
One run uses the majority cluster as the training label.
The other uses the execution-confidence cluster.
Figure~\ref{fig:mbpp-oracle} reports the hidden-check pass rate and the public-case pass rate of the winning cluster on the 200-question test set, before training and after DPO.
The debate bars are the two selectors applied to the same traces.
The DPO bars are each trained model scored by the selector it was trained on.

On the shared debate, execution confidence passes the hidden check on 42.0\% of questions (84/200), against 39.0\% (78/200) for the vote, and passes 53.3\% of public cases, against 48.6\%.
After DPO the same ordering holds, and the public-case gap widens.
The confidence-trained model, scored by execution confidence, reaches 43.5\% hidden (87/200) and 55.1\% public.
The vote-trained model, scored by majority voting, reaches 40.0\% hidden (80/200) and 49.1\% public.
Training on the execution-confidence label therefore improves the cluster that the oracle keeps, on both the public count that defines the label and the stricter check used for evaluation.
In this execution setting, selecting and training on how many public tests a cluster passes is the stronger rule.

\section{Tasks without a direct answer-equivalence relation}
\label{app:sec:non_equivalence_tasks}

The main experiments use tasks on which a final answer is a number, a choice letter, or a program, so two traces can be tested for equivalence and then counted.
RaR-Medicine~\citep{gunjal2026rubrics} does not have that kind of answer.
This subsection records the dataset, the checklist, the judge that turns a free-form reply of Qwen-2.5-1.5B-it into a checklist pattern, and the labeling rule that replaces a vote over those patterns.
Post-training is otherwise the same pipeline as in the main experiments: a sampling-diverse committee, followed by DPO on the preferred cluster.

\paragraph{Dataset.}
We use the medicine split of Rubrics as Rewards \citep{gunjal2026rubrics}, released as \texttt{ScaleAI/RaR-Medicine} on Huggingface.
Each example is a clinical question, a reference answer, and a checklist.
The reference answer is not a training target and is not shown to the committee.
The sweep uses 1500 training questions and 500 test questions, shuffled with the same data seed as the other datasets.
Agents are asked only to answer the question completely and generate and open-ended reply.

\paragraph{Checklist.}
A checklist is an ordered list of pass/fail criteria produced by the autorubric pipeline of \citet{gunjal2026rubrics}.
Each item has a title, a description of what the response must do, and an integer weight from the release.
An item contributes one hit when the judge marks it satisfied, and zero otherwise.
For a question with $K$ criteria, a response is summarized by a bit string $b\in\{0,1\}^K$.
The hit count is $c(b)=\sum_{k=1}^{K} b_k$, and the hit rate~(confidence) is $c(b)/K$.

\paragraph{Judgment.}
Equivalence is not computed from the text of the answer.
A frozen judge (GPT-4o at temperature 0) reads the question, the full assistant response, and the criteria in order, and returns JSON of the form \texttt{\{"satisfied": [true, false, ...]\}} with one boolean per criterion.
That boolean vector is the bit string $b$.
Two responses are the same cluster exactly when their bit strings match, so a cluster is one way of meeting the checklist, not a paraphrase class.
The same judge is used to score the committee during debate and to score a sample at evaluation time.
The reference answer is not an input to this judgment.

\paragraph{Evaluation.}
The primary number is the weighted hit rate on the 500 test questions.
For each question the judge scores the selected answer: satisfied positive criteria add their weight, satisfied negative criteria subtract, and the total is divided by the sum of the positive weights.
A missing or unparseable answer counts as zero.
Debate is the answer selected from the round-2 committee, before training.
Each DPO entry is the policy trained on that row's debate labels, then judged on its own test generations under the same selection rule.
Table~\ref{tab:rar-hit-rate} reports that comparison.
The rubric success rate is the fraction of test questions on which every positive criterion is satisfied and every negative criterion is avoided.
A method can raise the average hit rate while still rarely clearing the whole checklist, so that rate is reported separately and is not the selection rule used for training.

\begin{table}[t]
\centering
\caption{RaR-Medicine test split success rate and hit rate.
Each cell is the mean fraction of checklist items satisfied.}
\label{tab:rar-hit-rate}
\small
\begin{tabular}{lcccc}
\toprule
& \multicolumn{2}{c}{Full success rate}
& \multicolumn{2}{c}{Rubric hit rate} \\
\cmidrule(lr){2-3} \cmidrule(lr){4-5}
Oracle & Debate & DPO & Debate & DPO \\
\midrule
Majority Voting & 4.2 & 3.2 & 30.5 & 29.3 \\
Hit Confidence  & 5.6 & 6.0 & 41.9 & 42.8 \\
\bottomrule
\end{tabular}
\end{table}

\paragraph{Why a vote is the wrong label.}
Inside a cluster every member has the same bit string, so every member has the same hit count $c$.
A vote assigns the cluster the score $n$, the number of agents that landed on it.
The rubric oracle assigns each member the weight $c$ and, under the same sum reduction used in the main experiments, assigns the cluster the score $n\cdot c$.
The vote therefore keeps only how often a pattern was produced.
The hit count also keeps how much of the checklist that pattern covers.
A pattern that meets more criteria can outrank a pattern that more agents happened to repeat.
Table~\ref{tab:rar-vote-vs-hits} is the definition, not a measured outcome: three agents share a checklist with 2 of 10 items satisfied, and one agent meets 8 of 10.
The vote selects the first cluster.
The hit-count score selects the second, because $1\cdot 8 > 3\cdot 2$.
On these questions the patterns are usually all different, so a vote is a tie among singletons and the tie break is string order of the bit strings, which is unrelated to how many criteria are met.
The preferred cluster under \method{} is the one with the largest hit-count~(confidence) score.
Completions whose judged bit string matches that cluster are the preferred side of the preference pair.
The others are the rejected side.
DPO then use those pairs exactly as on the verifiable tasks.

\begin{table}[t]
\centering
\caption{Vote score and hit-count score on one question with $K{=}10$ criteria.
The vote selects the repeated partial checklist.
The hit-count score selects the checklist that meets more criteria.}
\label{tab:rar-vote-vs-hits}
\small
\begin{tabular}{lccc}
\toprule
Cluster & Agents $n$ & Criteria met $c$ & Score \\
\midrule
Vote ($n$) & 3 & 2 & 3 \\
Vote ($n$) & 1 & 8 & 1 \\
\midrule
Hit count ($n\cdot c$) & 3 & 2 & 6 \\
Hit count ($n\cdot c$) & 1 & 8 & 8 \\
\bottomrule
\end{tabular}
\end{table}

\section{Qualitative analysis}
\label{app:sec:qualitative_analysis}

The quantitative results report accuracies aggregated over prompts.
This section shows the same quantities on individual transcripts, so that the oracle rule in Section~\ref{sec:oracle} can be checked against the committee that produced it.
Each block is one selected prompt from an evaluation run.
The gold answer is printed only to interpret the transcript.
It is not an input to debate, to equivalence clustering, or to post-training.
A trace is unparseable when \(S(y_i)=\varnothing\), and it is placed in \(\mathcal G^-(x)\) rather than in a cluster, as in Section~\ref{sec:oracle}.
For Qwen-2.5-1.5B-it model on GSM8K dataset, every prompt on which majority voting and span confidence select different clusters has two clusters sharing the largest \(n_k\).
The blocks in Sections~\ref{app:sec:qual:tie} and~\ref{app:sec:qual:disagree} are prompts of that kind.

\subsection{Full responses}
\label{app:sec:qual:full}

The following blocks record question asked to the committee members and their full responses.
Round 1 is the independent answer.
Round 2 is the answer written after the other agents' round-1 answers have been shown.
One prompt ends on the gold answer.
The other ends on a shared incorrect answer after agents who had the gold answer adopt a calculation already present in the committee.

\paragraph{Convergence onto the gold answer.}

Katelyn sees 50 fairies, 25 more join, and 30 leave, so 45 remain.
The independent answers are 55, 45, 15, 85, and 45.
After the exchange every agent answers 45.
Agent 5's second trace says the revision follows the solutions given by the other agents.

\begin{tcolorbox}[
    breakable,
    colback=gray!5!white,
    colframe=black,
    title={Question asked to committee}
]
\textbf{Run.}

\begin{verbatim}
model: Qwen2.5-1.5B-it
dataset / split: GSM8K / test
example id: 392
seed: 10
committee: N=5, M=2, temperatures {0.01, 0.25, 0.5, 0.75, 1.0}
\end{verbatim}

\textbf{Question shown to the agents.}

\begin{verbatim}
While playing with her friends in their school playground, 
Katelyn saw 50 fairies flying above the nearby forest. 
After about twenty minutes, one of her friends saw half 
as many fairies as Katelyn saw come from the east and 
join the fairies that were there. Ten minutes later, 
30 fairies flew away. How many fairies are remaining?
\end{verbatim}

\textbf{Gold answer (evaluation only).}

\begin{verbatim}
45
\end{verbatim}
\end{tcolorbox}

\begin{tcolorbox}[
    breakable,
    colback=gray!5!white,
    colframe=black,
    title={Response: agent 1, round 1}
]
To solve this problem, we need to follow these steps:

\begin{enumerate}
    \item Determine how many fairies joined from the east after twenty minutes.

    \begin{itemize}
        \item Katelyn initially saw 50 fairies.
        \item Her friend saw half as many fairies as Katelyn did, so they saw
        \[
        \frac{50}{2} = 25
        \]
        fairies from the east.
    \end{itemize}

    \item Determine how many fairies remained after 30 fairies flew away.

    \begin{itemize}
        \item Ten minutes later, 30 fairies flew away.
    \end{itemize}
\end{enumerate}

So, the total number of fairies remaining is:

\[
50 - 25 + 30 = 55
\]

Therefore, the final answer is \(\boxed{55}\).
\end{tcolorbox}

\begin{tcolorbox}[
    breakable,
    colback=gray!5!white,
    colframe=black,
    title={Response: agent 2, round 1}
]
To solve this problem, let's break it down into steps:

\begin{enumerate}
    \item \textbf{Initial Count:} Katelyn initially sees 50 fairies.

    \item \textbf{Friends' Observation:}
    \begin{itemize}
        \item Her friend sees half as many fairies as Katelyn did.
        \item Calculation:
        \[
        \frac{50}{2} = 25
        \]
        fairies.
    \end{itemize}

    \item \textbf{Fairies Joining:}
    \begin{itemize}
        \item Thirty more fairies fly away after ten minutes.
        \item Calculation:
        \[
        50 + 25 - 30 = 45.
        \]
    \end{itemize}
\end{enumerate}

So, the total number of fairies remaining is 45.

\textbf{Final Answer:} \(\boxed{45}\).
\end{tcolorbox}

\begin{tcolorbox}[
    breakable,
    colback=gray!5!white,
    colframe=black,
    title={Response: agent 3, round 1}
]
Let's break down the problem step by step:

\begin{enumerate}
    \item \textbf{Initial Number of Fairies:}

    Katelyn initially saw 50 fairies.

    \item \textbf{Fairies Joining from East:}

    One friend saw half as many fairies as Katelyn did, which is
    \[
    \frac{50}{2} = 25.
    \]
    So, these 25 fairies joined from the east.

    The total now becomes:
    \begin{itemize}
        \item Katelyn: 50
        \item Friends: 25
    \end{itemize}

    \item \textbf{Fairies Leaving:}

    Ten minutes later, 30 fairies flew away. This leaves us with:
    \begin{itemize}
        \item Katelyn: \(50 - 30 = 20\)
        \item Friends: \(25 - 30 = -5\)
    \end{itemize}

    Since we cannot have negative fairies, this means all 25 fairies left.

    \item \textbf{Final Calculation:}

    Total remaining fairies:
    \[
    20 + (-5) = 15.
    \]
\end{enumerate}

Therefore, the final answer is \(\boxed{15}\).
\end{tcolorbox}

\begin{tcolorbox}[
    breakable,
    colback=gray!5!white,
    colframe=black,
    title={Response: agent 4, round 1}
]
To solve this problem, let's break it down into steps:

\begin{enumerate}
    \item \textbf{Initial Count:} At first, Katelyn sees 50 fairies.

    \item \textbf{Friends' Observation:}
    Her friend sees half as many fairies as she does. Therefore,
    \[
    \frac{50}{2} = 25
    \]
    fairies.

    \item \textbf{Fairy Migration:}
    Ten minutes after seeing her friend, 30 fairies fly away.

    \item \textbf{Final Calculation:}

    The total number of fairies initially seen is 50. Her friend then
    sees 25 more. In addition to these, 30 fairies leave after ten minutes.

    Therefore,
    \[
    50 + 25 - 30 = 85.
    \]

    This gives us 85, but since only 30 were actually left behind
    (the 25 seen plus the others minus those that flew away), ...
\end{enumerate}

The final answer is: \(\boxed{85}\).
\end{tcolorbox}

\begin{tcolorbox}[
    breakable,
    colback=gray!5!white,
    colframe=black,
    title={Response: agent 5, round 1}
]
\begin{enumerate}
    \item \textbf{Initial Count:} Initially, Katelyn sees 50 fairies
    flying above the forest.

    \item \textbf{Friends' Observation:}
    \begin{itemize}
        \item Her friend sees half as many fairies as Katelyn did.
        \item This means she observes:
        \[
        \frac{50}{2} = 25 \text{ fairies}.
        \]
    \end{itemize}

    \item \textbf{Fairies Joining:}
    Ten minutes after joining, 30 more fairies fly away.

    \item \textbf{Final Calculation:}
    \begin{itemize}
        \item Total fairies before these events:
        \[
        50 + 25 = 75.
        \]
        \item Fairies joined plus those who stayed minus those that left:
        \[
        75 - 30 = 45.
        \]
    \end{itemize}
\end{enumerate}

Therefore, the number of remaining fairies is \(\boxed{45}\).
\end{tcolorbox}

\begin{tcolorbox}[
    breakable,
    colback=gray!5!white,
    colframe=black,
    title={Response: agent 1, round 2}
]
\begin{enumerate}
    \item \textbf{Initial Count:} Katelyn initially sees 50 fairies.

    \item \textbf{Friends' Observation:}
    Her friend sees half as many fairies as Katelyn did, which is
    \[
    \frac{50}{2} = 25
    \]
    fairies.

    \item \textbf{Fairy Migration:}
    Ten minutes after seeing her friend, 30 fairies fly away.

    \item \textbf{Final Calculation:}
    Add the initial count of 50 fairies, the 25 fairies observed by her
    friend, and subtract the 30 fairies that flew away:
    \[
    50 + 25 - 30 = 45.
    \]
\end{enumerate}

Therefore, the final answer is \(\boxed{45}\).
\end{tcolorbox}

\begin{tcolorbox}[
    breakable,
    colback=gray!5!white,
    colframe=black,
    title={Response: agent 2, round 2}
]
\begin{enumerate}
    \item \textbf{Initial Count:} Katelyn initially sees 50 fairies.

    \item \textbf{Friends' Observation:}
    Her friend sees half as many fairies as Katelyn did. This means she sees
    \[
    \frac{50}{2} = 25
    \]
    fairies.

    \item \textbf{Fairies Joining:}
    These 25 fairies join the 50 fairies already present.

    \item \textbf{Total Before Departure:}
    \[
    50 + 25 = 75
    \]
    fairies.

    \item \textbf{Departure:}
    Ten minutes later, 30 fairies fly away.

    \item \textbf{Remaining Fairies:}
    \[
    75 - 30 = 45.
    \]
\end{enumerate}

Therefore, the final answer is \(\boxed{45}\).
\end{tcolorbox}

\begin{tcolorbox}[
    breakable,
    colback=gray!5!white,
    colframe=black,
    title={Response: agent 3, round 2}
]
To find out how many fairies are remaining, follow these steps:

\begin{enumerate}
    \item \textbf{Initial Count:} Katelyn initially sees 50 fairies.

    \item \textbf{Friends' Observation:}
    Her friend sees half as many fairies as Katelyn did. This means she observes:
    \[
    \frac{50}{2} = 25 \text{ fairies}.
    \]

    \item \textbf{Fairy Migration:}
    Ten minutes after joining, 30 fairies fly away.

    \item \textbf{Final Calculation:}
    \begin{itemize}
        \item Total fairies before these events:
        \[
        50 + 25 = 75.
        \]
        \item Fairies joined plus those who stayed minus those that left:
        \[
        75 - 30 = 45.
        \]
    \end{itemize}
\end{enumerate}

Therefore, the number of remaining fairies is \(\boxed{45}\).
\end{tcolorbox}

\begin{tcolorbox}[
    breakable,
    colback=gray!5!white,
    colframe=black,
    title={Response: agent 4, round 2}
]
\begin{itemize}
    \item Initial count: Katelyn sees 50 fairies.

    \item Friend observation: Katelyn's friend sees half as many fairies
    as Katelyn did, which is 25 fairies. These 25 fairies join from the east.

    \item Remaining fairies after the friend joins:
    Add 25 (friend's fairies) to 50 (initial Katelyn's fairies), giving
    a total of 75 fairies.

    \item Fairies leaving: Ten minutes later, 30 fairies fly away.

    \item Final count: Subtract 30 from the current total of 75 fairies,
    resulting in 45 fairies remaining.
\end{itemize}

Thus, the final answer is \(\boxed{45}\).
\end{tcolorbox}

\begin{tcolorbox}[
    breakable,
    colback=gray!5!white,
    colframe=black,
    title={Response: agent 5, round 2}
]
Here are the revised step-by-step solutions provided by other agents:

\begin{enumerate}
    \item Katelyn initially sees 50 fairies.

    \item Her friend sees half as many fairies as Katelyn, which is
    \[
    \frac{50}{2} = 25
    \]
    fairies.

    \item These 25 fairies join with the 50 already present, totaling
    \[
    50 + 25 = 75
    \]
    fairies.

    \item After ten minutes, 30 fairies fly away.

    \item Adding the newly arrived fairies to those still in place:
    \[
    75 - 30 = 45.
    \]
\end{enumerate}

The final answer is \(\boxed{45}\).
\end{tcolorbox}

\paragraph{Convergence onto a shared incorrect answer.}
Violetta has 20 dollars and needs five crayons at 2 dollars each, so the change is 10 dollars.
Two independent traces reach the decoding limit before a boxed answer.
Of the three that finish, one answers 0 and two answer 10.
The answer 0 comes from buying ten crayons and spending the whole 20 dollars.
After the exchange, four agents answer 0, including both agents whose independent answer was 10.
One agent moves from 0 to 10.

\begin{tcolorbox}[colback=gray!5!white, colframe=black, title={Question asked to committee}]
\textbf{Run.}
\begin{verbatim}
model: Qwen2.5-1.5B-it
dataset / split: GSM8K / test
example id: 237
seed: 10
committee: N=5, M=2, temperatures {0.01, 0.25, 0.5, 0.75, 1.0}
\end{verbatim}

\textbf{Question shown to the agents.}
\begin{verbatim}
Violetta wants to buy new crayons. 
She needs them in 5 different colors and 
prepared \$20 for this purchase. 
One crayon costs \$2. 
How much change will she get?
\end{verbatim}

\textbf{Gold answer (evaluation only).}
\begin{verbatim}
10
\end{verbatim}
\end{tcolorbox}

\begin{tcolorbox}[
    breakable,
    colback=gray!5!white,
    colframe=black,
    title={Response: agent 1, round 1}
]
To solve this problem, let's break it down into steps:

\begin{enumerate}
    \item \textbf{Determine the total cost of crayons Violetta can afford with her \$20:}
    \begin{itemize}
        \item Each crayon costs \$2.
        \item Violetta has \$20.
    \end{itemize}

    \item \textbf{Calculate how many crayons she can buy:}

    To find out how many crayons she can buy, divide the total amount
    of money (\$20) by the price per crayon (\$2).

    \[
    \text{Number of crayons}
    = \frac{\$20}{\$2\text{ per crayon}}
    = 10\text{ crayons}
    \]

    \item \textbf{Calculate the total cost of these crayons:}

    Since each crayon costs \$2 and she buys 10 crayons,

    \[
    \text{Total cost}
    = 10\text{ crayons} \times \$2
    = \$20
    \]

    \item \textbf{Calculate the change she gets back from her initial \$20:}

    Subtract the total cost of the crayons from the amount she initially had.

    \[
    \text{Change} = \$2
    \]
\end{enumerate}
\end{tcolorbox}

\begin{tcolorbox}[
    breakable,
    colback=gray!5!white,
    colframe=black,
    title={Response: agent 2, round 1}
]
To solve this problem, let's break it down into steps:

\begin{enumerate}
    \item \textbf{Determine the total cost of crayons Violetta can afford:}
    \begin{itemize}
        \item Violetta has \$20.
        \item Each crayon costs \$2.
    \end{itemize}

    \item \textbf{Calculate how many crayons she can buy with her money:}

    To find out how many crayons she can buy, divide her total amount
    (\$20) by the price per crayon (\$2).

    \[
    \text{Number of crayons}
    = \frac{\$20}{\$2}
    = 10
    \]

    \item \textbf{Verify that she can afford all five colors:}
    \begin{itemize}
        \item Since she can buy 10 crayons and there are only 5 different
        colors, she can indeed afford all the colors she wants.
    \end{itemize}

    \item \textbf{Calculate the total cost of these crayons:}

    Multiply the number of crayons (10) by the cost per crayon (\$2).

    \[
    \text{Total cost} = 10 \times \$2 = \$20
    \]

    \item \textbf{Determine the change she will receive:}

    Subtract the total cost from the amount she initially had (\$20).
\end{enumerate}
\end{tcolorbox}

\begin{tcolorbox}[
    breakable,
    colback=gray!5!white,
    colframe=black,
    title={Response: agent 3, round 1}
]
\begin{enumerate}
    \item Determine how many crayons Violetta can buy with her \$20 if each
    crayon costs \$2.

    To find out how many crayons she can afford, divide her total money
    (\$20) by the cost per crayon (\$2).

    \[
    \text{Number of crayons}
    = \frac{\$20}{\$2}
    = 10\text{ crayons}
    \]

    \item Calculate the total amount spent on the crayons.

    Since each crayon costs \$2 and Violetta buys 10 crayons, the total
    cost is:

    \[
    \text{Total cost}
    = 10 \times \$2
    = \$20
    \]

    \item Subtract the total cost from the initial amount Violetta has
    to determine her remaining balance.

    If Violetta started with \$20 and spends all \$20 on crayons, she
    will have no change left.

    \[
    \text{Change} = \$20 - (\$20)
    \]

    \[
    \text{Change} = \$0
    \]
\end{enumerate}

Therefore, Violetta will get $\boxed{0}$ as her change.
\end{tcolorbox}

\begin{tcolorbox}[
    breakable,
    colback=gray!5!white,
    colframe=black,
    title={Response: agent 4, round 1}
]
To determine how much change Violetta will receive after her purchase,
follow these steps:

\begin{enumerate}
    \item \textbf{Calculate the total cost of the crayons Violetta buys:}

    \begin{itemize}
        \item Each crayon costs \$2.
        \item Violetta is buying 5 different colors.
    \end{itemize}

    \[
    2\text{ dollars} \times 5 = 10\text{ dollars}
    \]

    So, the total amount spent on crayons is \$10.

    \item \textbf{Subtract the total amount spent from the initial amount
    of money Violetta has:}

    \[
    \$20 - \$10 = \$10
    \]
\end{enumerate}

Thus, Violetta will have \$10 left as change after making her purchase.

\textbf{Final answer:} \(\boxed{10}\)
\end{tcolorbox}

\begin{tcolorbox}[
    breakable,
    colback=gray!5!white,
    colframe=black,
    title={Response: agent 5, round 1}
]
\begin{enumerate}
    \item \textbf{Calculate the total cost of buying one crayon:}
    \begin{itemize}
        \item One crayon costs \$2.
    \end{itemize}

    \item \textbf{Determine the total cost if Violetta buys five crayons:}
    \begin{itemize}
        \item Total = Number of crayons $\times$ Cost per crayon.
        \item Total = 5 crayons $\times$ \$2/crayon.
        \item Total = \$10.
    \end{itemize}

    \item \textbf{Subtract the total cost from the amount Violetta initially
    prepared:}
    \begin{itemize}
        \item Amount Violetta prepared = \$20.
        \item Change received = Initial amount - Total cost.
        \item Change received = \$20 - \$10.
    \end{itemize}

    \item \textbf{Therefore, the amount Violetta gets as change is:}

    \[
    \boxed{10}
    \]
\end{enumerate}
\end{tcolorbox}

\begin{tcolorbox}[
    breakable,
    colback=gray!5!white,
    colframe=black,
    title={Response: agent 1, round 2}
]
\begin{itemize}
    \item Calculate the total cost of crayons Violetta can afford with her \$20
    if each crayon costs \$2.
    \begin{itemize}
        \item Divide her total money (\$20) by the cost per crayon (\$2):
        \[
        \text{Number of crayons}
        = \frac{\$20}{\$2}
        = 10\text{ crayons}
        \]
    \end{itemize}

    \item Verify that she can afford all five colors.
    \begin{itemize}
        \item Since she can buy 10 crayons and there are only 5 different
        colors, she can indeed afford all the colors she wants.
    \end{itemize}

    \item Calculate the total amount spent on the crayons.
    \begin{itemize}
        \item Multiply the number of crayons (10) by the cost per crayon (\$2):
        \[
        \text{Total cost} = 10 \times \$2 = \$20
        \]
    \end{itemize}

    \item Determine the change she will receive.
    \begin{itemize}
        \item Subtract the total cost from the initial amount Violetta has:
        \[
        \text{Change} = \$20 - \$20 = \$0
        \]
    \end{itemize}
\end{itemize}

\textbf{Final Answer:} Violetta will get $\boxed{0}$ as her change.
\end{tcolorbox}

\begin{tcolorbox}[
    breakable,
    colback=gray!5!white,
    colframe=black,
    title={Response: agent 2, round 2}
]
\begin{itemize}
    \item Calculate the total cost of crayons Violetta can afford with her \$20.
    \begin{itemize}
        \item Each crayon costs \$2.
        \item Violetta has \$20.
    \end{itemize}

    \item Determine how many crayons she can buy.
    \begin{itemize}
        \item Divide the total amount of money (\$20) by the price per crayon (\$2).
        \[
        \text{Number of crayons}
        = \frac{\$20}{\$2\text{ per crayon}}
        = 10\text{ crayons}
        \]
    \end{itemize}

    \item Calculate the total cost of these crayons.
    \begin{itemize}
        \item Since each crayon costs \$2 and Violetta buys 10 crayons,
        \[
        \text{Total cost}
        = 10\text{ crayons} \times \$2
        = \$20
        \]
    \end{itemize}

    \item Calculate the change she will receive.
    \begin{itemize}
        \item Subtract the total cost from the amount Violetta initially has.
        \[
        \text{Change} = \$20 - \$20 = \$0
        \]
    \end{itemize}
\end{itemize}

Therefore, Violetta will get \(\boxed{0}\) as her change.
\end{tcolorbox}

\begin{tcolorbox}[
    breakable,
    colback=gray!5!white,
    colframe=black,
    title={Response: agent 3, round 2}
]
\begin{enumerate}
    \item Calculate the total cost of one crayon: Each crayon costs \$2.

    \item Determine the total cost for 5 different colors:
    \[
    5\text{ crayons} \times \$2/\text{crayon} = \$10
    \]

    \item Subtract the total cost from the amount Violetta initially prepared:
    \[
    \$20 - \$10 = \$10
    \]
\end{enumerate}

\textbf{Final Answer:} \(\boxed{10}\)
\end{tcolorbox}

\begin{tcolorbox}[
    breakable,
    colback=gray!5!white,
    colframe=black,
    title={Response: agent 4, round 2}
]
\textbf{Step-by-Step Reasoning Summary}

\begin{enumerate}
    \item \textbf{Determine the total cost of crayons Violetta can afford
    with her \$20:}
    \begin{itemize}
        \item Each crayon costs \$2.
        \item Violetta has \$20.
        \item The total number of crayons Violetta can buy is calculated as:
        \[
        \text{Number of crayons}
        = \frac{\$20}{\$2\text{ per crayon}}
        = 10\text{ crayons}
        \]
    \end{itemize}

    \item \textbf{Calculate the total cost of these crayons:}

    Since each crayon costs \$2 and Violetta buys 10 crayons,

    \[
    \text{Total cost}
    = 10\text{ crayons} \times \$2
    = \$20
    \]

    \item \textbf{Calculate the change she gets back from her initial \$20:}

    Subtract the total cost of the crayons from the amount she initially had.

    \[
    \text{Change} = \$20 - \$20 = \$0
    \]
\end{enumerate}

\textbf{Final Answer}

\[
\boxed{0}
\]
\end{tcolorbox}

\begin{tcolorbox}[
    breakable,
    colback=gray!5!white,
    colframe=black,
    title={Response: agent 5, round 2}
]
\begin{itemize}
    \item Determine the total cost of crayons Violetta can afford:
    Divide the total money by the price per crayon.

    \[
    \text{Number of crayons}
    = \frac{\$20}{\$2\text{ per crayon}}
    = 10\text{ crayons}
    \]

    \item Calculate the total cost: Multiply the number of crayons by
    the price per crayon.

    \[
    \text{Total cost}
    = 10\text{ crayons} \times \$2
    = \$20
    \]

    \item Calculate the change: Subtract the total cost from the initial
    amount Violetta had.

    \[
    \text{Change} = \$20 - \$20 = \$0
    \]
\end{itemize}

The answer is \(\boxed{0}\).
\end{tcolorbox}

\subsection{Count tie resolved by span confidence}
\label{app:sec:qual:tie}

Majority voting selects \(\operatorname*{arg\,max}_k n_k\).
When two or more clusters share that maximum, vote count does not name a winner, and the span-confidence oracle uses \(w_k\), with cluster size only as a secondary key.
The prompt below is one on which the largest \(n_k\) is shared and the two rules select different clusters.
Table~\ref{tab:oracle} summarizes how often this happens; the block below is one such example.

\begin{tcolorbox}[
    breakable,
    colback=gray!5!white,
    colframe=black,
    title={Question asked to committee}
]
\textbf{Run.}
\begin{verbatim}
model: Qwen2.5-1.5B-it
dataset / split: GSM8K / test
example id: 19
seed: 10
committee: N=5, M=2, temperatures {0.01, 0.25, 0.5, 0.75, 1.0}
\end{verbatim}

\textbf{Question shown to the agents.}
\begin{verbatim}
In Mr. Roper's class of 30 students, 20% of the class are
football players. Out of the remaining class, 25% of the
students are cheerleaders or part of band. These 3 groups
of students will need to leave early today to travel to an
away game. How many students are leaving early?
\end{verbatim}

\textbf{Gold answer (evaluation only).}
\begin{verbatim}
12
\end{verbatim}
\end{tcolorbox}

\begin{tcolorbox}[
    breakable,
    colback=gray!5!white,
    colframe=black,
    title={Count tie response analysis}]
\textbf{Final-round traces.}
\begin{verbatim}
agent   temp   parsed answer a_i    c_i^span    parseable
1       0.01   24                   0.9906      yes
2       0.25   12                   0.9996      yes
3       0.50   12                   0.9987      yes
4       0.75   6                    0.2426      yes
5       1.00   6                    0.8191      yes
\end{verbatim}

\textbf{Equivalence clusters.}
\begin{verbatim}
cluster   representative   n_k   w_k     matches gold
A         24               1     0.9906  no
B         12               2     1.9983  yes
C         6                2     1.0617  no
\end{verbatim}

\textbf{Selection.}
\begin{verbatim}
majority cluster: C
  (B and C share n_k=2; the string tie-break selects 6)
span-confidence cluster k*: B
tie broken by w_k: yes
alpha(x): 0.40
winning cluster correct: yes
\end{verbatim}
\end{tcolorbox}

\subsection{Majority voting and span confidence disagree}
\label{app:sec:qual:disagree}

On most prompts the two rules select the same cluster (Table~\ref{app:tab:oracle_comparison}).
The block below is a prompt on which they do not.
The two prompts below are one case in which the span-selected cluster is correct and one case in which the majority-selected cluster is correct.

\begin{tcolorbox}[
    breakable,
    colback=gray!5!white,
    colframe=black,
    title={Question asked to committee}
]
\textbf{Run.}
\begin{verbatim}
model: Qwen2.5-1.5B-it
dataset / split: GSM8K / test
example id: 308
seed: 10
committee: N=5, M=2, temperatures {0.01, 0.25, 0.5, 0.75, 1.0}
\end{verbatim}

\textbf{Question shown to the agents.}
\begin{verbatim}
Jack decides to visit a museum 150 miles from home. He
drives 75 mph there and back. He spends 6 hours at the
museum. How long is he gone from home?
\end{verbatim}

\textbf{Gold answer (evaluation only).}
\begin{verbatim}
10
\end{verbatim}

\end{tcolorbox}

\begin{tcolorbox}[
    breakable,
    colback=gray!5!white,
    colframe=black,
    title={Disagreement: span-selected cluster is correct}]

\textbf{Final-round traces.}
\begin{verbatim}
agent   temp   parsed answer a_i    c_i^span    parseable
1       0.01   480                  0.9963      yes
2       0.25   10                   0.9992      yes
3       0.50   8                    0.9992      yes
4       0.75   8                    0.1139      yes
5       1.00   10                   0.9993      yes
\end{verbatim}

\textbf{Equivalence clusters.}
\begin{verbatim}
cluster   representative   n_k   w_k     matches gold
A         480              1     0.9963  no
B         10               2     1.9985  yes
C         8                2     1.1131  no
\end{verbatim}

\textbf{Selection.}
\begin{verbatim}
majority cluster / correct?: C (8) / no
  (B and C share n_k=2; the string tie-break selects 8)
span-confidence cluster / correct?: B (10) / yes
alpha(x): 0.40
\end{verbatim}
\end{tcolorbox}

\begin{tcolorbox}[
    breakable,
    colback=gray!5!white,
    colframe=black,
    title={Question asked to committee}
]
\textbf{Run.}
\begin{verbatim}
model: Qwen2.5-1.5B-it
dataset / split: GSM8K / test
example id: 328
seed: 10
committee: N=5, M=2, temperatures {0.01, 0.25, 0.5, 0.75, 1.0}
\end{verbatim}

\textbf{Question shown to the agents.}
\begin{verbatim}
Janet had 22 green pens and 10 yellow pens. Then she
bought 6 bags of blue pens and 2 bags of red pens. There
were 9 pens in each bag of blue and 6 pens in each bag of
red. How many pens does Janet have now?
\end{verbatim}

\textbf{Gold answer (evaluation only).}
\begin{verbatim}
98
\end{verbatim}
\end{tcolorbox}

\begin{tcolorbox}[
    breakable,
    colback=gray!5!white,
    colframe=black,
    title={Disagreement: majority-selected cluster is correct}]
\textbf{Final-round traces.}
\begin{verbatim}
agent   temp   parsed answer a_i    c_i^span    parseable
1       0.01   98                   0.9988      yes
2       0.25   98                   0.9986      yes
3       0.50   100                  0.9996      yes
4       0.75   100                  0.9998      yes
5       1.00   unparseable          na          no
\end{verbatim}

\textbf{Equivalence clusters.}
\begin{verbatim}
cluster   representative   n_k   w_k     matches gold
A         98               2     1.9975  yes
B         100              2     1.9995  no
\end{verbatim}

\textbf{Selection.}
\begin{verbatim}
majority cluster / correct?: A (98) / yes
  (A and B share n_k=2; the string tie-break selects 98)
span-confidence cluster / correct?: B (100) / no
alpha(x): 0.40
\end{verbatim}
\end{tcolorbox}

\subsection{Later rounds raise agreement without raising correctness}
\label{app:sec:qual:rounds}

Section~\ref{sec:committee} fixes \(M{=}2\) because further exchange rounds increase agreement while consensus accuracy stops improving and can fall.
The block below follows one prompt across rounds.
The rows below are the parsed answer and \(c_i^{\mathrm{span}}\) after the independent generation and after each exchange, together with the selected cluster and whether it matches the gold answer.

\begin{tcolorbox}[
    breakable,
    colback=gray!5!white,
    colframe=black,
    title={Question asked to committee}
]
\textbf{Run.}
\begin{verbatim}
model: Qwen2.5-1.5B-it
dataset / split: GSM8K / test
example id: 324
seed: 10
committee: N=5 clones, temperature 0.5
rounds recorded: 1 (independent), 2, 3, 4, 5
\end{verbatim}

\textbf{Question shown to the agents.}
\begin{verbatim}
Christina is planning a birthday party and needs .75 gift
bags per invited guest, because 1/4 of attendees don't
show up. She invited 16 friends. Gift bags are $2 each.
How much will she spend?
\end{verbatim}

\textbf{Gold answer (evaluation only).}
\begin{verbatim}
24
\end{verbatim}

\end{tcolorbox}

\begin{tcolorbox}[
    breakable,
    colback=gray!5!white,
    colframe=black,
    title={Agreement rises, selected answer stays incorrect}]

\textbf{Committee by round.}
\begin{verbatim}
round   agent   parsed answer a_i    c_i^span    parseable
1       1       unparseable          na          no
1       2       24                   0.9994      yes
1       3       18                   0.9995      yes
1       4       24                   0.9981      yes
1       5       18                   0.9995      yes
2       1       18                   0.9996      yes
2       2       18                   0.9996      yes
2       3       18                   0.9992      yes
2       4       24                   0.9997      yes
2       5       18                   0.9975      yes
3       1       18                   0.9991      yes
3       2       24                   0.9990      yes
3       3       18                   0.9998      yes
3       4       18                   0.9998      yes
3       5       18                   0.9998      yes
4       1       18                   0.9994      yes
4       2       18                   0.9998      yes
4       3       18                   0.9997      yes
4       4       18                   0.9998      yes
4       5       18                   0.9998      yes
5       1       18                   0.9997      yes
5       2       18                   0.9998      yes
5       3       18                   0.9998      yes
5       4       18                   0.9999      yes
5       5       18                   0.9998      yes
\end{verbatim}

\textbf{Selected cluster by round.}
The selected answer is the span-confidence cluster.
\begin{verbatim}
round   n_max   w_of_selected   alpha(x)   consensus   correct
1       2       1.9990          0.40          18         no
2       4       3.9959          0.80          18         no
3       4       3.9986          0.80          18         no
4       5       4.9985          1.00          18         no
5       5       4.9990          1.00          18         no
\end{verbatim}
\end{tcolorbox}

\subsection{Unanimous incorrect debate}
\label{app:sec:qual:unanimous}
Preference objectives~(DPO \& KTO) use a prompt only when both \(\mathcal G^+(x)\) and \(\mathcal G^-(x)\) are nonempty (Section~\ref{sec:training}).
A debate in which every parseable agent names the same incorrect answer has \(\mathcal G^-(x)=\emptyset\) once unparseable traces are set aside, or a \(\mathcal G^-(x)\) that contains only unparseable traces.
Such a prompt does not supply a competing-cluster comparison.
The block documents one prompt of this kind.
It is the case in which agreement is high and the selected answer is wrong, which is the regime where \(\alpha(x)\) is large and correctness is not.

\begin{tcolorbox}[
    breakable,
    colback=gray!5!white,
    colframe=black,
    title={Question asked to committee}
]
\textbf{Run.}
\begin{verbatim}
model: Qwen2.5-1.5B-it
dataset / split: GSM8K / test
example id: 153
seed: 10
committee: N=5, M=2, temperatures {0.01, 0.25, 0.5, 0.75, 1.0}
\end{verbatim}

\textbf{Question shown to the agents.}
\begin{verbatim}
Sandy wants to lose as much weight as Joey does but needs
4 weeks to lose the same amount of weight that Joey loses
in a single week. If Joey loses 8 pounds in 4 weeks, how
many weeks will it take Sandy to lose the same amount of
weight?
\end{verbatim}

\textbf{Gold answer (evaluation only).}
\begin{verbatim}
16
\end{verbatim}

\end{tcolorbox}

\begin{tcolorbox}[
    breakable,
    colback=gray!5!white,
    colframe=black,
    title={Unanimous incorrect committee}]

\textbf{Final-round traces.}
\begin{verbatim}
agent   temp   parsed answer a_i    c_i^span    parseable
1       0.01   4                    0.9988      yes
2       0.25   4                    0.9985      yes
3       0.50   4                    0.9988      yes
4       0.75   4                    0.9978      yes
5       1.00   4                    0.9991      yes
\end{verbatim}

\textbf{Preference partition.}
\(\hat a(x)\) is the span-confidence answer. Every parseable trace names it, so there is no competing cluster.
\begin{verbatim}
hat a(x): 4
|G+(x)|: 5
|G-(x)|: 0
unparseable traces in G-(x): 0
included in D_pref: no
alpha(x): 1.00
selected answer correct: no
\end{verbatim}
\end{tcolorbox}

\subsection{The same prompt before and after post-training}
\label{app:sec:qual:posttrain}
Consensus accuracy after post-training is still computed by running the committee and applying the oracle.
The block below uses one test prompt under the instruction-tuned checkpoint and under the post-trained checkpoint, with the same \(N\), \(M\), temperatures, and decoding limit.
Record pre-debate agent answers as well as final-round answers, matching the three metrics in Section~\ref{sec:experimental_setup}.

\begin{tcolorbox}[
    breakable,
    colback=gray!5!white,
    colframe=black,
    title={Question asked to committee}
]
\textbf{Run.}
\begin{verbatim}
model: Qwen2.5-1.5B-it
dataset / split: GSM8K / test
example id: 430
seed: 10
objective: DPO
checkpoint: instruction-tuned vs. one QLoRA iteration
committee: N=5, M=2, temperatures {0.01, 0.25, 0.5, 0.75, 1.0}
\end{verbatim}

\textbf{Question shown to the agents.}
\begin{verbatim}
Tara bought 8 packs of 5 canvas bags for $4 each. She
painted them and sold them at a craft fair for $8 each.
How much profit did she earn on her bags?
\end{verbatim}

\textbf{Gold answer (evaluation only).}
\begin{verbatim}
160
\end{verbatim}

\end{tcolorbox}

\begin{tcolorbox}[
    breakable,
    colback=gray!5!white,
    colframe=black,
    title={Before and after post-training}]

\textbf{Instruction-tuned checkpoint.}
\begin{verbatim}
round   agent   temp   parsed answer a_i    c_i^span
1       1       0.01   32                   0.9981
1       2       0.25   32                   0.9979
1       3       0.50   608                  0.9975
1       4       0.75   8                    0.9965
1       5       1.00   32                   0.9980
2       1       0.01   32                   0.9996
2       2       0.25   32                   0.9991
2       3       0.50   32                   0.9998
2       4       0.75   32                   0.9998
2       5       1.00   608                  0.9992
selected cluster: 32
consensus correct: no
mean agent correct (final round): 0/5
\end{verbatim}

\textbf{Post-trained checkpoint.}
\begin{verbatim}
round   agent   temp   parsed answer a_i    c_i^span
1       1       0.01   32                   0.9983
1       2       0.25   288                  0.9964
1       3       0.50   32                   0.9993
1       4       0.75   288                  0.9981
1       5       1.00   160                  0.9973
2       1       0.01   160                  0.9995
2       2       0.25   160                  0.9997
2       3       0.50   160                  0.9998
2       4       0.75   unparseable          na
2       5       1.00   32                   0.9991
selected cluster: 160
consensus correct: yes
mean agent correct (final round): 3/5
\end{verbatim}
\end{tcolorbox}

\subsection{Heterogeneous committee}
\label{app:sec:qual:heterogeneous}

In the heterogeneous committee of Section~\ref{sec:ablations}, span confidence is the geometric mean of token probabilities from different tokenizers and different scales.
Those scores are summed as if they were on one scale.
The first prompt is one on which the winning cluster is carried by one model's confidence mass.
The second prompt is one on which a lower-confidence model has the correct answer.
Per-agent accuracies for this committee are in Table~\ref{tab:ablate_arch}.

\begin{tcolorbox}[
    breakable,
    colback=gray!5!white,
    colframe=black,
    title={Question asked to committee}
]
\textbf{Run.}
\begin{verbatim}
dataset / split: GSM8K / test
example id: 288
seed: 10
members: L1=Llama-3.2-1B, Q1=Qwen2.5-1.5B,
         Q2=Qwen-3-1.7B, P4=Phi-4-mini, G4=Gemma-3-4B
temperature 0.5, M=2
\end{verbatim}

\textbf{Question shown to the agents.}
\begin{verbatim}
Jason was told he could earn $3.00 for doing his laundry,
$1.50 for cleaning his room, $0.75 for taking the trash
to the curb each week and $0.50 for emptying the
dishwasher. In a two week period, Jason emptied the
dishwasher 6 times, did his laundry once, took the trash
out twice and cleaned his room once. How much money did
Jason earn?
\end{verbatim}

\textbf{Gold answer (evaluation only).}
\begin{verbatim}
9
\end{verbatim}
\end{tcolorbox}

\begin{tcolorbox}[
    breakable,
    colback=gray!5!white,
    colframe=black,
    title={Heterogeneous scores are not on one scale}]

\textbf{Final-round traces.}
\begin{verbatim}
member   parsed answer a_i    c_i^span    correct
L1       8.00                 0.9773      no
Q1       unparseable          na          no
Q2       unparseable          na          no
P4       9.00                 0.9056      yes
G4       unparseable          na          no
\end{verbatim}

\textbf{Clusters and selection.}
Both parseable answers have \(n_k=1\). The winning mass is L1's score alone, and it exceeds P4's score on the gold answer.
\begin{verbatim}
cluster   members   representative   n_k   w_k     matches gold
A         L1        8.00             1     0.9773  no
B         P4        9.00             1     0.9056  yes

selected by span confidence: A
member with the largest c_i^span: L1
that member correct: no
\end{verbatim}
\end{tcolorbox}

\begin{tcolorbox}[
    breakable,
    colback=gray!5!white,
    colframe=black,
    title={Question asked to committee}
]
\textbf{Run.}
\begin{verbatim}
dataset / split: GSM8K / test
example id: 106
seed: 10
members: L1=Llama-3.2-1B, Q1=Qwen3-1.7B,
         Q2=Qwen2.5-1.5B, P4=Phi-4-mini, G4=Gemma-3-4B
temperature 0.5, M=2
\end{verbatim}

\textbf{Question shown to the agents.}
\begin{verbatim}
Jackson is planting tulips. He can fit 6 red tulips in a
row and 8 blue tulips in a row. If Jackson buys 36 red
tulips and 24 blue tulips, how many rows of flowers will
he plant?
\end{verbatim}

\textbf{Gold answer (evaluation only).}
\begin{verbatim}
9
\end{verbatim}

\end{tcolorbox}

\begin{tcolorbox}[
    breakable,
    colback=gray!5!white,
    colframe=black,
    title={Lower-confidence member holds the correct answer}]

\textbf{Final-round traces.}
\begin{verbatim}
member   parsed answer a_i    c_i^span    correct
L1       3                    0.9996      no
Q1       unparseable          na          no
Q2       3                    0.9903      no
P4       9                    0.9967      yes
G4       unparseable          na          no
\end{verbatim}

\textbf{Clusters and selection.}
P4 has the gold answer and a lower span confidence than L1. The wrong cluster wins because the two wrong scores sum to more than P4's score.
\begin{verbatim}
cluster   members   representative   n_k   w_k     matches gold
A         Q2, L1    3                2     1.9899  no
B         P4        9                1     0.9967  yes
selected by span confidence: A
member with the largest c_i^span: L1
that member correct: no
\end{verbatim}
\end{tcolorbox}

\subsection{MBPP example}
\label{app:sec:mbpp_example}

The following example shows a MBPP task trace generated using Llama-3.1-8B model committee.
The committee must compute sum of positive integers down to the last positive term. 
There are three public tests given unit-testing the program.
Before DPO, three agents agree on a program that starts at $n-2$ and passes none of these tests. 
Execution confidence keeps the single program that starts at $n$ and passes all three. 
After DPO on the respective oracle chosen the final responses are shown below. 
A majority of three again passes $0/3$ tests, while the confidence selection passes $3/3$.
The vote follows the larger cluster. 
The execution-confidence oracle follows the cluster that passes the most public tests.

\begin{tcolorbox}[ breakable, colback=gray!5!white, colframe=black, title={Execution confidence versus majority voting on MBPP} ] 
\textbf{Run.} 
\begin{verbatim} 
model: Llama-3.1-8B 
dataset / split: MBPP / test 
example id: 162 
seed: 10 
objective: DPO on execution confidence (public cases passed) 
checkpoint: instruction-tuned debate vs. one DPO iteration 
committee: N=5, M=2
temperatures {0.01, 0.25, 0.5, 0.75, 1.0} 
\end{verbatim}

\textbf{Question shown to the agents.} 
\begin{verbatim} 
Write a function to calculate the sum of the positive 
integers of n+(n-2)+(n-4)... (until n-x =< 0).

Implement the function sum_series. 
\end{verbatim}

\textbf{Public tests (oracle and reported pass rate).} 
\begin{verbatim} 
sum_series(6) == 12 
sum_series(10) == 30 
sum_series(9) == 25 
\end{verbatim}

\textbf{Before DPO. Majority, 3 agents, public 0/3.} 
\begin{verbatim} 
def sum_series(n): 
    total = 0 
    x = 2 
    while n - x >= 0: 
        total += n - x 
        x += 2 
    return total 
\end{verbatim}

\textbf{Before DPO. Execution confidence, 1 agent, public 3/3.} 
\begin{verbatim} 
def sum_series(n): 
    total = 0 
    x = 0 
    while n - x >= 0: 
        if n - x > 0: 
            total += n - x 
        x += 2 
    return total 
\end{verbatim}

\textbf{After DPO. Majority, 3 agents, public 0/3.} 
\begin{verbatim} 
def sum_series(n): 
    total = 0 
    x = 2 
    while n - x >= 0: 
        total += n - x 
        x += 2 
        n -= 2 
    return total 
\end{verbatim}

\textbf{After DPO. Execution confidence, 1 agent, public 3/3.} 
\begin{verbatim} 
def sum_series(n): 
    total = 0 
    x = 0
    while n - x >= 0: 
        total += n - x 
        x += 2 
    return total
\end{verbatim} 
\end{tcolorbox}

\end{document}